\documentclass[twoside,11pt]{article}

\usepackage{appendix}
\usepackage{mathtools}
\usepackage[abbrvbib]{styles/jmlr2e}

\usepackage{dsfont}
\usepackage[shellescape,nowrite]{styles/gmp}
\usepackage{xcolor}
\usepackage{enumitem}
\usepackage{wrapfig}
\usepackage{booktabs}
\usepackage{microtype}

\usepackage{subcaption}
\usepackage{algorithm}
\usepackage[noend]{algpseudocode}
\newcommand\epsilonDelta{\{ (\varepsilon_t, \delta_t) \}_{t=1}^T}
\DeclareMathOperator\Span{span}
\newcommand\Minkowski{\rho}
\newcommand\Contraction{\kappa}
\newcommand\ALPHABET{\mathsf}
\newcommand\IND{\mathds{1}}

\newcommand\GRAD{\nabla}

\newcommand\PR {\mathbb{P}}
\newcommand\EXP{\mathbb{E}}
\newcommand\EXPsAIS{\EXP_{\Ais_t \sim \ainfo^s_t(\his_t)}}
\newcommand\EXPsAISt{\EXP_{\Ais_{t+1} \sim \ainfo^s_{t+1}(\his_{t+1})}}

\newcommand*\info{\sigma}
\newcommand*\ainfo{\hat \info}
\newcommand*\aquant{\psi}
\newcommand*\rewinfo{\hat r}
\newcommand*\nextinfo{\hat P}
\newcommand*\nextobs{\hat P^{\ob}}
\newcommand*\hide[1]{\ignorespaces}
\newcommand*\update{\varphi}
\newcommand*\aupdate{\hat{\update}}

\newcommand*\new{\textsc{new}}

\newcommand{\TV}{\mathrm{TV}}
\newcommand{\KL}{\mathrm{KL}}
\newcommand{\Lip}{\mathrm{Lip}}
\newcommand*\DEFINED{\coloneqq}

\DeclareMathOperator\Supp{Supp}
\newcommand\reals{\mathbb{R}}
\newcommand\integers{\mathbb{Z}}
\newcommand\TRANS{\intercal}
\newtheorem{assumption}{Assumption}

\newcommand{\St}{S} % State - RV
\newcommand{\st}{s} % Realization of state
\newcommand{\StSp}{\ALPHABET{\St}} % State space
\newcommand{\Ob}{Y}
\newcommand{\ob}{y}
\newcommand{\ObSp}{\ALPHABET{\Ob}}
\newcommand{\aOb}{\hat Y}

\newcommand{\aObSp}{\hat {\ALPHABET{\Ob}}}
\newcommand{\Act}{A}
\newcommand{\act}{a}
\newcommand{\ActSp}{\ALPHABET{\Act}}

\newcommand{\aact}{\hat a}
\newcommand{\aActSp}{\hat {\ALPHABET{\Act}}}
\newcommand{\W}{W}

\newcommand{\F}{\mathfrak{F}}
\newcommand{\WSp}{\ALPHABET{\W}}

\newcommand{\pol}{\pi}
\newcommand\polPars{\theta}

\newcommand{\apol}{\hat{\pi}}
\newcommand{\His}{H}
\newcommand{\his}{h}
\newcommand{\HisSp}{\ALPHABET{\His}}
\newcommand{\aHis}{\hat H}

\newcommand{\aHisSp}{\hat {\ALPHABET{\His}}}

\newcommand{\Istruct}{I}

\newcommand{\IstructSp}{\ALPHABET{\Istruct}}
\newcommand{\Is}{Z}
\newcommand{\is}{z}
\newcommand{\IsSp}{\ALPHABET{\Is}}
\newcommand{\Ais}{\hat Z}
\newcommand{\ais}{\hat z}
\newcommand{\AisSp}{\hat{\ALPHABET{Z}}}
\newcommand{\Rew}{R} % Reward - RV
\newcommand{\RewFn}{r} % Reward function

\newcommand\discount{\gamma}
\newcommand{\perf}{J}

\newcommand{\Psc}{\hat \xi}
\newcommand{\PSC}{\hat \Xi}
\newcommand{\PscSp}{\mathcal{X}}
\DeclareMathOperator\Linear{Linear}
\DeclareMathOperator\Tanh{Tanh}
\DeclareMathOperator\LSTM{LSTM}
\DeclareMathOperator\Relu{ReLU}
\DeclareMathOperator\Softmax{Softmax}
\DeclareMathOperator\GMM{GMM}
\DeclareMathOperator\ELU{ELU}
\DeclareMathOperator\Conv{Conv2d}
\DeclareMathOperator\MP{MaxPool2d}

\newcommand\Arrow{\rotatebox[origin=c]{-90}{$\Rightarrow$}}
\newcommand\LLArrow{\rotatebox[origin=c]{-135}{$\Rightarrow$}}
\newcommand\LRArrow{\rotatebox[origin=c]{-45}{$\Rightarrow$}}

\usepackage{acronym}

\newacro{AIS}{approximate information state}
\newacro{IPM}{integral probability metric}

\newcommand*\blue{\textcolor{black}}

\ShortHeadings{Approximate information state}{Subramanian, Sinha, Seraj, and Mahajan}
\firstpageno{1}

\begin{document}

\title{Approximate information state for approximate planning and
reinforcement learning in partially observed systems}

\author{\name Jayakumar Subramanian\thanks{This work was done when Jayakumar
  Subramanian was at McGill University.} \email jasubram@adobe.com \\
  \addr Media and Data Science Research Lab, Digital Experience Cloud,\\
  Adobe Systems India Private Limited, Noida, Uttar Pradesh, India\\
  %Plot A5, Noida-Greater Noida Expressway, Block A, Sector 132, 201304\\
       \AND
       \name Amit Sinha \email amit.sinha@mail.mcgill.ca \\
       \name Raihan Seraj \email raihan.seraj@mail.mcgill.ca \\
       \name Aditya Mahajan \email aditya.mahajan@mcgill.ca \\
       \addr Department of Electrical and Computer Engineering \\
       McGill University, 
       Montreal, QC, Canada}

\editor{}

\maketitle

\begin{abstract}
We propose a theoretical framework for approximate planning and learning in partially observed systems. Our framework is based on the fundamental notion of information state. We provide two equivalent definitions of information state---i) a function of history which is sufficient to compute the expected reward and predict its next value; ii) equivalently, a function of the history which can be recursively updated and is sufficient to compute the expected reward and predict the next observation. An information state always leads to a dynamic programming decomposition. Our key result is to show that if a function of the history (called \acf{AIS}) approximately satisfies the properties of the information state, then there is a corresponding approximate dynamic program. We show that the policy computed using this is approximately optimal with bounded loss of optimality. We show that several approximations in state, observation and action spaces in literature can be viewed as instances of \acs{AIS}. In some of these cases, we obtain tighter bounds. A salient feature of \acs{AIS} is that it can be learnt from data. We present \acs{AIS} based  multi-time scale policy gradient algorithms. and detailed numerical experiments with low, moderate and high dimensional environments.
\end{abstract}


\section{Introduction}

Reinforcement learning (RL) provides a conceptual framework for designing
agents which learn to act optimally in an unknown environment. RL has been
successfully used in various applications ranging from robotics, industrial
automation, finance, healthcare, and natural language processing. The success
of RL is based on a solid foundation of combining the theory of exact and
approximate Markov decision processes (MDPs) with iterative algorithms that
are guaranteed to learn an exact or approximate action-value function and/or
an approximately optimal policy \citep{SuttonBarto_2018,
BertsekasTsitsiklis1996}. However, for the most part, the research on RL
theory is focused primarily on systems with full state observations. 

In various applications including robotics, finance, and healthcare, the agent
only gets a partial observation of the state of the environment. Such partially
observed systems are mathematically modeled as partially observable Markov
decision processes (POMDPs) and there is a fairly good understanding of how to
identify optimal or approximately optimal policies for POMDPs when the system
model is known to the agent. 

Since the initial work on POMDPs \citep{Astrom_1965}, it is known that POMDPs can be
modeled as fully observed MDPs by considering the belief state (i.e., the
posterior belief of the unobserved state given all the observations made by the
agent) as an information state. Therefore, the theory and algorithms for exact
and approximate planning for MDPs are also applicable to POMDPs. One
computational challenge is that the belief state is continuous valued.
However, the value function based on the belief state has a nice property---it
is piecewise linear and a convex function of the belief state---which can be
exploited to develop efficient algorithms to identify the optimal policy.
Building on the one-pass algorithm of \citep{SmallwoodSondik_1973}, various
such algorithms have been proposed in the literature including the linear
support algorithm \citep{Cheng1988}, the witness algorithm
\citep{CassandraKaelblingLittman_1994}, incremental pruning \citep{Zhang1996,
Cassandra1997}, the duality based approach \citep{Zhang2009}, and others. Since
POMDPs are PSPACE-complete \citep{Papadimitriou1999}, the worst case complexity of
such algorithms is exponential in the size of the unobserved state space. To
overcome the worst case complexity of finding an optimal policy, various
point-based methods have been proposed in the literature which obtain an
approximate solution by sampling from the belief space \citep{Pineau2003,
Smith2004, Spaan2005, Shani2007, Kurniawati2008, Poupart2011}; see
\cite{ShaniPineauKaplow_2013} for an overview and comparison. 

However, the exact and approximate planning results are of limited value for
partially observed reinforcement learning (PORL) because they are based on the belief state, constructing which
requires the knowledge of the system model. So, when an agent is operating in
an unknown environment, it cannot construct a belief state based on its
observations. An attempt to circumvent this difficulty was to use
memoryless policies (i.e., choose the action based only on the current
observation) \citep{Littman1994a,Loch1998,Jaakkola1998, Williams1999, Li2011,
Azizzadenesheli2016}. A related idea is to choose the action based on $k$
recent observations \citep{Littman1994a, Loch1998} or choose the action
based on a memory which is updated using a finite state machine
\citep{Whitehead1995,McCallum_1993,Hansen1997,Meuleau1999,Amato2010}. Such
finite memory policies are also amenable to policy search methods
\citep{Hansen1998,Baxter2001,Poupart2004}. However, there are no approximation
guarantees available for such methods.

Another approach taken in the literature is to use a Bayesian RL framework
\citep{Ross2008,Poupart2008,Ross2011,Katt2019} where a posterior distribution
over the models of the environment is maintained; at each step, a model is
sampled from the posterior and the corresponding optimal policy is executed.
Appproximation error bounds in using such methods are derived in
\cite{Ross2011}.

A completely different class of model-based RL algorithms are methods
using predictive state representations (PSRs) \citep{LittmanSutton_2002,
SinghLittmanJongPardoeStone_2003}. PSRs are constructed only based on
observational data so they can easily be adapted to the RL setup. There have
been a number of papers which use PSRs to propose model based RL algorithms
\citep{James2004,RosencrantzGordonThrun_2004, Boots2011, Hamilton2014,
Kulesza2015, Kulesza2015a, Jiang2016}.

Inspired by the recent successes of deep reinforcement learning, there are
many recent results which suggest using RNNs (Recurrent Neural
Networks~\citep{Rumelhart1986}) or LSTMs (Long Short-Term
Memories~\citep{Hochreiter1997}) for modeling the action-value
function and/or the policy function
\citep{bakker2002reinforcement,WierstraFoersterPetersSchmidhuber_2007,
WierstraFoersterPetersSchmidhuber_2010,HausknechtStone_2015,HeessHuntLillicrapSilver_2015,Zhu2017,ha2018world,
BaiseroAmato_2018,Igl2018,Zhang2019}. It is shown that these approaches perform well on empirical benchmarks, but there are no approximation guarantees available for such methods.

Our main contribution is to present a rigorous approach for PORL which is
based on a principled theory of approximate planning for POMDPs that we
develop.
In particular:
\begin{enumerate}
  \item In Sec.~\ref{sec:prelim}, we formalize the notion of information state for partially observed
      systems and provide equivalent methods of identifying information
      states. 

    \item In Secs.~\ref{sec:ais} and~\ref{sec:infinite}, we present the notion
      of an \acf{AIS} as a compression of history which approximately
      satisfies the properties of an information state. The two equivalent
      formulations of information state lead to two equivalent formulations of
      \acs{AIS}. We present bounds on the loss in performance (compared to the
      optimal history dependent policy) when planning using an \acs{AIS}. We
      generalize these results to cover approximation in action spaces as
      well. We show that various existing approximation results for MDPs and
      POMDPs in the literature may be viewed as special cases of \acs{AIS}
      (and in some cases, our bounds are tighter than those in the
      literature). 

    \item In Sec.~\ref{sec:dec}, we present a theory for approximate planning
      for decentralized (i.e., multi-agent) partially observed systems using a
      common-information based \acs{AIS}.

    \item In Secs.~\ref{sec:RL} and~\ref{sec:experiments}, we then present
      policy gradient based online RL algorithms for PORL which learn an
      \acs{AIS} representation using multi-timescale stochastic gradient
      descent. We provide detailed numerical experiments on several classes
      of partially observed environments ranging from classical low-dimensional toy
      environments, to moderate-dimensional environments, and high-dimensional grid-world
      environments.
\end{enumerate}

\section{Preliminaries: Information state and dynamic programming
decomposition for partially observed systems}\label{sec:prelim}

\subsection{General model for a partially observed system} \label{sec:model}

Traditionally, partially observed systems are modeled as partially observable
Markov decision processes (POMDPs)~\citep{Astrom_1965,SmallwoodSondik_1973}, where
there is a controlled state and an agent which makes noise corrupted
observations of the state. However, for the purpose of understanding
approximation for partially observed systems, it is conceptually cleaner to
start with an input-output model of the system as described below.

\begin{figure}[ht]
  \begin{minipage}{0.475\textwidth}
    \centering
    \resizebox{0.95\linewidth}{!}{%
    \begin{mpost}[mpsettings={input boxes;}]
      defaultdx := 10bp;
      defaultdy := 20bp;
      boxit.system(\btex System etex);
      system.c = origin;

      drawboxed(system);

      z1 = 0.5[system.w, system.nw];
      z2 = 0.5[system.w, system.sw];

      z3 = z1 - (1cm,0);
      z4 = z2 - (1cm,0);

      drawarrow z3 -- lft z1;
      drawarrow z4 -- lft z2;

      label.lft(\btex Stochastic input $W_t$ etex, z3);
      label.lft(\btex Controlled input $A_t$ etex, z4);

      z5 = 0.5[system.e, system.ne];
      z6 = 0.5[system.e, system.se];

      z7 = z5 + (1cm, 0);
      z8 = z6 + (1cm, 0);
      
      drawarrow z5 -- z7;
      drawarrow z6 -- z8;
      label.rt(\btex Observation $Y_t$ etex, z7);
      label.rt(\btex Reward $R_t$ etex, z8);
    \end{mpost}}
    \caption{A stochastic input-output system}
    \label{fig:model}
  \end{minipage}
  \hfill
  \begin{minipage}{0.475\textwidth}
    \centering
    \resizebox{0.95\linewidth}{!}{%
    \begin{mpost}
      draw (-0.5cm,0) -- (3cm, 0);
      draw (3cm, 0) -- (4.5cm, 0) dashed evenly;
      draw (4.5cm, 0) -- (6.5cm, 0);
      drawarrow (6.5cm, 0) -- (8cm, 0) dashed evenly;

      drawarrow (0, -0.75cm) -- (0, 0);
      label.bot(\btex $A_1$ etex, (0, -0.75cm));

      drawarrow (0.5cm, -0.75cm) -- (0.5cm, 0);
      label.bot(\btex $W_1$ etex, (0.5cm, -0.75cm));

      drawarrow (1cm, 0) -- (1cm, 0.75cm);
      label.top(\btex $(Y_1, R_1)$ etex, (1cm, 0.75cm));

      drawarrow (1.5cm, -0.75cm) -- (1.5cm, 0);
      label.bot(\btex $A_2$ etex, (1.5cm, -0.75cm));

      drawarrow (2cm, -0.75cm) -- (2cm, 0);
      label.bot(\btex $W_2$ etex, (2cm, -0.75cm));

      drawarrow (2.5cm, 0) -- (2.5cm, 0.75cm);
      label.top(\btex $(Y_2, R_2)$ etex, (2.5cm, 0.75cm));

      drawarrow (5cm, -0.75cm) -- (5cm, 0);
      label.bot(\btex $A_t$ etex, (5cm, -0.75cm));

      drawarrow (5.5cm, -0.75cm) -- (5.5cm, 0);
      label.bot(\btex $W_t$ etex, (5.5cm, -0.75cm));

      drawarrow (6cm, 0) -- (6cm, 0.75cm);
      label.top(\btex $(Y_t,R_t)$ etex, (6cm, 0.75cm));
    \end{mpost}}
    \caption{The timing diagram of the input-output system.}
    \label{fig:timeline}
  \end{minipage}
\end{figure}

We view a partially observed system as a black-box input-output system shown
in Fig.~\ref{fig:model}. At each time~$t$, the system has two inputs and
generates two outputs. The inputs to the system are a control input (also
called an action) $\Act_t \in \ActSp$ and a disturbance $\W_t \in \WSp$.
The outputs of the system are an observation $\Ob_t \in \ObSp$ and a reward
$\Rew_t \in \reals$. For the ease of exposition, we assume that $\ActSp$,
$\WSp$, and $\ObSp$ are finite sets. The analysis extends to general spaces
under appropriate technical conditions. The order in which the input and
output variables are generated is shown in Fig.~\ref{fig:timeline}. 

As stated before, we do not impose a state space model on the system.
Therefore, all we can say is that the outputs $(\Ob_t, \Rew_t)$ at time~$t$
are some function of all the inputs $(\Act_{1:t}, \W_{1:t})$ up to time~$t$,
i.e.,

\[
  \Ob_{t} = f_t(\Act_{1:t}, \W_{1:t})
  \quad\text{and}\quad
  \Rew_t = \RewFn_t(\Act_{1:t}, \W_{1:t}),
\]
where $\{f_t \colon \ActSp^{t} \times \WSp^{t} \to \ObSp \}_{t = 1}^{T}$ 
are called the system output functions and $\{\RewFn_t \colon \ActSp^{t} \times \WSp^{t} \to \reals \}_{t=1}^T$ are called the 
system reward functions. 

There is an agent which observes the output $\Ob_t$ and generates a control
input or the action $\Act_t$ as a (possibly stochastic) function of the
history $\His_t = (\Ob_{1:t-1}, \Act_{1:t-1})$ of the past observations and
actions, i.e., 
\[
  \Act_t \sim \pol_t(\His_t),
\]
where $\pol \DEFINED (\pol_t)_{t \ge 1}$ is a (history-dependent and possibly
stochastic) policy. We
use $\HisSp_t$ to denote the space of all histories up to time~$t$. Then the
policy $\pol_t$ is a mapping from $\HisSp_t$ to $\Delta(\ActSp)$ (which denotes
the space of probability measures on $\ActSp$). We will use
$\pol_t(\act_t|\his_t)$ to denote the probability of choosing action $\act_t$
at time~$t$ given history $h_t$ and use $\Supp(\pol_t(\his_t))$ to denote the
support of $\pol_t$ (i.e., the set of actions chosen with positive
probability).

We assume that the disturbance $\{\W_t\}_{t \ge 1}$ is a sequence of independent random
variables defined on a common probability space $(\Omega, \mathcal{F}, \PR)$.
Thus, if the control input process $\{\Act_{t} \}_{t \ge 1}$ is specified, then the
output processes $\{ \Ob_{t}, \Rew_{t}\}_{t \ge 1}$ are random variables on
$(\Omega, \mathcal{F}, \PR)$. Specifying a policy $\pol$ for the agent
induces a probability measure on the output processes $\{ \Ob_t, \Rew_t\}_{t \ge
1}$, which we denote by $\PR^{\pol}$. 

We start our discussion by looking at the planning problem faced by the agent
when the system runs for a finite horizon~$T$. We will generalize our results
to the infinite horizon discounted reward setup later. In the finite horizon
setup, the performance of any policy $\pol$ is given by
\begin{equation} \label{eq:optimal}
  J(\pol) \coloneqq \EXP^{\pol}\biggl[ \sum_{t=1}^T \Rew_t \biggr],
\end{equation}
where $\EXP^{\pol}$ denotes the expectation with respect to the probability
measure $\PR^{\pol}$. 

We assume
that the agent knows the system dynamics $\{f_t\}_{t \ge 1}$, the reward
functions $\{r_t\}_{t\ge 1}$, and the probability measure $\PR$ on the
primitive random variables $\{\W_t\}_{t\ge1}$. The objective of the agent is
to choose a policy $\pol$ which maximizes the expected total reward
$\perf(\pol)$. 

Since all system variables are assumed to be finite valued and the
system runs for a finite horizon, there are only a finite number of policies
$\pol$. So, an optimal policy always exists and the important question is to
determine an efficient algorithm to compute the optimal policy.

In Sec.~\ref{sec:history}, we start by presenting a trivial dynamic
programming decomposition which uses the entire history of observations as a
state. Such a history-dependent dynamic program is not an efficient method to
compute the optimal policy; rather it serve as a reference with which we
compare the more efficient exact and approximate dynamic programs that we
derive later. 

In Sec.~\ref{sec:info-state}, we present sufficient conditions to identify an
information state for dynamic programming. Our main result, presented in
Secs.~\ref{sec:ais} and~\ref{sec:infinite}, is to identify a notion of
approximate information state and derive approximation bounds when an
approximate policy is computed using an approximate information state.

\subsection{A dynamic programming decomposition}\label{sec:history}

To obtain a dynamic program to identify an optimal policy
for~\eqref{eq:optimal}, we can view the history $\His_t$ as a ``state'' of a
Markov decision process (MDP) with transition probability 
\[
  \PR(\His_{t+1} = (\his'_t, \act'_t, \ob_t) \mid \His_t = \his_t, \Act_t = \act_t)
  = \begin{cases}
    \PR(\Ob_t = \ob_t | \His_t = \his_t, \Act_t = \act_t), &
    \text{if $\his'_t = \his_t$ \& $\act'_t = \act_t$} \\
    0, & \text{otherwise}
  \end{cases}
\]
and per-step reward $\EXP[ \Rew_t | \His_t, \Act_t]$. Therefore, from standard
results from Markov decision processes \cite{Bellman1957}, we can recursively
compute the performance of a given policy as well as the best possible
performance using ``standard'' dynamic program.

\begin{proposition}[Policy evaluation]\label{prop:eval}
  For any given (history dependent) policy $\pol$, define the
  \emph{reward-to-go} function for any time~$t$ and realization $\his_t$ of
  history $\His_t$ as
  \begin{equation}\label{eq:J1_defn}
    V^\pol_t(\his_t) \coloneqq \EXP^{\pol}\biggl[ \sum_{s=t}^T \Rew_s \biggm| 
    \His_t = \his_t \biggr]. 
  \end{equation}
  The reward-to-go functions defined above satisfy the following recursion.
  Define $V^\pol_{T+1}(h_{T+1}) = 0$ and for any $t \in \{T, \dots, 1\}$, 
  \begin{equation} \label{eq:pol-eval}
    V^\pol_t(\his_t) = \EXP^\pol
    \bigl[ \Rew_t + V^\pol_{t+1}(\His_{t+1}) \bigm| \His_t = \his_t \bigr].
  \end{equation}
\end{proposition}
The reward-to-go function $V^\pol_t(\his_t)$ denotes the expected cumulative
rewards obtained in the future when starting from history $\his_t$ at time~$t$
and following policy~$\pol$.
Note that $V^\pol_t(\his_t)$ only depends on the policy $\pol$ only through
the choice of the future policy $(\pol_{t}, \dots, \pol_T)$ and therefore can
be computed without the knowledge of the past policy $(\pol_1, \dots, \pol_{t-1})$. 

Note that $h_1 = \emptyset$ and the performance $J(\pol)$ defined
in~\eqref{eq:optimal} equals $V^\pol(h_1)$. Thus, Proposition~\ref{prop:eval}
gives a recursive method to evaluate the performance of any history dependent
policy~$\pol$. Following the standard argument for Markov decision processes,
we can modify the recursion~\eqref{eq:pol-eval} to obtain a
dynamic program to identify an optimal policy as follows.

\begin{proposition}[Dynamic programming]\label{prop:optimal}
  Recursively define \emph{value functions} $\{ V_t \colon \ALPHABET H_t \to
  \reals\}_{t=1}^{T+1}$ as follows. $V_{T+1}(\His_{t+1}) \coloneqq 0$ and for $t \in
  \{T, \dots 1\}$, 
  \begin{equation} \label{eq:DP-hist}
    V_t(\his_t) \coloneqq \max_{\act_t \in \ActSp} \EXP
    \bigl[ \Rew_t + V_{t+1}(\His_{t+1}) \bigm| \His_t = \his_t, \Act_t = \act_t \bigr].
  \end{equation}
  Then, a stochastic policy $\pol = (\pol_1, \dots, \pol_T)$ is
  optimal if and only if for all $t \in \{1, \dots T\}$ it satisfies
  \begin{equation} \label{eq:opt-policy}
    \Supp(\pol_t(\his_t)) \subseteq \arg\max_{\act_t \in \ActSp} \EXP
    \bigl[ \Rew_t + V_{t+1}(\His_{t+1}) \bigm| \His_t = \his_t, \Act_t = \act_t
    \bigr].
  \end{equation}
\end{proposition}
Note that the expectation in~\eqref{eq:DP-hist} is with respect to the
probability measure $\PR$ on $(\Omega, \mathcal{F})$ and can be computed
without the knowledge of the policy~$\pol$.

\subsection{Information state and simplified dynamic programs}
\label{sec:info-state}

The dynamic program of Proposition~\ref{prop:optimal} uses the entire history
as state and may not be efficient for identifying an optimal policy. In this
section, we present a general class of dynamic programming decompositions
which may be more efficient. This class of dynamic programs is based on the
notion of information state, which we describe next. 

\begin{definition}\label{def:info_state}
  Let $\{\IsSp_t\}_{t=1}^T$ be a pre-specified collection of Banach spaces. A
  collection $\{\info_t \colon \HisSp_t \to \IsSp_t \}_{t=1}^T$ of history
  compression functions is called an \emph{information state generator} if the
  process $\{\Is_t\}_{t =1}^T$, where $\Is_t = \info_t(\His_t)$, satisfies the
  following properties:
  \begin{description}
    \item[(P1)] \textbf{\textup{Sufficient for performance evaluation}}, i.e., 
      for any time~$t$, any realization $\his_t$ of $\His_t$ and any choice
      $\act_t$ of $\Act_t$, we have
      \[
        \EXP[ \Rew_t \mid \His_t = \his_t, \Act_t = \act_t ] =
        \EXP[ \Rew_t \mid \Is_t = \info_t(\his_t), \Act_t = \act_t].
      \]

    \item[(P2)] \textbf{\textup{Sufficient to predict itself}}, i.e., 
      for any time~$t$, any realization $\his_t$ of $\His_t$ and any choice
      $\act_t$ of $\Act_t$, we have that for any Borel subset $\ALPHABET B$ of
      $\IsSp_{t+1}$,
      \[
        \PR(\Is_{t+1} \in \ALPHABET B \mid \His_t = \his_t, \Act_t = \act_t ) = 
        \PR(\Is_{t+1} \in \ALPHABET B \mid \Is_t = \info_t(\his_t), \Act_t = \act_t ).
      \]
  \end{description}
\end{definition}

In the sequel, we will sometimes use the phrase ``let $\{\Is_t\}_{t=1}^T$ be an
information state'' to specify an information state and will implicitly assume
that the corresponding information state spaces are $\{\IsSp_t\}_{t=1}^T$ and
the corresponding compression functions are $\{\info_t\}_{t=1}^T$.

Note that both the probabilities in Property (P2) can be computed without the
knowledge of the policy~$\pol$. Furthermore, there are no restrictions on the spaces
$\{\IsSp_t\}_{t =1}^T$ although in practice an information state is useful
only when these spaces are ``small'' in an appropriate sense. 

Condition (P1) is easy to verify but condition (P2) can be a bit abstract. 
For some models, instead of (P2), it is easier to verify the following
stronger conditions:
\begin{description}
\item[(P2a)] \textbf{Evolves in a state-like manner}, i.e., there exist
  measurable functions $\{\update_t\}_{t=1}^T$ such that for any time~$t$ and
  any realization $\his_{t+1}$ of $\His_{t+1}$, we have
  \[
    \sigma_{t+1}(\his_{t+1}) = \update_t( \sigma_t(\his_t), \ob_t, \act_t).
  \]
  Informally, the above condition may be written as
  \(
    \Is_{t+1} = \update_t(\Is_t, \Ob_{t}, \Act_t).
  \)
\item[(P2b)] \textbf{Is sufficient for predicting future observations}, i.e.,
  for any time~$t$, any realization $\his_t$ of $\His_t$ and any choice
  $\act_t$ of $\Act_t$, we have that for any subset $\ALPHABET D$ of $\ObSp$, 
  \[
    \PR(\Ob_{t} \in \ALPHABET D \mid \His_t = \his_t, \Act_t = \act_t) = 
    \PR(\Ob_{t} \in \ALPHABET D \mid \Is_t = \info_t(\his_t), \Act_t = \act_t).
  \]
\end{description}

\begin{proposition}\label{prop:alt-info-state}
  \textup{(P2a)} and \textup{(P2b)} imply \textup{(P2)}.
\end{proposition}
\begin{proof}
  For any Borel subset $\ALPHABET D$ of $\IsSp_{t+1}$, we have
  \begin{align*}
    \PR(\Is_{t+1} \in \ALPHABET D &\mid \His_t = \his_t, \Act_t = \act_t) \\
  &\stackrel{(a)}{=}\sum_{\ob_{t} \in \ObSp}
  \PR(\Ob_{t} = \ob_{t}, \Is_{t+1} \in \ALPHABET D \mid \His_t = \his_t, \Act_t = \act_t)\\
  &\stackrel{(b)}{=}\sum_{\ob_{t} \in \ObSp}
  \IND\{\update_t(\info_t(\his_t),\ob_{t},\act_t) \in \ALPHABET D \}
   \PR(\Ob_{t} = \ob_{t} \mid \His_t = \his_t, \Act_t = \act_t)\\
  &\stackrel{(c)}{=}\sum_{\ob_{t} \in \ObSp}
  \IND\{\update_t(\info_t(\his_t),\ob_{t},\act_t) \in \ALPHABET D \}
   \PR(\Ob_{t} = \ob_{t} \mid \Is_t = \info_t(\his_t), \Act_t = \act_t)\\
  &\stackrel{(d)}{=}\PR(\Is_{t+1} \in \ALPHABET D \mid \Is_t = \info_t(\his_t), \Act_t = \act_t)
  \end{align*}
  where $(a)$ follows from the law of total probability, $(b)$ follows from
  (P2a), $(c)$ follows from (P2b) and $(d)$ from the law of total probability.
\end{proof}

\blue{
The following example illustrates how (P2a) and (P2b) are stronger conditions than 
(P2). Consider a Markov decision process (MDP) with state $(S^1_t, S^2_t) \in
\mathcal{S}^1 \times \mathcal{S}^2$ and action $A_t \in \mathcal{A}$, where the
dynamics of the two components of the state are conditionally independent
given the action, i.e., 
\begin{multline*}
  \PR(S^1_{t+1} = s^1_{+}, S^2_{t+1} = s^2_{+} | S^1_t = s^1, S^2_t = s^2, A_t = a)
  \\
  = 
  \PR(S^1_{t+1} = s^1_{+} | S^1_t = s^1, A_t = a)
  \PR(S^2_{t+1} = s^2_{+} | S^2_t = s^2, A_t = a).
\end{multline*}
Furthermore, suppose the reward $R_t$ at any time $t$ is given by $R_t  =
r_t(S^1_t, A_t)$. Since the model is an MDP, the observation at time~$t$ is
the same as the state. For this model, the component $\{S^1_t\}_{t\ge1}$ of the state
satisfies properties (P1) and (P2). Therefore, $\{S^1_t\}_{t \ge 1}$ is an
information state process. However, $\{S^1_t\}_{t \ge 1}$ is not sufficient to
predict the next observation $(S^1_{t+1}, S^2_{t+1})$. Therefore, $\{S^1_t\}_{t
\ge 1}$ does not satisfy property (P2b). This shows that properties (P2a) and
(P2b) are stronger than property (P2). 
The above example may be considered as an instance of what is called the Noisy-TV 
problem~\citep{burda2018exploration}.
}

Next, we show that an information state is useful because it is
always possible to write a dynamic program based on the information state. To
explain this dynamic programming decomposition, we first write the
history-based dynamic programs of Proposition~\ref{prop:eval}
and~\ref{prop:optimal} in a more compact manner as follows: Let
$V_{T+1}(\his_{T+1}) \coloneqq 0$ and for $t \in \{T, \dots, 1\}$, define
\begin{subequations}\label{eq:DP-opt}
\begin{align}
  Q_t(\his_t, \act_t) &\coloneqq \EXP\bigl[ \Rew_t + V_{t+1}(\His_{t+1}) \bigm|
  \His_t = \his_t, \Act_t = \act_t \bigr],
  \\
  V_t(\his_t) &\coloneqq \max_{\act_t \in \ActSp} Q_t(\his_t, \act_t).
\end{align}
\end{subequations}
The function $Q_t(\his_t, \act_t)$ is called the action-value function.
Moreover, for a given stochastic policy $\pol = (\pol_1, \dots, \pol_T)$,
where $\pol_t \colon \HisSp_t \to \Delta(\ActSp_t)$, let
$V^\pol_{T+1}(\his_{T+1}) = 0$ and for $t \in \{T, \dots, 1\}$, define
\begin{subequations} \label{eq:DP-eval}
\begin{align}
  Q^\pol_t(\his_t, \act_t) &\coloneqq \EXP\bigl[ \Rew_t + V^\pol_{t+1}(\His_{t+1}) \bigm|
  \His_t = \his_t, \Act_t = \act_t \bigr],
  \\
  V^\pol_t(\his_t) &\coloneqq  \sum_{\act_t \in \ActSp} \pol_t(\act_t \mid \his_t).
  Q^\pol_t(\his_t, \act_t).
\end{align}
\end{subequations}

\begin{theorem}\label{thm:info-state}
  Let $\{\Is_t\}_{t=1}^T$ be an information state. Recursively
  define value functions $\{\bar V_t \colon \IsSp_t \to \reals\}_{t=1}^{T+1}$, 
  as follows: $\bar V_{T+1}(\is_{T+1}) \coloneqq 0$ and for $t \in
  \{T, \dots, 1\}$:
  \begin{subequations}\label{eq:DP-info}
  \begin{align}
    \bar Q_t(\is_t, \act_t) &\coloneqq \EXP[ \Rew_t + \bar V_{t+1}(\Is_{t+1}) \mid
    \Is_t = \is_t, \Act_t = \act_t ],  \\ 
    \bar V_t(\is_t) &\coloneqq \max_{\act_t \in \ActSp} \bar Q_t(\is_t, \act_t).
  \end{align}
  \end{subequations}

  Then, we have the following:
  \begin{enumerate}
    \item For any time~$t$, history $\his_t$, and action $\act_t$, we have
      that
      \begin{equation} \label{eq:equiv}
        Q_t(\his_t, \act_t) = \bar Q_t(\info_t(\his_t), \act_t)
        \text{ and }
        V_t(\his_t) = \bar V_t(\info_t(\his_t)). 
  %  \pi_t(h_t) &= \bar \pi_t(\info_t(h_t)).
      \end{equation}
    \item Let $\bar \pol = (\bar \pol_1, \dots \bar \pol_T)$,
      where $\bar \pol_t \colon \IsSp_t \to \Delta(\ActSp)$, be a stochastic
      policy. Then, the policy $\pol = (\pol_1, \dots, \pol_T)$ given by
      $\pol_t = \bar \pol_t \circ \info_t$ is optimal if and
      only if for all $t$ and all realizations $\is_t$ of information states
      $\Is_t$, $\Supp(\bar \pol_t(\is_t)) \subseteq \arg \max_{\act_t \in
      \ActSp} \bar Q_t(\is_t, \act_t)$. 
  \end{enumerate}
\end{theorem}
\begin{proof}
  We prove the result by backward induction. By construction, \eqref{eq:equiv}
  is true at time $T+1$. This forms the basis of induction. Assume
  that~\eqref{eq:equiv} is true at time $t+1$ and consider the system at
  time~$t$. Then, 
  \begin{align*}
    Q_t(\his_t, \act_t) &= \EXP[ \Rew_t + V_{t+1}(\His_{t+1}) \mid \His_t = \his_t, \Act_t = \act_t ]\\
    &\stackrel{(a)}= \EXP[ \Rew_t + \bar V_{t+1}(\info_{t+1}(\His_{t+1})) \mid \His_t
    = \his_t, \Act_t = \act_t ] 
    \\
    &\stackrel{(b)}= \EXP[ \Rew_t + \bar V_{t+1}(\Is_{t+1}) \mid 
    \Is_t = \info_t(\his_t), \Act_t = \act_t]\\
    &\stackrel{(c)}= \bar Q_t(\info_t(\his_t), \act_t),
  \end{align*}
  where $(a)$ follows from the induction hypothesis, $(b)$ follows from the
  properties (P1) and (P2) of information state, and $(c)$ follows from the definition of $\bar
  Q$. This shows that the action-value functions are equal. By maximizing over
  the actions, we get that the value functions are also equal. The optimality
  of the policy follows immediately from~\eqref{eq:equiv}.
\end{proof}

\subsection{Examples of information state} \label{ex:info-state}

For a general model, it is not immediately evident that a non-trivial
information state exists. The question of existence will depend on the
specifics of the observation and reward functions $\{ f_t, r_t \}_{t \ge 1}$
as well as the properties of the probability measure on the primitive random
variables $\{ \W_t\}_{t \ge 1}$. We do not pursue the question of existence in
this paper, but present various specific models where information state
exists and show that the corresponding results for these models in the
literature may be viewed as
a special case of Theorem~\ref{thm:info-state}.

\begin{enumerate}
  \item For any partially observed model, the history $\His_t$ is always a
    trivial information state. Therefore, the dynamic program of
    Proposition~\ref{prop:optimal} may be viewed as a special case of
    Theorem~\ref{thm:info-state}.

  \item \textsc{Markov decision process (MDP):} Consider a Markov decision
    process (MDP) with state $\St_t \in \StSp$ and action $\Act_t \in \ActSp$
    \citep{Bellman1957}. At each time, the state evolves in a controlled
    Markovian manner with 
    \[
      \PR(\St_{t+1} = \st_{t+1} \mid \St_{1:t} = \St_{1:t}, \Act_{1:t} = \Act_{1:t})
      =
      \PR(\St_{t+1} = \st_{t+1} \mid \St_t = \St_t, \Act_t = \Act_t).
    \]
    The observation of the agent is $\Ob_t = \St_{t+1}$ and the reward output
    is $R_t = r(\St_t, \Act_t)$. An information state for an MDP is given by
    the current state $\St_t$ (the corresponding compression function
    is $\info_t(\St_{1:t}, \Act_{1:t-1}) = \St_t$). The standard dynamic
    program for MDPs may be viewed as a special case of
    Theorem~\ref{thm:info-state}.
    
  \item \textsc{Even MDPs:} Consider an MDP where the state space $\StSp$ is
    either $\reals$ or a symmetric subset of $\reals$ of the form $[-B, B]$,
    the controlled transition matrix is even, i.e., for every $\act \in
    \ActSp$ and $\st, \st' \in \StSp$,
    \[
      \PR(\St_{t+1} = \st' \mid \St_t = \st, \Act_t = \act) =
      \PR(\St_{t+1} = -\st' \mid \St_t = -\st, \Act_t = \act),
    \]
    and for every $\act \in \Act$, the per-step reward function $r(\st,
    \act)$ is even in $\st$. Such MDPs are called \emph{even}
    MDPs~\citep{Chakravorty2018} and an information state for such MDPs is
    given by the absolute value state $|\St_t|$ (the corresponding compression
    function is $\info_t(\St_{1:t}, \Act_{1:t-1}) = |\St_t|$). The dynamic
    program for even MDPs derived in \cite{Chakravorty2018} may be viewed as a
    special case of Theorem~\ref{thm:info-state}.

  \item \textsc{MDP with irrelevant components:} Consider an MDP with state space
    $\StSp = \StSp^1 \times \StSp^2$, action space $\ActSp$, transition matrix
    $P(\st^1_+, \st^2_+ | \st^1, \st^2, \act) = P^1(\st^1_{+} | \st^1, \act)
    P^2(\st^2_{+} | \st^1, \st^2, \act)$, and per-step reward $r(\st^1,
    \act)$, which does not depend on the second component of the state. As
    explained in \cite{Feinberg2005}, such models arise in control of queues
    and transformation of continuous time Markov decision processes to
    discrete time MDPs using uniformization. An information state for such
    MDPs is given by the first component $\St^1_t$ (the corresponding
      compression function is $\info_t(\St^1_{1:t}, \St^2_{1:t}, \Act_{1:t}) =
    \St^1_t$). The qualitative properties of optimal policies for such models
    derived in \cite{Feinberg2005} may be viewed as a special case of
    Theorem~\ref{thm:info-state}.

  \item \textsc{MDP with delayed state observation:} Consider an MDP where the
    observation $\Ob_t$ of the agent is the $\delta$-step delayed state
    $\St_{t-\delta+1}$ of the system \citep{Altman1992}. An information state
    for such MDPs is given by the vector $(\St_{t-\delta+1},
    U_{t-\delta+1:t-1})$. The dynamic program for such models derived in 
    \cite{Altman1992} may be viewed as a special case of
    Theorem~\ref{thm:info-state}.

  \item \textsc{Partially observable Markov decision processes (POMDPs):}
    Consider a partially observable Markov decision process (POMDP) where
    there is a state space model as for an MDP but the observation $\Ob_t$ is
    some function of the state and the disturbance, i.e., $\Ob_t =
    f^{\ob}_t(\St_t, \W_t)$ \citep{Astrom_1965, SmallwoodSondik_1973}. An
    information state for the POMDP is given by the belief state $B_t \in
    \Delta(\StSp)$ which is given by
    \(
      B_t(\st) = \PR(\St_t = \st \mid \His_t = \his_t)
    \). The corresponding compression function may be identified via the
    update functions $\{\update_t\}_{t=1}^T$ of Property~(P2a), which are the
    standard belief update functions for non-linear filtering. The standard
    belief state dynamic program for POMDPs \citep{Astrom_1965,
    SmallwoodSondik_1973} may be viewed as a special case of
    Theorem~\ref{thm:info-state}.

  \item \textsc{Linear quadratic and Gaussian (LQG) models:} Consider a POMDP where
    the state and action spaces are Euclidean spaces, the system dynamics
    $\PR(\St_{t+1} \mid \St_t, \Act_t)$ and the observation $f^\ob_t(\St_t,
    \W_t)$ are linear, the disturbance $\W_t$ is Gaussian, and the per-step
    \emph{cost} is a quadratic function of the state and action
    \citep{Astrom_1970}. For such a \emph{linear-quadratic-and-Gaussian}
    POMDP, an information state is given by the state estimate $\hat \St_t =
    \EXP[ \St_t \mid \His_t = \his_t]$. The corresponding compression function
    may be identified via the update functions $\{\update_t\}_{t=1}^T$ of
    Property~(P2a), which in this case are Kalman filtering update equations.
    The standard conditional estimate based
    dynamic program for LQG models \citep{Astrom_1970} may be viewed as a
    special case of Theorem~\ref{thm:info-state}.

  \item \textsc{POMDPs with delayed observations:}
    Consider a POMDP where the observation is delayed by $\delta$ time steps
    \citep{Bander1999}. For such a system the belief on $\delta$ step delayed
    state based on the $\delta$-step delayed observations and control, as well
    as the vector of last $\delta$ control actions is an information state.
    The structure of the optimal policy and the dynamic program derived in
    \cite{Bander1999} may be viewed as a special case of
    Theorem~\ref{thm:info-state}.

  \item \textsc{Machine maintenance:} Consider the following model for machine
    maintenance \citep{Eckles_1968}. A machine can be in one of $n$ ordered
    states where the first state is the best and the last state is the worst.
    The production cost increases with the state of the machine. The state
    evolves in a Markovian manner. At each time, an agent has the option to
    either run the machine or stop and inspect it for a cost. After
    inspection, the agent may either repair it (at a cost that depends on the
    state) or replace it (at a fixed cost). The objective is to identify a
    maintenance policy to minimize the cost of production, inspection, repair,
    and replacement.

    Let $\tau$ denote the time of last inspection and $\St_\tau$ denote the
    state of the machine after inspection, repair, or replacement. Then, it can
    be shown that $(\St_\tau, t-\tau)$ is an information state for the system.
    This is an instance of an incrementally expanding representation for a
    POMDP described in \cite{AM:rl-pomdp}.
\end{enumerate}

The above examples show that there are generic information states for certain
class of models (e.g., MDPs, MDPs with delays, POMDPs, POMDPs with delays)
as well as specific information states tuned to the model (e.g., even MDPs,
MDPs with irrelevant components, LQG models, machine repair).

\subsection{Discussion and related work} \label{discuss:info-state}

Although we are not aware of a previous result which formally defines an
information state and shows that an information state always implies a dynamic
programming decomposition (Theorem~\ref{thm:info-state}), yet the notion of
information state is not new and has always existed in the stochastic control
literature. Information state may be viewed as a generalization of the
traditional notion of state~\citep{Nerode:1958}, which is defined as a
statistic (i.e., a function of the observations) sufficient for input-output
mapping. In contrast, we define an information state as a statistic sufficient
for performance evaluation (and, therefore, for dynamic programming). Such a
definition is hinted in \cite{Witsenhausen:1976}. The notion of information
state is also related to sufficient statistics for optimal control defined in
\cite{Striebel:1965} for systems with state space models. 

As far as we are aware, the informal definition of information state was first
proposed by~\cite{Kwakernaak:1965} for adaptive control systems. Formal
definitions for linear control systems were given by~\cite{Bohlin:1970} for
discrete time systems and by~\cite{DavisVaraiya:1972} for continuous time
systems. \cite{Kumar_1986} define an information state as a compression of
past history which satisfies property (P2a) but do not formally show that
such an information state always leads to a dynamic programming decomposition.
A formal definition of information state appears in our previous work
\citep{MM:dec-control} where the result of Theorem~\ref{thm:info-state} is
asserted without proof. Properties of information states for 
multi-agent teams were asserted in \cite{Mahajan:phd}. \cite{Adlakha2012}
provide a definition which is stronger than our definition. 
They require that in a POMDP with unobserved state $\St_t \in
\StSp$, $\info_t(\his_t)$ should satisfy (P1) and (P2) as well be sufficient
to predict $\St_t$, i.e., for any Borel subset $\ALPHABET B$ of $\StSp$ and
any realization $\his_t$ of $\His_t$,
\[
  \PR(\St_t \in \ALPHABET B \mid \His_t = \his_t) = 
  \PR(\St_t \in \ALPHABET B \mid \Ais_t = \info_t(\his_t)).
\]
A similar definition is also used in \cite{FrancoisLavet2019}. We had
presented a definition similar to Definition~\ref{def:info_state} in the
preliminary version of this paper \citep{CDC}.

The notion of information state is also related to $\Gamma$-trace equivalence
for MDPs and POMDPs defined by \cite{CastroPanangadenPrecup_2009}. For MDPs.
$\Gamma$-trace equivalence takes a partition of the state space and returns a
finer partition such that for any choice of future actions any two states in
the same cell of the finer partition have the same distribution on future
states and rewards. \cite{CastroPanangadenPrecup_2009} show that recursive
applications of $\Gamma$-trace equivalence has a fixed point, which is
equivalent to bisimulation based partition \citep{Givan_2003} of the state
space of the MDP. Similar results were shown for MDPs in
\cite{FernsPanangadenPrecup_2004, FernsPanangadenPrecup_2011}.

\cite{CastroPanangadenPrecup_2009} extend the notion of trace equivalence for
MDPs to belief trajectory equivalence for POMDPs. In particular, two belief
states are said to be belief trajectory equivalent if for any choice of future
actions, they generate the same distribution on future observations and
rewards. Such belief trajectory equivalence is related to predictive state
representation (PSR) \citep{LittmanSutton_2002,
  SinghLittmanJongPardoeStone_2003, Izadi2003, James2004, RosencrantzGordonThrun_2004, WolfeJamesSingh_2005} and observable operator models (OOM) \citep{Jaeger2000,
Jaeger2006}, which are a compression of the past history which is
sufficient to predict the future observations (but not necessarily rewards).
Information state may be viewed as a ``Markovianized'' version of belief
trajectory equivalence and PSRs, which has the advantage that both (P1) and
(P2) are defined in terms of ``one-step'' equivalence while belief trajectory
equivalence and PSR are defined in terms of ``entire future trajectory''
equivalence. It should be noted that PSR and bisimulation based equivalences
are defined for infinite horizon models, while the information state is
defined for both finite and infinite horizon models (see
Sec.~\ref{sec:infinite}).

Another related notion is the notion of causal states (or
$\varepsilon$-machines) used in computational
mechanics~\citep{Crutchfield:1989,Crutchfield:2001}. and forecasting in
dynamical systems~\citep{Grassberger:1986, Grassberger:1988}. These
definitions are for uncontrolled Markov chains and the emphasis is on the
minimal state representation for time-invariant infinite-horizon systems.

\section{Approximate planning in partially observed systems}\label{sec:ais}

Our key insight is that information states provide a principled 
approach to approximate planning and learning in partially observed systems. 
To illustrate this, reconsider the machine maintenance example presented
earlier in Sec.~\ref{ex:info-state}.
Theorem~\ref{thm:info-state} implies that we can write a dynamic program for
that model using the information state $(\St_\tau, t - \tau)$, which takes
values in a countable set. This countable state dynamic program is
considerably simpler than the standard belief state dynamic program typically
used for that model. Moreover, it is possible to approximate the countable
state model by a finite-state model by truncating the state space, which
provides an approximate planning solution to the problem. Furthermore, the
information state $(\St_\tau, t-\tau)$ does not depend on the transition
probability of the state of the machine or the cost of inspection or repair.
Thus, if these model parameters were unknown, we can use a standard
reinforcement learning algorithm to find an optimal policy which maps
$(\St_\tau, t-\tau)$ to current action.

Given these benefits of a good information state, it is natural to consider a
data-driven approach to identify an information state. An information state
identified from data will not be exact and it is important to understand what
is the loss in performance when using an approximate information state. 
Theorem~\ref{thm:info-state} shows that a compression of the history which
satisfies properties (P1) and (P2) is sufficient to identify a dynamic
programming decomposition. Would a compression of history that approximately
satisfied properties (P1) and (P2) lead to an approximate dynamic program? In
this section, we show that the answer to this question is yes. First, we need
to precisely define what we mean by ``approximately satisfy properties (P1)
and (P2)''. For that matter, we need to fix a distance metric on probability
spaces. There are various metrics on probability space and it turns out that
the appropriate distance metric for our purposes is the 
integral probability metric (IPM) \citep{muller1997integral}.

\subsection{Integral probability metrics (IPM)}
\begin{definition}\label{def:ipm}
  Let $(\ALPHABET X, \mathcal G)$ be a measurable space and $\F$
  denote a class of uniformly bounded measurable functions on $(\ALPHABET X,
  \mathcal G)$. The integral probability metric (IPM) between two probability
  distributions $\mu, \nu \in \Delta(\ALPHABET X)$ with respect to the
  function class $\F$ is defined as
  \[
    d_{\F}(\mu, \nu) \coloneqq \sup_{f \in \F}\biggl |
    \int_{\ALPHABET{X}} fd\mu - \int_{\ALPHABET{X}} fd\nu \biggr|.
  \]
\end{definition}

In the literature, IPMs are also known as probability metrics with a
$\zeta$-structure; see e.g., \cite{Zolotarev1983,Rachev1991}. They are useful
to establish weak convergence of probability measures. Methods for estimating
IPM from samples are discussed in \cite{Sriperumbudur2012}.

\subsubsection*{Examples of integral probability metrics (IPMs)}
When $(\ALPHABET X, \mathcal G)$ is a metric space, then various commonly used
distance metrics on $(\ALPHABET X, \mathcal G)$ lead to specific instances of IPM
for a particular choice of function space $\F$. We provide some
examples below:
\begin{enumerate}
  \item \textsc{Total variation distance:} If $\F$ is chosen as $\{f :
    \lVert f \rVert_\infty \le 1\}$, then $d_{\F}$ is the total
    variation distance.\footnote{\label{fnt:TV}%
      In particular, if $\mu$ and $\nu$ are
      absolutely continuous with respect to some measure $\lambda$ and let $p
      = d\mu/d\lambda$ and $q = d\nu/d\lambda$, then
      \[
        \left| \int_{\ALPHABET X} f d\mu - \int_{\ALPHABET X} f d\nu \right|
        =
        \left| \int_{\ALPHABET X} f(x) p(x) \lambda(dx) 
             - \int_{\ALPHABET X} f(x) q(x) \lambda(dx)
        \right|
        \le 
        \| f \|_{\infty} \int_{\ALPHABET X} 
        \bigl| p(x) - q(x) \bigr| \lambda(dx).
      \]
      In this paper, we are defining total variation distance as
      $\int_{\ALPHABET X}| p(x) - q(x)| \lambda(dx)$. Typically, it is defined
      as half of that quantity. Note that it is possible to get a tighter bound
      than above where $\| f\|_\infty$ is replaced by $\tfrac12\Span(f)=
    \tfrac12(\max(f) - \min(f))$. }
    
  \item \textsc{Kolmogorov distance:} If $\ALPHABET X = \reals^m$ and $\mathfrak
    F$ is chosen as $\{ \IND_{(-\infty, t]} \colon t \in \reals^m \}$, then
    $d_{\F}$ is the Kolmogorov distance.

  \item \textsc{Kantorovich metric or Wasserstein distance:} Let $\lVert f
    \rVert_\Lip$ denote the Lipschitz semi-norm of a function. If $\F$
    is chosen as $\{ f : \lVert f \rVert_\Lip \le 1 \}$, then
    $d_{\F}$ is the Kantorovich metric. When $\ALPHABET X$ is
    separable, the Kantorovich metric is the dual representation of the
    Wasserstein distance via the Kantorovich-Rubinstein
    duality~\citep{Villani_2008}.

  \item \textsc{Bounded-Lipschitz metric:} If $\F$ is chosen as $\{f : \lVert f
    \rVert_\infty + \lVert f \rVert_\Lip \le 1\}$, then $d_{\F}$ is the
    bounded-Lipschitz (or Dudley) metric.

  \item \textsc{Maximum mean discrepancy (MMD):} Let $\mathcal H$ be a reproducing
    kernel Hilbert space (RKHS) of real valued functions on $\ALPHABET X$ 
    and let $\F = \{ f \in \mathcal H : \lVert f \rVert_{\mathcal H} \le 1
    \}$, then $d_{\F}$ is the maximum mean discrepancy\footnote{One of
      features of MMD is that the optimizing $f$ can be identified in closed
      form. In particular, if $k$ is the kernel of the RKHS, then (see
      \cite{Gretton2006, Sriperumbudur2012} for details)
      \begin{align*}
        d_\F(\mu, \nu) &= \biggl\| \int_{\ALPHABET X} k(\cdot, x) d\mu(x) 
        - \int_{\ALPHABET X} k(\cdot, x) d\nu(x) \biggr\|_{\mathcal H}
        \\
        &= \biggl[ 
          \int_{\ALPHABET X} \int_{\ALPHABET X} k(x,y) \mu(dx) \mu(dy)
          +                                                         
          \int_{\ALPHABET X} \int_{\ALPHABET X} k(x,y) \nu(dx) \nu(dy)
          - 2                                                       
          \int_{\ALPHABET X} \int_{\ALPHABET X} k(x,y) \mu(dx) \nu(dy)
        \biggr]^{1/2}.
      \end{align*}
    We use an MMD as a IPM in the PORL algorithms proposed in
  Sec.~\ref{sec:RL}, where we exploit this property.}
    \citep{Sriperumbudur2008}. The energy distance studied in statistics
    \citep{Szekely2004} is a special case of maximum mean discrepancy;
    see~\cite{Sejdinovic2013} for a discussion.

\end{enumerate}

We say that $\F$ is a closed set if it is closed under the topology of
pointwise convergence. We say that $\F$ is a convex set if $f_1, f_2 \in \F$
implies that for any $\lambda \in (0,1)$, $\lambda f_1 + (1 - \lambda)f_2 \in
\F$. Note that all the above function classes are convex and all except
Kolmogorov distance are closed. 

We now list some useful properties of IPMs, which immediately follow from
definition.
\begin{enumerate}
  \item Given a function class $\F$ and a function $f$ (not
    necessarily in $\F$), 
    \begin{equation} \label{eq:minkowski}
      \biggl| \int_{\ALPHABET X} f d\mu - \int_{\ALPHABET X} f d\nu \biggr|
      \le \Minkowski_{\F}(f) \cdot d_{\F}(\mu, \nu),
    \end{equation}
    where $\Minkowski_{\F}(f)$ is the Minkowski functional with respect to $\F$
    given by
    \begin{equation}
      \Minkowski_{\F}(f) \coloneqq \inf\{
      \Minkowski \in \reals_{> 0} : \Minkowski^{-1} f \in \F \}.
    \end{equation}

    For the total variation distance, 
    \( \bigl| \int_{\ALPHABET X} f d\mu - \int_{\ALPHABET X} f d\nu \bigr|
    \le \tfrac12 \Span(f) d_\F(\mu, \nu)\). Thus, for total variation, $\Minkowski_\F(f)
    = \tfrac12 \Span(f)$. For the Kantorovich metric, 
    \( \bigl| \int_{\ALPHABET X} f d\mu - \int_{\ALPHABET X} f d\nu \bigr|
    \le \|f\|_\Lip d_\F(\mu,\nu)\). Thus, for Kantorovich
    metric, $\Minkowski_{\F}(f) = \lVert f \rVert_\Lip$. For the maximum mean
    discrepancy, 
    \( \bigl| \int_{\ALPHABET X} f d\mu - \int_{\ALPHABET X} f d\nu \bigr|
    \le \| f\|_{\mathcal H} d_\F(\mu, \nu) \). Thus, for maximum mean
    discrepancy, $\Minkowski_\F(f) = \lVert f \rVert_{\mathcal H}$.

  \item Let $\ALPHABET X$ and $\ALPHABET Y$ be Banach spaces and let
    $\F_{\ALPHABET X}$ and $\F_{\ALPHABET Y}$ denote the function class for
    $d_{\F}$ with domain $\ALPHABET X$ and $\ALPHABET Y$, respectively. Then,
    for any $\ell \colon \ALPHABET X \to \ALPHABET Y$, any real-valued
    function $f \in \F_{\ALPHABET Y}$ and any measures $\mu$ and $\nu$ on
    $\Delta(\ALPHABET X)$, we have
    \begin{equation*}
      \biggl| \int_{\ALPHABET X} f(\ell(x)) \mu (dx) 
            - \int_{\ALPHABET X} f(\ell(x)) \nu (dx) \biggr|
      \le 
      \Minkowski_{\F_{\ALPHABET X}}(f \circ \ell) d_{\F_{\ALPHABET X}}(\mu, \nu).
    \end{equation*}
    We define the contraction factor of the function $\ell$ as
    \begin{equation}\label{eq:F-contraction}
      \Contraction_{\F_{\ALPHABET X}, \F_{\ALPHABET Y}}(\ell) =
      \sup_{f \in \F_{\ALPHABET Y}}  
      \Minkowski_{\F_{\ALPHABET X}}
      ( f \circ \ell).
    \end{equation}
    Therefore, we can say that for any $f \in \F_{\ALPHABET Y}$,
    \begin{equation}\label{eq:contraction}
      \biggl| \int_{\ALPHABET X} f(\ell(x)) \mu (dx) 
            - \int_{\ALPHABET X} f(\ell(x)) \nu (dx) \biggr|
      \le 
      \Contraction_{\F_{\ALPHABET X}, \F_{\ALPHABET Y}}(\ell) d_{\F_{\ALPHABET X}}(\mu, \nu).
    \end{equation}

    For the total variation distance, $\tfrac12 \Span( f \circ \ell) \le  \| f \circ
    \ell \|_{\infty} \le  \| f \|_{\infty} \le 1$. Thus,
    $\Contraction_\F(\ell) \le 1$. For the Kantorovich metric, $\| f \circ
    \ell \|_\Lip \le \| f \|_\Lip \| \ell \|_\Lip$ Thus, $\Contraction_\F(\ell) \le \|
    \ell \|_\Lip$.
\end{enumerate}

\subsection{Approximate information state (AIS) and approximate dynamic
programming}%\label{sec:ais}

Now we define a notion of \ac{AIS} as a compression of
the history of observations and actions which approximately satisfies
properties (P1) and (P2).

\begin{definition} \label{def:ais}
  Let $\{\AisSp_t\}_{t=1}^T$ be a pre-specified collection of Banach spaces, 
  $\F$ be a function class for \textup{IPMs}, and
  $\epsilonDelta$ be pre-specified positive real numbers. A collection $\{
  \ainfo_t \colon \HisSp_t \to \AisSp_t \}_{t=1}^T$ of history compression
  functions, along with approximate update kernels $\{\nextinfo_t \colon
  \AisSp_t \times \ActSp \to \Delta(\AisSp_{t+1})\}_{t=1}^T$ and reward
  approximation functions $\{\rewinfo_t \colon \AisSp_t \times \ActSp \to
  \reals\}_{t=1}^T$, is called an \emph{$\epsilonDelta$-\ac{AIS}
  generator} if the process $\{\Ais_t\}_{t=1}^T$, where
  $\Ais_t = \ainfo_t(\His_t)$, satisfies the following properties:
  \begin{description}
    \item[(AP1)] \textbf{\textup{Sufficient for approximate performance
      evaluation}}, i.e., for any time~$t$, any realization $\his_t$ of
      $\His_t$ and any choice $\act_t$ of $\Act_t$, we have
      \[
        \bigl\lvert \EXP[ \Rew_t \mid \His_t = \his_t, \Act_t = \act_t ] - 
        \rewinfo_t(\ainfo_t(\his_t), \act_t) \bigr\rvert 
        \le \varepsilon_t.
      \]
    \item[(AP2)] \textbf{\textup{Sufficient to predict itself approximately}}.
      i.e., for any time~$t$, any realization $\his_t$ of $\His_t$, any choice
      $\act_t$ of $\Act_t$, and for any Borel subset $\ALPHABET B$ of
      $\AisSp_{t+1}$, define
      \(
        \mu_t(\ALPHABET B) \coloneqq \PR(\Ais_{t+1} \in B \mid \His_t = \his_t, \Act_t = \act_t)
      \)
      and
      \(
        \nu_t(\ALPHABET B) \coloneqq \nextinfo_t(B \mid \ainfo_t(\his_t), \act_t);
      \)
      then, 
      \[
        d_\F( \mu_t, \nu_t) \le \delta_t.
      \]
  \end{description}
\end{definition}
We use the phrase ``$(\varepsilon, \delta)$-\acs{AIS}'' when $\varepsilon_t$ and
$\delta_t$ do not depend on time. 

Similar to Proposition~\ref{prop:alt-info-state}, we can provide an
alternative characterization of an \acs{AIS} where we replace (AP2) with the
following approximations of (P2a) and (P2b). 

\begin{description}
  \item[(AP2a)] \textbf{Evolves in a state-like manner}, i.e., there exist
    measurable update functions $\{\aupdate_t \colon \AisSp_t \times \ObSp
    \times \ActSp\}_{t=1}^T$ such that for any realization $\his_{t+1}$ of
    $\His_{t+1}$, we have
    \[
      \ainfo_{t+1}(\his_{t+1}) = \aupdate(\ainfo_t(\his_t), \ob_t, \act_t).
    \]
  \item[(AP2b)] \textbf{Is sufficient for predicting future observations
    approximately}, i.e., there exist measurable observation prediction kernels
    $\{\nextobs_t \colon \AisSp_t \times \ActSp \to \Delta(\ObSp)\}_{t=1}^T$
    such that for any time~$t$, any realization $\his_t$ of $\His_t$,
    any choice $\act_t$ of $\Act_t$, and for any Borel subset $\ALPHABET B$ of
    $\ObSp$ define,
    \(
      \mu^\ob_t(\ALPHABET B) \coloneqq \PR(\Ob_{t} \in \ALPHABET B \mid \His_t = \his_t, \allowbreak \Act_t = \act_t)
    \)
    and
    \(
      \nu^\ob_t(\ALPHABET B) = \nextobs_t(\ALPHABET B | \ainfo_t(\his_t), \act_t)
    \);
    then, 
    \[
      d_{\F}( \mu^\ob_t, \nu^\ob_t) \le
      \delta/\Contraction_{\F}(\aupdate_t),
    \]
    where $\Contraction_{\F}(\aupdate_t)$ is defined as
    $\sup_{\his_t \in \HisSp_t, \act_t \in \ActSp_t} 
    \Contraction_{\F}(\aupdate_t(\ainfo_t(\his_t), \cdot, \act_t))$. Note
    that for the total variation distance $\Contraction_{\F}(\aupdate_t) = 1$; for
    the Kantorovich distance $\Contraction_{\F}(\aupdate_t)$ is equal to the
    Lipschitz uniform bound on the Lipschitz constant of $\aupdate_t$ with
    respect to $\ob_t$. 
\end{description}

\begin{proposition}\label{prop:alt-ais}
  \textup{(AP2a)} and \textup{(AP2b)} imply \textup{(AP2)} holds with 
  transition kernels $\{\nextobs_t\}_{t=1}^T$ defined as follows: for any
  Borel subset $\ALPHABET B$ of $\AisSp$, 
  \[
    \nextinfo_t(\ALPHABET B \mid \ainfo_t(\his_t), \act_t) =
    \int_{\ObSp} 
    \IND_{\ALPHABET B}
    (\aupdate_t(\ainfo_t(\his_t), \ob_t, \act_t))
    \nextobs_t(d\ob_t | \ainfo_t(\his_t), \act_t ).
  \]
  Therefore, we can alternatively define an $\epsilonDelta$-\acs{AIS} generator
  as a tuple $\{(\ainfo_t, \rewinfo_t, \aupdate_t, \nextobs_t)\}_{t=1}^T$
  which satisfies \textup{(AP1)}, \textup{(AP2a)}, and \textup{(AP2b)}.
\end{proposition}
\begin{proof}
  Note that by the law of total probability, $\mu_t$ and $\nu_t$ defined in
  (AP2) are
  \begin{align*}
    \mu_t(\ALPHABET B) &= \int_{\ObSp} 
    \IND_{\ALPHABET B}
    (\aupdate_t(\ainfo_t(\his_t), \ob_t, \act_t))
    \mu^\ob_t(d\ob_t),
    \\
    \nu_t(\ALPHABET B) &= \int_{\ObSp} 
    \IND_{\ALPHABET B}
    (\aupdate_t(\ainfo_t(\his_t), \ob_t, \act_t))
    \nu^\ob_t(d\ob_t).
  \end{align*}
  Thus, for any function
  $f \colon \AisSp_{t+1} \to \reals$, 
  \begin{align*}
    \int_{\AisSp_{t+1}} f d\mu_t 
    &=
    \int_{\ObSp_t} f( \aupdate_t(\ainfo_t(\his_t), \ob_t, \act_t))
    \mu^\ob_t(d\ob_t),
    \\
    \int_{\AisSp_{t+1}} f d\nu_t 
    &=
    \int_{\ObSp_t} f( \aupdate_t(\ainfo_t(\his_t), \ob_t, \act_t))
    \nu^\ob_t(d\ob_t).
  \end{align*}
  The result then follows from~\eqref{eq:contraction}.
\end{proof}

Our main result is to establish that any \acs{AIS} gives rise to an approximate
dynamic program.
\begin{theorem}\label{thm:ais}
  Suppose $\{\ainfo_t, \nextinfo_t, \rewinfo_t\}_{t=1}^T$ is an
  $\epsilonDelta$-\acs{AIS} generator.
  Recursively define approximate action-value functions $\{\hat Q_t \colon
    \AisSp_t \times \ActSp \to \reals \}_{t=1}^T$ and value functions $\{\hat
    V_t \colon \AisSp_t \to \reals\}_{t=1}^T$ as follows:
   $\hat V_{T+1}(\ais_{T+1}) \coloneqq 0$ and for $t \in \{T, \dots,
  1\}$:
  \begin{subequations}\label{eq:DP-ais}
  \begin{align}
    \hat Q_t(\ais_t, \act_t) &\coloneqq \rewinfo_t(\ais_t, \act_t) 
    + \int_{\AisSp_{t+1}} \hat V_{t+1}(\ais_{t+1}) 
    \nextinfo_t(d \ais_{t+1} \mid \ais_t, \act_t),
    \\
    \hat V_t(\ais_t) &\coloneqq \max_{\act_t \in \ActSp} \hat Q_t(\ais_t, \act_t).
  \end{align}
  \end{subequations}
  Then, we have the following:
  \begin{enumerate}
    \item \textbf{\textup{Value function approximation:}} For any time~$t$,
      realization~$\his_t$ of $\His_t$, and choice $\act_t$ of $\Act_t$, we have
      \begin{equation}\label{eq:value-approx-finite}
        \lvert Q_t(\his_t, \act_t) - \hat Q_t(\ainfo_t(\his_t), \act_t)\rvert 
        \le \alpha_t
        \quad\text{and}\quad
        \lvert V_t(\his_t) - \hat V_t(\ainfo_t(\his_t)) \rvert 
        \le \alpha_t,
      \end{equation}
      where $\alpha_t$ satisfies the following recursion: $\alpha_{T+1} =
      0$ and for $t \in \{T, \dots, 1 \}$,
      \[
        \alpha_t = \varepsilon_t + \Minkowski_{\F}(\hat V_{t+1})
        \delta_{t} + \alpha_{t+1}.
      \]
      Therefore,
      \[
        \alpha_t = \varepsilon_t + \sum_{\tau=t+1}^{T}\big[  
        \Minkowski_\F(\hat V_{\tau}) \delta_{\tau-1} + \varepsilon_\tau \bigr].
      \]
    \item \textbf{\textup{Approximately optimal policy:}} Let $\hat \pol = (\hat \pol_1, \dots,
      \hat \pol_T)$, where $\hat \pol_t \colon \AisSp_t \to \Delta(\ActSp)$,
      be a stochastic policy that satisfies 
      \begin{equation}\label{eq:ais-opt}
        \Supp(\hat \pol(\ais_t)) \subseteq 
        \arg \max_{\act_t \in \ActSp} \hat Q_t(\ais_t, \act_t).
      \end{equation}
      Define policy $\pol = (\pol_1, \dots, \pol_T)$, where $\pol_t
      \colon \HisSp_t \to \Delta(\ActSp)$ by $\pol_t \coloneqq \hat \pol_t \circ
      \ainfo_t$. Then, for any time~$t$, realization~$\his_t$ of $\His_t$, and
      choice $\act_t$ of $\Act_t$, we
      have
      \begin{equation}\label{eq:policy-approx}
        \lvert Q_t(\his_t, \act_t) - Q^\pol_t(\his_t, \act_t)\rvert 
        \le 2\alpha_t
        \quad\text{and}\quad
        \lvert V_t(\his_t) - V^\pol_t(\his_t) \rvert 
        \le 2\alpha_t.
      \end{equation}
  \end{enumerate}
\end{theorem}
\begin{proof}
  We prove both parts by backward induction. We start with value function
  approximation. Eq.~\eqref{eq:value-approx-finite} holds at $T+1$ by definition.
  This forms the basis of induction. Assume that~\eqref{eq:value-approx-finite}
  holds at time~$t+1$ and consider the system at time~$t$. We have that
  \begin{align*}
    \bigl| Q_t(\his_t, \act_t) &-
    \hat Q_t(\ainfo_t(\his_t), \act_t) \bigr|
    \\
    &\stackrel{(a)}\le
     \bigl| \EXP[ R_t \mid \His_t = \his_t, \Act_t = \act_t ]
       - \rewinfo_t(\ainfo_t(\his_t), \act_t) \bigr|
    \\
    &\quad + 
    \EXP\bigl[ \bigl|  V_{t+1}(\His_{t+1}) - \hat V_{t+1}(\ainfo_{t+1}(\His_{t+1}))
      \bigr| \bigm| \His_t = \his_t, \Act_t = \act_t \bigr] 
    \\
    &\quad+ 
    \biggl| \EXP[ \hat V_{t+1}(\ainfo_{t+1}(\His_{t+1})) \mid \His_t = \his_t,
    \Act_t = \act_t ] - 
    \int_{\AisSp_{t+1}} \hat V_{t+1}(\ais_{t+1}) \nextinfo_t(d\ais_{t+1} \mid
    \ainfo_t(\his_t), \act_t) \biggr|
    \\
    &\stackrel{(b)}\le
    \varepsilon_t + \alpha_{t+1} + \Minkowski_{\F}(\hat V_{t+1})
    \delta_{t} = \alpha_t
  \end{align*}
  where $(a)$ follows from triangle inequality and $(b)$ follows from (AP1),
  the induction hypothesis, (AP2) and~\eqref{eq:minkowski}. This proves the
  first part of~\eqref{eq:value-approx-finite}. The second part follows from
  \[
    \bigl| V_t(\his_t) - \hat V_t(\ainfo_t(\his_t)) \bigr| 
    \stackrel{(a)}\le 
    \max_{\act_t \in \ActSp} 
    \bigl| Q_t(\his_t, \act_t) - \hat Q_t(\ainfo_t(\his_t), \act_t) \bigr|
    \le \alpha_t,
  \]
  where $(a)$ follows from the inequality $\max f(x) \le \max | f(x) - g(x) |
  +  \max g(x)$. 

  To prove the policy approximation, we first prove an intermediate result.
  For policy $\hat \pol$ recursively define $\{ \hat Q^{\hat \pol}_t \colon
  \AisSp \times \ActSp \to \reals \}_{t=1}^T$ and $\{\hat V^{\hat \pol}_t
  \colon \AisSp \to \reals \}_{t=1}^{T+1}$ as follows: $\hat V^{\hat
  \pol}_{T+1}(\ais_{T+1}) \coloneqq 0$ and for $t \in \{T, \dots, 1\}$:
  \begin{subequations}
  \begin{align}
    \hat Q^{\hat \pol}_t(\ais_t, \act_t) &\coloneqq 
    \rewinfo_t(\ais_t, \act_t) + 
    \int_{\AisSp_{t+1}} \hat V^{\hat \pol}_t(\ais_{t+1})
    \nextinfo_t(d\ais_{t+1} \mid \ais_t, \act_t) 
    \\
    \hat V^{\hat \pol}_t(\ais_t) &\coloneqq  \sum_{\act_t \in \ActSp}
    \hat \pol_t(\act_t \mid \ais_t).
    \hat Q^{\hat \pol}_t(\ais_t, \act_t).
  \end{align}
  \end{subequations}
  Note that~\eqref{eq:ais-opt} implies that 
  \begin{equation}\label{eq:ais-policy}
    \hat Q^{\hat \pol}_t(\ais_t, \act_t) = \hat Q_t(\ais_t, \act_t)
    \quad\text{and}\quad
    \hat V^{\hat \pol}_t(\ais_t) = \hat V_t(\ais_t).
  \end{equation}
  Now, we prove that
  \begin{equation}\label{eq:policy-approx-2}
    \lvert Q^\pol_t(\his_t, \act_t) -
    \hat Q^{\hat \pol}_t(\ainfo_t(\his_t), \act_t) \rvert 
      \le \alpha_t
      \quad\text{and}\quad
      \lvert 
      V^\pol_t(\his_t)
      -
      \hat V^{\hat \pol}_t(\ainfo_t(\his_t))
      \rvert 
      \le \alpha_t.
  \end{equation}
  We prove the result by backward induction. By construction,
  Eq.~\eqref{eq:policy-approx-2} holds at time~$T+1$. This forms the basis of
  induction. Assume that~\eqref{eq:policy-approx-2} holds at time~$t+1$ and
  consider the system at time~$t$. We have
  \begin{align*}
    \bigl| Q^\pol_t(\his_t, \act_t) &-
    \hat Q^{\hat \pol}_t(\ainfo_t(\his_t), \act_t) \bigr|
    \\
    &\stackrel{(a)}\le
     \bigl| \EXP[ R_t \mid \His_t = \his_t, \Act_t = \act_t ]
       - \rewinfo_t(\ainfo_t(\his_t), \act_t) \bigr|
    \\
    &\quad + 
    \EXP\bigl[ \bigl|  V^\pol_{t+1}(\His_{t+1}) - \hat V^{\hat
      \pol}_{t+1}(\ainfo_{t+1}(\His_{t+1}))
      \bigr| \bigm| \His_t = \his_t, \Act_t = \act_t \bigr] 
    \\
    &\quad+ 
    \biggl| \EXP[ \hat V^{\hat \pol}_{t+1}(\ainfo_{t+1}(\His_{t+1})) \mid \His_t = \his_t,
    \Act_t = \act_t ] - 
    \int_{\AisSp_{t+1}} \hat V^{\hat \pol}_{t+1}(\ais_{t+1}) \nextinfo_t(d\ais_{t+1} \mid
    \ainfo_t(\his_t), \act_t) \biggr|
    \\
    &\stackrel{(b)}\le
    \varepsilon_t + \alpha_{t+1} + \Minkowski_{\F}(\hat V_{t+1})
    \delta_{t} = \alpha_t
  \end{align*}
  where $(a)$ follows from triangle inequality and $(b)$ follows from (AP1),
  the induction hypothesis, (AP2) and~\eqref{eq:minkowski}. This proves the
  first part of~\eqref{eq:policy-approx-2}. The second part follows from the
  triangle inequality:
  \[
    \bigl| V^\pol_t(\his_t) - \hat V^{\hat \pol}_t(\ainfo_t(\his_t)) \bigr|
    \le \sum_{\act_t \in \ActSp} \hat \pol_t(\act_t | \ainfo_t(\his_t) )
    \bigl| Q^\pol(\his_t, \act_t) - \hat Q^{\hat \pol}_t(\ainfo_t(\his_t),
    \act_t) \bigr| \le \alpha_t.
  \]
  
  Now, to prove the policy approximation, we note that
  \[
    \bigl| Q_t(\his_t, \act_t) - Q^{\pol}_t(\his_t, \act_t) \bigr|
    \le 
    \bigl| Q_t(\his_t, \act_t) - \hat Q^{\hat \pol}_t(\ainfo_t(\his_t), \act_t) \bigr|
    + 
    \bigl| Q^\pol_t(\his_t, \act_t) - \hat Q^{\hat \pol}_t(\ainfo_t(\his_t), \act_t) \bigr|
    \le \alpha_t + \alpha_t,
  \]
  where the first inequality follows from the triangle inequality, the first
  part of  the second inequality follows from~\eqref{eq:value-approx-finite}
  and~\eqref{eq:ais-policy} and the second part follows
  from~\eqref{eq:policy-approx-2}. This proves the first part
  of~\eqref{eq:policy-approx}. The second part of~\eqref{eq:policy-approx}
  follows from the same argument. 
\end{proof}

An immediate implication of Theorems~\ref{thm:info-state} and~\ref{thm:ais} is
the following.
\begin{corollary}
  Let $\{\info_t\}_{t=1}^T$ be an information state generator and
  $\{(\ainfo_t, \nextinfo_t, \rewinfo_t)\}_{t=1}^T$ be an \acs{AIS} generator.
  Then, for any time~$t$, realization $\his_t$ of history $\His_t$, and choice
  $\act_t$ of action $\Act_t$, we have
  \[
    \bigl| \bar Q_t(\info_t(\his_t), \act_t) - \hat Q_t(\ainfo_t(\his_t), \act_t) \bigr| 
    \le \alpha_t
    \quad\text{and}\quad
    \bigl| \bar V_t(\info_t(\his_t)) - \hat V_t(\ainfo_t(\his_t)) \bigr| 
    \le \alpha_t,
  \]
  \blue{where $\bar Q_t$ and $\bar V_t$ are defined as in
  Theorem~\ref{thm:info-state}.}
\end{corollary}

\blue{
\begin{remark}
  It is possible to derive a tighter bound in Theorem~\ref{thm:ais} and show
  that
  \[
    \alpha_t = \varepsilon_t + \Delta^*_t(\hat V_{t+1}) + \alpha_{t+1}
  \]
  where
  \[
    \Delta^*_t(\hat V_{t+1}) = 
    \sup_{h_t, a_t} \biggl| \EXP[ \hat V_{t+1}(\ainfo_{t+1}(\His_{t+1})) \mid \His_t = \his_t,
    \Act_t = \act_t ] - 
    \int_{\AisSp_{t+1}} \hat V_{t+1}(\ais_{t+1}) \nextinfo_t(d\ais_{t+1} \mid
    \ainfo_t(\his_t), \act_t) \biggr|
  \]
  The bound presented in Theorem~\ref{thm:ais} can be then thought of as an
  upper bound on $\Delta^*_t(\hat V_{t+1}) \le \rho_\F(\hat V_{t+1})\delta$ using~\eqref{eq:minkowski}.
\end{remark}
}

\blue{
\begin{remark}
  In part~1 of Theorem~\ref{thm:ais}, it is possible to derive an alternative
  bound 
  \[
    |Q_t(\his_t, \act_t) - \hat Q_t(\ainfo_t(\his_t), \act_t)| \le \alpha'_t
    \quad\text{and}\quad
    |V_t(\his_t) - \hat V_t(\ainfo_t(\his_t))| \le \alpha'_t
  \]
  where $\alpha'_t$ satisfies the recursion: $\alpha'_{T+1} = 0$ and for $t
  \in \{T, \dots, 1\}$, 
  \[
    \alpha'_t = \varepsilon_t + \rho_\F(V_{t+1})\delta_t + \alpha'_{t+1}.
  \]
  This is because while using the triangle inequality in step $(a)$ in the proof of Theorem~\ref{thm:ais}, we could have 
alternatively added and subtracted the term $\EXP[ V^{\pol}_{t+1}(\His_{t+1}) \mid \His_t = \his_t,
    \Act_t = \act_t ]$ instead of $\EXP[ \hat V^{\hat \pol}_{t+1}(\ainfo_{t+1}(\His_{t+1})) \mid \His_t = \his_t,
    \Act_t = \act_t ]$. Using this bound, we can also derive an alternative
    bound for part~2 of the Theorem and show that
    \[
      \lvert Q_t(\his_t, \act_t) - Q^\pol_t(\his_t, \act_t)\rvert 
      \le \alpha_t + \alpha'_t
      \quad\text{and}\quad
      \lvert V_t(\his_t) - V^\pol_t(\his_t) \rvert 
      \le \alpha_t + \alpha'_t.
    \]
\end{remark}
}

\subsection{Examples of approximate information states}
\label{ex:ais}

We now present various examples of information state and show that many
existing results in the literature may be viewed as a special case of
Theorem~\ref{thm:ais}. \blue{Some of these examples are for infinite horizon
discounted reward version of Theorem~\ref{thm:ais} (with discount factor
$\discount \in (0,1)$), which we prove later in Theorem~\ref{thm:inf-ais}.}

\begin{enumerate}
  \item \textsc{Model approximation in MDPs:} Consider an
    MDP with state space $\StSp$, action space $\ActSp$, transition kernel
    $P_t \colon \StSp \times \ActSp \to \Delta(\StSp)$, and per-step reward
    $r_t \colon \StSp \times \ActSp \to \reals$. Consider an approximate model
    defined on the same state and action spaces with transition kernel $\hat
    P_t \colon \StSp \times \ActSp \to \Delta(\StSp)$ and per-step reward
    $\hat r_t \colon \StSp \times \ActSp \to \reals$. Define
    $\ainfo_t(\St_{1:t}, \Act_{1:t-1}) = \St_t$. Then $\{ (\ainfo_t, \hat
    P_t, \hat r_t)\}_{t=1}^T$ is an \acs{AIS} with
    \[
      \varepsilon_t \coloneqq \sup_{\st \in \StSp, \act \in \ActSp}
      \bigl| r_t(\st, \act) - \hat r_t(\st, \act) \bigr|
      \quad\text{and}\quad
      \delta_t = \sup_{\st \in \StSp, \act \in \ActSp}
      d_{\F}\bigl( P_t(\cdot | \st, \act), \hat P_t (\cdot | \st, \act) \bigr).
    \]
    A result similar in spirit to Theorem~\ref{thm:ais} for this setup for
    general $d_{\F}$ is given in Theorem~4.2 of \cite{Muller1997}. When $d_\F$
    is the Kantorovich metric, a bound for model approximation for infinite
    horizon setup is provided in Theorem~2 of \cite{Asadi_2018}. This is
    similar to our result generalization of Theorem~\ref{thm:ais} to infinite
    horizon, which is given in Theorem~\ref{thm:inf-ais}; 
    a bound on $\Minkowski_\F(\hat V)$ in this case can be 
    obtained using results of \cite{Hinderer2005, Rachelson2010}. 

  \item \textsc{State abstraction in MDPs:} Consider an MDP with state space
    $\StSp$, action space $\ActSp$, transition kernel $P_t \colon \StSp \times
    \ActSp \to \Delta(\StSp)$, and per-step reward $r_t \colon \StSp \times
    \ActSp \to \reals$. Consider an abstract model defined over a
    state space $\hat \StSp$ (which is ``smaller'' than $\StSp$) and the same
    action space with transition kernel $\hat P_t \colon \hat \StSp \times
    \ActSp \to \Delta(\hat \StSp)$ and per-step reward $\hat r_t \colon \hat
    \StSp \times \ActSp \to \reals$. Suppose there is an abstraction function
    $q \colon \StSp \to \hat \StSp$ and, in state $\St \in \StSp$, we
    choose an action based on $q(\StSp)$. For such a model, define
    $\ainfo_t(\St_{1:t}, \Act_{1:t-1}) = q(\St_t)$. Then
    $\{ (\ainfo_t, \hat P_t, \hat r_t)\}_{t=1}^T$ is an \acs{AIS} with
    \[
      \varepsilon_t \coloneqq \sup_{\st \in \StSp, \act \in \ActSp}
      \bigl| r_t(\st, \act) - \hat r_t(q(\st), \act) \bigr|
      \quad\text{and}\quad
      \delta_t \coloneqq \sup_{\st \in \StSp, \act \in \ActSp}
      d_{\F}\bigl(\mu_t(\cdot | \st, \act), \hat P_t(\cdot | q(\st), \act)\bigr),
    \]
    where for any Borel subset $\ALPHABET B$ of $\hat \StSp$, 
    \(
      \mu_t(B | \st, \act) \coloneqq P_t( q^{-1}(\ALPHABET B) |  \st, \act)
    \).

    There is a rich literature on state abstraction starting with 
    \cite{Bertsekas_1975} and \cite{Whitt_1978}, but the error bounds in those
    papers are of a different nature. There are some recent papers which
    derive error bounds similar to Theorem~\ref{thm:ais} for the infinite
    horizon setup with state abstraction. We generalize Theorem~\ref{thm:ais}
    to infinite horizon later in Theorem~\ref{thm:inf-ais}. 

    When $d_{\F}$ is the Kantorovich metric, a bound on $\Minkowski_\F(\hat V)
    = \| \hat V \|_{\Lip}$ can be
    obtained using results of \cite{Hinderer2005, Rachelson2010}. Substituting
    this bound in Theorem~\ref{thm:inf-ais} gives us the following bound on the policy
    approximation error by using \acs{AIS}. 
    \[
        \bigl| V(\st) - V^\pol(s) \bigr| \le
        \frac{2\varepsilon}{(1-\discount)} + \frac{2\discount \delta
        \| \hat V \|_\Lip}{(1-\discount)}.
    \]
    Similar bound has been obtained in Theorem~5 of \cite{DeepMDP}. A detailed
    comparison with this model is presented in Appendix~\ref{app:LipMDP}.

    When $d_{\F}$ is the total variation distance, a bound on $d_\F(\hat V)$
    is given by $\Span(r)/(1-\discount)$. Substituting this in
    Theorem~\ref{thm:inf-ais}, we get that 
    \[
      | V(s) - V^\pol(s) | \le \frac{2\varepsilon}{(1-\discount)} 
      + \frac{\discount \delta \Span(r) }{(1-\discount)^2}.
    \]
    A $\mathcal O(1/(1-\discount)^3)$ bound on the policy approximation error in this
    setup was obtained in Lemma~2 and Theorem~2 of \cite{Abel2016}. \textbf{Directly
      using the \acs{AIS} bound of Theorems~\ref{thm:ais} and~\ref{thm:inf-ais} gives
      a factor of $1/(1-\discount)$ improvement in the error bound of
    \cite{Abel2016}.} See
    Appendix~\ref{app:Abel} for a detailed comparison.

  \item \textsc{Belief approximation in POMDPs:} Consider a POMDP with state space
    $\StSp$, action space $\ActSp$, observation space $\ObSp$, and a per-step
    reward function $r_t \colon \StSp \times \ActSp \to \reals$. Let
    $b_t(\cdot | \His_t) \in \Delta(\StSp)$ denote the belief of the current
    state given the history, i.e., 
    \(
      b_t(\st | \His_t) = \PR(\St_t = \st \mid \His_t)
    \).
    Suppose there are history compression functions $\{\phi_t \colon
    \HisSp_t \to \Phi_t \}_{t=1}^T$ (where $\Phi_t$ is some arbitrary space)
    along with
    belief approximation functions $\{\hat b_t \colon \Phi_t \to
    \Delta(\StSp)\}_{t=1}^T$, such that for any time~$t$ and any realization
    $\his_t$ of $\His_t$, we have
    \[
      \| \hat b_t(\cdot \mid \phi_t(\his_t)) - b_t(\cdot \mid \his_t) \|_1 \le
      \varepsilon.
    \]
    Such a $\{(\phi_t, \hat b_t)\}_{t=1}^T$ was called an
    $\varepsilon$-sufficient statistic in \cite{FrancoisLavet2019}. An example
    of $\varepsilon$-sufficient statistic is belief quantization, where the
    belief is quantized to the nearest point in the \emph{type lattice}
    (here $m = |\StSp|$)
    \[
      Q_n \coloneqq \bigl\{ (p_1, \dots, p_m) \in \Delta(\StSp) : 
      n p_i  \in \integers_{\ge 0} \bigr\}.
    \]
    An efficient algorithm to find the nearest point in $Q_n$ for any given
    belief $b_t \in \Delta(\StSp)$ is presented in \cite{Reznik2011}. Under
    such a quantization, the maximum $\ell_1$ distance between a belief vector
    and its quantized value is given by $2\lfloor m/2 \rfloor \lceil m/2
    \rceil/ mn \approx m/2n$ (see Proposition~2 of \cite{Reznik2011}). Thus,
    by taking $n > m/2\varepsilon$, we get an $\varepsilon$-sufficient
    statistic.

    \cite{FrancoisLavet2019} showed
    that the bias of using the optimal policy based on $\hat b_t(\his_t)$ in the
    original model is $2\varepsilon \|r\|_{\infty}/(1-\discount)^3$. This result
    uses the same proof argument as \cite{Abel2016} discussed in the
    previous bullet point, which is not tight. By metricizing the belief space
    using total variation distance and using the bounded-Lipschitz metric on
    the space of probability measures on beliefs, we can show that an
    $\varepsilon$-sufficient statistic induces a $(\varepsilon \Span(r),
    3\varepsilon)$-AIS. When $d_\F$ is the bounded-Lipschitz metric, a bound
    on $\Minkowski_\F(\hat V)$ is given by $2\|r\|_\infty/(1-\discount)$.
    Substituting this in Theorem~\ref{thm:inf-ais}, we get that
    \[
      | V(s) - V^\pol(s) | \le 
      \frac{2 \varepsilon \| r \|_{\infty} }{(1-\discount)}
      + 
      \frac{6 \discount \varepsilon \| r\|_{\infty}}{(1-\discount)^2}.
    \]
    Thus, \textbf{directly using the AIS bound of Theorems~\ref{thm:ais}
      and~\ref{thm:inf-ais} gives a factor of $1/(1-\discount)$ improvement
    in the error bound of \cite{FrancoisLavet2019}.}
    See Appendix~\ref{app:FrancoisLavet} for details.

    In a slightly different vein, belief quantization in POMDPs with finite or 
    Borel valued unobserved state was investigated in~\cite{saldi2018finite},
    who showed that under appropriate technical conditions the value
    function and optimal policies for the quantized model converge to the value
    function and optimal policy of the true model. However~\cite{saldi2018finite}
    did not provide approximation error for a fixed quantization level.

\end{enumerate}

{\color{black}
\subsection{Approximate policy evaluation}\label{sec:ais-poleval}
In some settings, we are interested in comparing the performance of an
arbitrary policy in an approximate model with its performance in the real
model. The bounds of Theorem~\ref{thm:ais} can be adapted to such a setting as
well. 

\begin{theorem}\label{thm:ais-poleval-fin}
  Suppose $\{\ainfo_t, \nextinfo_t, \rewinfo_t\}_{t=1}^T$ is an
  $\epsilonDelta$-\acs{AIS} generator. Let $\hat \pol^{\#} = (\hat \pol^{\#}_1, \dots,
      \hat \pol^{\#}_T)$, where $\hat \pol^{\#}_t \colon \AisSp_t \to \Delta(\ActSp)$,
      be an arbitrary stochastic policy. Recursively define approximate policy action-value 
      functions $\{\hat Q^{\hat \pol^{\#}}_t \colon
    \AisSp_t \times \ActSp \to \reals \}_{t=1}^T$ and value functions $\{\hat
    V^{\hat \pol^{\#}}_t \colon \AisSp_t \to \reals\}_{t=1}^T$ as follows:
   $\hat V^{\hat \pol^{\#}}_{T+1}(\ais_{T+1}) \coloneqq 0$ and for $t \in \{T, \dots,
  1\}$:
  \begin{subequations}\label{eq:DP-ais-poleval}
  \begin{align}
    \hat Q^{\hat \pol^{\#}}_t(\ais_t, \act_t) &\coloneqq 
    \rewinfo_t(\ais_t, \act_t) + 
    \int_{\AisSp_{t+1}} \hat V^{\hat \pol^{\#}}_t(\ais_{t+1})
    \nextinfo_t(d\ais_{t+1} \mid \ais_t, \act_t) 
    \\
    \hat V^{\hat \pol^{\#}}_t(\ais_t) &\coloneqq  \sum_{\act_t \in \ActSp}
    \hat \pol^{\#}_t(\act_t \mid \ais_t).
    \hat Q^{\hat \pol^{\#}}_t(\ais_t, \act_t).
  \end{align}
  \end{subequations}
  Define policy $\pol^{\#} = (\pol^{\#}_1, \dots, \pol^{\#}_T)$, where
  $\pol^{\#}_t
  \colon \HisSp_t \to \Delta(\ActSp)$ by $\pol^{\#}_t \coloneqq \hat \pol^{\#}_t \circ
      \ainfo_t$. Then, for any time~$t$, realization~$\his_t$ of $\His_t$, and
      choice $\act_t$ of $\Act_t$, we have:
  \begin{equation}\label{eq:policy-approx-2-poleval}
    \lvert Q^{\pol^{\#}}_t(\his_t, \act_t) -
    \hat Q^{\hat \pol^{\#}}_t(\ainfo_t(\his_t), \act_t) \rvert 
    \le \alpha^{\#}_t
      \quad\text{and}\quad
      \lvert 
      V^{\pol^{\#}}_t(\his_t)
      -
      \hat V^{\hat \pol^{\#}}_t(\ainfo_t(\his_t))
      \rvert 
      \le \alpha^{\#}_t,
  \end{equation}
  where $\alpha^{\#}_t$ satisfies the following recursion: $\alpha^{\#}_{T+1} =
      0$ and for $t \in \{T, \dots, 1 \}$,
      \[
        \alpha^{\#}_t = \varepsilon_t + \Minkowski_{\F}(\hat V^{\hat \pol^{\#}}_{t+1})
        \delta_{t} + \alpha^{\#}_{t+1}.
      \]
      Therefore,
      \[
        \alpha^{\#}_t = \varepsilon_t + \sum_{\tau=t+1}^{T}\big[  
        \Minkowski_\F(\hat V^{\hat \pol^{\#}}_{\tau}) \delta_{\tau-1} + \varepsilon_\tau \bigr].
      \]
\end{theorem}
\begin{proof}
  The proof proceeds by backward induction along the same lines as the proof
  of Theorem~\ref{thm:ais}.
  By construction,
  Eq.~\eqref{eq:policy-approx-2-poleval} holds at time~$T+1$. This forms the basis of
  induction. Assume that~\eqref{eq:policy-approx-2-poleval} holds at time~$t+1$ and
  consider the system at time~$t$. We have
  \begin{align*}
    \bigl| Q^{\pol^{\#}}_t(\his_t, \act_t) &-
    \hat Q^{\hat \pol^{\#}}_t(\ainfo_t(\his_t), \act_t) \bigr|
    \\
    &\stackrel{(a)}\le
     \bigl| \EXP[ R_t \mid \His_t = \his_t, \Act_t = \act_t ]
       - \rewinfo_t(\ainfo_t(\his_t), \act_t) \bigr|
    \\
    &\quad + 
    \EXP\bigl[ \bigl|  V^{\pol^{\#}}_{t+1}(\His_{t+1}) - \hat V^{\hat
      \pol^{\#}}_{t+1}(\ainfo_{t+1}(\His_{t+1}))
      \bigr| \bigm| \His_t = \his_t, \Act_t = \act_t \bigr] 
    \\
    &\quad+ 
    \biggl| \EXP[ \hat V^{\hat \pol^{\#}}_{t+1}(\ainfo_{t+1}(\His_{t+1})) \mid \His_t = \his_t,
    \Act_t = \act_t ] - 
    \int_{\AisSp_{t+1}} \hat V^{\hat \pol^{\#}}_{t+1}(\ais_{t+1}) \nextinfo_t(d\ais_{t+1} \mid
    \ainfo_t(\his_t), \act_t) \biggr|
    \\
    &\stackrel{(b)}\le
    \varepsilon_t + \alpha^{\#}_{t+1} + \Minkowski_{\F}(\hat V^{\hat \pol^{\#}}_{t+1})
    \delta_{t} = \alpha^{\#}_t
  \end{align*}
  where $(a)$ follows from triangle inequality and $(b)$ follows from (AP1),
  the induction hypothesis, (AP2) and~\eqref{eq:minkowski}. This proves the
  first part of~\eqref{eq:policy-approx-2-poleval}. The second part follows from the
  the fact that $\pol^{\#}(\act_t | \his_t) = \hat \pol^{\#}(\act_t |
  \ainfo_t(\his_t))$ and the triangle inequality:
  \[
    \bigl| V^{\pol^{\#}}_t(\his_t) - \hat V^{\hat \pol^{\#}}_t(\ainfo_t(\his_t)) \bigr|
    \le \sum_{\act_t \in \ActSp} \hat \pol^{\#}_t(\act_t | \ainfo_t(\his_t) )
    \bigl| Q^{\pol^{\#}}(\his_t, \act_t) - \hat Q^{\hat \pol^{\#}}_t(\ainfo_t(\his_t),
    \act_t) \bigr| \le \alpha^{\#}_t.
  \]
\end{proof}
}

\subsection{Stochastic AIS} \label{sec:stochais}

We have so far assumed that the history compression functions $\ainfo_t
\colon \HisSp_t \to \AisSp_t$ are deterministic functions. 
When learning a
discrete-valued \acs{AIS} from data, it is helpful to consider stochastic
mappings of history, so that quality of the mapping may be improved via
stochastic gradient descent. In general, the definition of deterministic
\acs{AIS} also covers the case of stochastic \acs{AIS} because a stochastic
function from $\HisSp_t$ to $\AisSp_t$ may be viewed as a deterministic
function from $\HisSp_t$ to $\Delta(\AisSp_t)$. However, a more explicit
characterization is also possible, which we present next.

\begin{definition} \label{def:stoc-ais}
  Let $\{\AisSp_t\}_{t=1}^T$ be a pre-specified collection of Banach spaces, 
  $\F$ be a function class for \textup{IPMs}, and
  $\epsilonDelta$ be pre-specified positive real numbers. A collection $\{
  \ainfo^s_t \colon \HisSp_t \to \Delta(\AisSp_t) \}_{t=1}^T$ of stochastic history compression
  functions, along with approximate update kernels $\{\nextinfo_t \colon
  \AisSp_t \times \ActSp \to \Delta(\AisSp_{t+1})\}_{t=1}^T$ and reward
  approximation functions $\{\rewinfo_t \colon \AisSp_t \times \ActSp \to
  \reals\}_{t=1}^T$, is called an \emph{$\epsilonDelta$-stochastic \ac{AIS}
  generator} if the process $\{\Ais_t\}_{t=1}^T$, where
  $\Ais_t = \ainfo_t(\His_t)$, satisfies the following properties:
  \begin{description}
    \item[(AP1)] \textbf{\textup{Sufficient for approximate performance
      evaluation}}, i.e., for any time~$t$, any realization $\his_t$ of
      $\His_t$ and any choice $\act_t$ of $\Act_t$, we have
      \[
        \bigl\lvert \EXP[ \Rew_t \mid \His_t = \his_t, \Act_t = \act_t ] - 
         \EXPsAIS[\rewinfo_t(\Ais_t, \act_t)] \bigr\rvert 
        \le \varepsilon_t.
      \]
    \item[(AP2)] \textbf{\textup{Sufficient to predict itself approximately}}.
      i.e., for any time~$t$, any realization $\his_t$ of $\His_t$, any choice
      $\act_t$ of $\Act_t$, and for any Borel subset $\ALPHABET B$ of
      $\AisSp_{t+1}$, define
      \(
        \mu_t(\ALPHABET B) \coloneqq \PR(\Ais_{t+1} \in B \mid \His_t = \his_t, \Act_t = \act_t)
      \)
      and
      \(
        \nu_t(B) \coloneqq \EXPsAIS[\nextinfo_t(B |
        \Ais_t, \act_t)];
      \)
      then, 
      \[
        d_\F( \mu_t, \nu_t) \le \delta_t.
      \]
  \end{description}
\end{definition}

Similar to Theorem~\ref{thm:ais}, we then have the following result.
\begin{theorem}\label{thm:stoc-ais}
  Given a stochastic \acs{AIS} generator $\{\ainfo^s_t, \nextinfo_t,
  \rewinfo_t\}_{t=1}^T$, define value functions $\{ \hat V_t \colon \AisSp_t
  \to \reals\}_{t=1}^T$ and action-value functions $\{ \hat Q_t \colon \AisSp_t
\times \ActSp \to \reals \}_{t=1}^T$ as in Theorem~\ref{thm:ais}. 
  Then, we have the following:
  \begin{enumerate}
    \item \textbf{\textup{Value function approximation:}} For any time~$t$,
      realization~$\his_t$ of $\His_t$, and choice $\act_t$ of $\Act_t$ we have
      \begin{equation}\label{eq:stoch-value-approx}
        \lvert Q_t(\his_t, \act_t) - \EXPsAIS[ \hat Q_t(\Ais_t, \act_t)] \rvert 
        \le \alpha_t
        \quad\text{and}\quad
        \lvert V_t(\his_t) - \EXPsAIS[ \hat V_t(\Ais_t) ] \rvert 
        \le \alpha_t,
      \end{equation}
      where $\alpha_t$ is defined as in Theorem~\ref{thm:ais}.

    \item \textbf{\textup{Approximately optimal policy:}} Let $\hat \pol = (\hat \pol_1, \dots,
      \hat \pol_T)$, where $\hat \pol_t \colon \AisSp_t \to \Delta(\ActSp)$,
      be a stochastic policy that satisfies 
      \begin{equation}\label{eq:stoch-ais-opt}
        \Supp(\hat \pol(\ais_t)) \subseteq 
        \arg \max_{\act_t \in \ActSp} \hat Q_t(\ais_t, \act_t).
      \end{equation}
      Define policy $\pol = (\pol_1, \dots, \pol_T)$, where $\pol_t
      \colon \HisSp_t \to \Delta(\ActSp)$ by $\pol_t(\his_t) = \EXPsAIS[ \hat
      \pol_t(\Ais_t) ]$. Then, for any time~$t$, realization~$\his_t$ of $\His_t$,
      and choice $\act_t$ of $\Act_t$, we
      have
      \begin{equation}\label{eq:stoc-policy-approx}
        \lvert Q_t(\his_t, \act_t) - Q^\pol_t(\his_t, \act_t)\rvert 
        \le 2\alpha_t
        \quad\text{and}\quad
        \lvert V_t(\his_t) - V^\pol_t(\his_t) \rvert 
        \le 2\alpha_t.
      \end{equation}
  \end{enumerate}
\end{theorem}
\begin{proof}
  The proof is almost the same as the proof of Theorem~\ref{thm:ais}. The main
  difference is that for the value and action-value functions of the
  stochastic approximation state, we take an additional expectation over the
  realization of the stochastic AIS. We only show the details of the proof of
  the first part of the result (value approximation). The second part (policy
  approximation) follows along similar lines.

  Eq.~\eqref{eq:stoch-value-approx} holds at $T+1$ by definition.
  This forms the basis of induction. Assume that~\eqref{eq:stoch-value-approx}
  holds at time~$t+1$ and consider the system at time~$t$. We have that
  \begin{align*}
    \bigl| Q_t&(\his_t, \act_t) -
    \EXPsAIS[ \hat Q_t(\Ais_t, \act_t)] \bigr|
    \\
    &\stackrel{(a)}\le
     \bigl| \EXP[ R_t \mid \His_t = \his_t, \Act_t = \act_t ]
     - \EXPsAIS[ \rewinfo_t(\Ais_t, \act_t) ] \bigr|
    \\
    &\quad + 
    \EXP\bigl[ \bigl|  V_{t+1}(\His_{t+1}) - 
      \EXPsAISt[ \hat V_{t+1}(\Ais_{t+1}) ]
      \bigr| \bigm| \His_t = \his_t, \Act_t = \act_t \bigr] 
    \\
    &\quad+ 
    \biggl| \EXP[ \hat V_{t+1}(\ainfo_{t+1}(\His_{t+1})) \mid \His_t = \his_t,
    \Act_t = \act_t ] - 
    \EXPsAIS \biggl[
      \int_{\AisSp_t} \hat V_{t+1}(\ais_{t+1}) \nextinfo_t(d\ais_{t+1} \mid
    \Ais_t, \act_t) \biggr] \biggr|
    \\
    &\stackrel{(b)}\le
    \varepsilon_t + \alpha_{t+1} + \Minkowski_{\F}(\hat V_{t+1})
    \delta_{t} = \alpha_t
  \end{align*}
  where $(a)$ follows from triangle inequality and $(b)$ follows from (AP1),
  the induction hypothesis, (AP2) and~\eqref{eq:minkowski}. This proves the
  first part of~\eqref{eq:stoch-value-approx}. The second part follows from
  \[
    \bigl| V_t(\his_t) - \hat V_t(\ainfo^s_t(\his_t)) \bigr| 
    \stackrel{(a)}\le 
    \max_{\act_t \in \ActSp} 
    \bigl| Q_t(\his_t, \act_t) - \hat Q_t(\ainfo^s_t(\his_t), \act_t) \bigr|
    \le \alpha_t,
  \]
  where $(a)$ follows from the inequality $\max f(x) \le \max | f(x) - g(x) |
  +  \max g(x)$. This completes the proof of value approximation. The proof of
  policy approximation is similar to that of Theorem~\ref{thm:ais} adapted in
  the same manner as above.
\end{proof}

\subsection{AIS with action compression} \label{sec:actais}

So far we have assumed that the action space for the AIS is the same as the
action space for the original model. In some instances, for example, for
continuous or large action spaces, it may be desirable to quantize or compress
the actions as well. In this section, we generalize the notion of AIS to
account for action compression. 

\begin{definition} \label{def:ais-ac}
  As in the definition of \acs{AIS}, suppose $\{\AisSp_t\}_{t=1}^T$ are
  pre-specified collection of Banach spaces, $\F$ be a function class for
  \textup{IPMs}, and $\epsilonDelta$ be pre-specified positive real numbers. In
  addition, suppose we have a subset $\aActSp \subset \ActSp$ of quantized
  actions. Then, a collection $\{\ainfo_t \colon \HisSp_t \to \AisSp_t
  \}_{t=1}^T$ of history compression functions, along with action quantization
  function $\aquant \colon \ActSp \to \aActSp$, 
  approximate update kernels $\{\nextinfo_t \colon
  \AisSp_t \times \aActSp \to \Delta(\AisSp_{t+1})\}_{t=1}^T$ and reward
  approximation functions $\{\rewinfo_t \colon \AisSp_t \times \aActSp \to
  \reals\}_{t=1}^T$, is called an
  \emph{$\epsilonDelta$-action-quantized \acs{AIS}
  generator} if the process $\{\Ais_t\}_{t=1}^T$, where
  $\Ais_t = \ainfo_t(\His_t)$, satisfies the following properties:
  \begin{description}
    \item[(AQ1)] \textbf{\textup{Sufficient for approximate performance
      evaluation}}, i.e., for any time~$t$, any realization $\his_t$ of
      $\His_t$ and any choice $\act_t$ of $\Act_t$, we have
      \[
        \bigl\lvert \EXP[ \Rew_t \mid \His_t = \his_t, \Act_t = \act_t ] - 
        \rewinfo_t(\ainfo_t(\his_t), \aquant(\act_t)) \bigr\rvert 
        \le \varepsilon_t.
      \]
    \item[(AQ2)] \textbf{\textup{Sufficient to predict itself approximately}}.
      i.e., for any time~$t$, any realization $\his_t$ of $\His_t$, any choice
      $\act_t$ of $\Act_t$, and for any Borel subset $\ALPHABET B$ of
      $\AisSp_{t+1}$, define
      \(
        \mu_t(\ALPHABET B) \coloneqq \PR(\Ais_{t+1} \in B \mid \His_t = \his_t, \Act_t = \act_t)
      \)
      and
      \(
        \nu_t(\ALPHABET B) \coloneqq \nextinfo_t(B \mid \ainfo_t(\his_t),
        \aquant(\act_t));
      \)
      then, 
      \[
        d_\F( \mu_t, \nu_t) \le \delta_t.
      \]
  \end{description}
\end{definition}

Similar to Theorem~\ref{thm:ais}, we show that an action-quantized \acs{AIS}
can be used to determine an approximately optimal policy.

\begin{theorem}\label{thm:ais-ac}
  Suppose $\{\ainfo_t, \aquant, \nextinfo_t, \rewinfo_t\}_{t=1}^T$ is an
  action-quantized \acs{AIS} generator.
  Recursively define approximate action-value functions $\{\hat Q_t \colon
    \AisSp_t \times \aActSp \to \reals \}_{t=1}^T$ and value functions $\{\hat
    V_t \colon \AisSp_t \to \reals\}_{t=1}^T$ as follows:
   $\hat V_{T+1}(\ais_{T+1}) \coloneqq 0$ and for $t \in \{T, \dots,
  1\}$:
  \begin{subequations}\label{eq:DP-ais-ac}
  \begin{align}
    \hat Q_t(\ais_t, \aact_t) &\coloneqq \rewinfo_t(\ais_t, \aact_t) 
    + \int_{\AisSp_t} \hat V_{t+1}(\ais_{t+1}) 
    \nextinfo_t(d \ais_{t+1} \mid \ais_t, \aact_t),
    \\
    \hat V_t(\ais_t) &\coloneqq \max_{\aact_t \in \aActSp} \hat Q_t(\ais_t, \aact_t).
  \end{align}
  \end{subequations}
  Then, we have the following:
  \begin{enumerate}
    \item \textbf{\textup{Value function approximation:}} For any time~$t$,
      realization~$\his_t$ of $\His_t$, and choice $\act_t$ of $\Act_t$, we have
      \begin{equation}\label{eq:value-approx-ac}
        \lvert Q_t(\his_t, \act_t) - \hat Q_t(\ainfo_t(\his_t), \aquant(\act_t))\rvert 
        \le \alpha_t
        \quad\text{and}\quad
        \lvert V_t(\his_t) - \hat V_t(\ainfo_t(\his_t)) \rvert 
        \le \alpha_t,
      \end{equation}
      where $\alpha_t$ is defined as in Theorem~\ref{thm:ais}.
    \item \textbf{\textup{Approximately optimal policy:}} Let $\hat \pol = (\hat \pol_1, \dots,
      \hat \pol_T)$, where $\hat \pol_t \colon \AisSp_t \to \Delta(\aActSp)$,
      be a stochastic policy that satisfies 
      \begin{equation}\label{eq:ais-opt-ac}
        \Supp(\hat \pol_t(\ais_t)) \subseteq 
        \arg \max_{\aact_t \in \aActSp} \hat Q_t(\ais_t, \aact_t).
      \end{equation}
      Define policy $\pol = (\pol_1, \dots, \pol_T)$, where $\pol_t
      \colon \HisSp_t \to \Delta(\ActSp)$ by $\pol_t \coloneqq \hat \pol_t \circ
      \ainfo_t$. Then, for any time~$t$, realization~$\his_t$ of $\His_t$, and
      choice $\act_t$ of $\Act_t$, we
      have
      \begin{equation}\label{eq:policy-approx-ac}
        \lvert Q_t(\his_t, \act_t) - Q^\pol_t(\his_t, \aquant(\act_t))\rvert 
        \le 2\alpha_t
        \quad\text{and}\quad
        \lvert V_t(\his_t) - V^\pol_t(\his_t) \rvert 
        \le 2\alpha_t.
      \end{equation}
  \end{enumerate}
\end{theorem}
\begin{proof}
  The proof is similar to the proof of Theorem~\ref{thm:ais}. We only show the
  details of the first part (value approximation). The second part (policy
  approximation) follows along similar lines.

  As before, we prove the result by backward induction.
  Eq.~\eqref{eq:value-approx-ac} holds at $T+1$ by definition. This forms the
  basis of induction. Assume that~\eqref{eq:value-approx-ac} holds at time~$t+1$
  and consider the system at time~$t$. We have that
  \begin{align*}
    \bigl| Q_t(\his_t, \act_t) &-
    \hat Q_t(\ainfo_t(\his_t), \aquant(\act_t)) \bigr|
    \\
    &\stackrel{(a)}\le
     \bigl| \EXP[ R_t \mid \His_t = \his_t, \Act_t = \act_t ]
          - \rewinfo_t(\ainfo_t(\his_t), \aquant(\act_t)) \bigr|
    \\
    &\quad + 
    \EXP\bigl[ \bigl|  V_{t+1}(\His_{t+1}) - \hat V_{t+1}(\ainfo_{t+1}(\His_{t+1}))
      \bigr| \bigm| \His_t = \his_t, \Act_t = \act_t \bigr] 
    \\
    &\quad+ 
    \biggl| \EXP[ \hat V_{t+1}(\ainfo_{t+1}(\His_{t+1})) \mid \His_t = \his_t,
    \Act_t = \act_t ] - 
    \int_{\AisSp_t} \hat V_{t+1}(\ais_{t+1}) \nextinfo_t(d\ais_{t+1} \mid
    \ainfo_t(\his_t), \aact_t) \biggr|
    \\
    &\stackrel{(b)}\le
    \varepsilon_t + \alpha_{t+1} + \Minkowski_{\F}(\hat V_{t+1})
    \delta_{t} = \alpha_t
  \end{align*}
  where $(a)$ follows from triangle inequality and $(b)$ follows from (AQ1),
  the induction hypothesis, (AQ2) and~\eqref{eq:minkowski}. This proves the
  first part of~\eqref{eq:value-approx-ac}. The second part follows from
  \[
    \bigl| V_t(\his_t) - \hat V_t(\ainfo_t(\his_t)) \bigr| 
    \stackrel{(a)}\le 
    \max_{\act_t \in \ActSp} 
    \bigl| Q_t(\his_t, \act_t) - \hat Q_t(\ainfo_t(\his_t), \aquant(\act_t)) \bigr|
    \le \alpha_t,
  \]
  where $(a)$ follows from the inequality $\max f(x) \le \max | f(x) - g(x) |
  +  \max g(x)$. We have also used the fact that if $\aquant$ is an onto
  function, then
  \(
    \max_{\aact \in \aActSp} \hat Q_t(\ais_t, \aact_t) = 
    \max_{\act \in \ActSp} \hat Q_t(\ais_t, \aquant(\act_t))
  \). 
  This completes the proof of value approximation. The proof of policy
  approximation is similar to that of Theorem~\ref{thm:ais} adapted in the
  same manner as above. 
\end{proof}

Action quantization in POMDPs with finite or Borel valued unobserved state was
investigated in~\cite{saldi2018finite}, who showed that under
appropriate technical conditions the value function and optimal policies for
the quantized model converge to the value function and optimal policy of the
true model. However~\cite{saldi2018finite} did not provide approximation
error for a fixed quantization level.  

\paragraph{Simplification for perfectly observed case:}
The approximation bounds for action compression 
derived in Theorem~\ref{thm:ais-ac} can be simplified when the system is
perfectly observed. In particular, consider an MDP with state space $\StSp$,
action space $\ActSp$, transition probability $P \colon \StSp \times \ActSp
\to \Delta(\StSp)$, per-step reward function $r \colon \StSp \times \ActSp \to
\reals$, and discount factor $\discount$.

For MDPs, we can simplify the definition of action quantized
\acs{AIS}-generator as follows.

\begin{definition} \label{def:ais-ac-only}
  Given an MDP as defined above, let $\F$ be a function class for
  \textup{IPMs}, and $(\varepsilon, \delta)$ be pre-specified positive real numbers. In
  addition, suppose we have a subset $\aActSp \subset \ActSp$ of quantized
  actions. Then, an action quantization function $\aquant \colon \ActSp \to \aActSp$,
  where $\aActSp \subset \ActSp$,
  is called an
  \emph{$(\varepsilon, \delta)$-action-quantizer}
   if the following properties are satisfied:
  \begin{description}
    \item[(AQM1)] \textbf{\textup{Sufficient for approximate performance
      evaluation}}, i.e., for any $\st \in \StSp$ and $\act \in \ActSp$, we
      have
      \[
        \bigl\lvert \RewFn(\st, \act) - 
        \RewFn(\st, \aquant(\act)) \bigr\rvert 
        \le \varepsilon.
      \]
    \item[(AQM2)] \textbf{\textup{Sufficient to predict the next state approximately}}.
      i.e., for any $\st \in \StSp$ and $\act \in \ActSp$,
      \[
        d_\F(P(\cdot | \st, \act), P(\cdot | \st, \aquant(\act))
         \le \delta.
      \]
  \end{description}
\end{definition}

Then, the approximation in Theorem~\ref{thm:ais-ac} simplifies for an MDP as
follows.

\begin{corollary}\label{cor:ais-ac-only}
  Suppose $\aquant$ is an $(\varepsilon, \delta)$-action-quantizer.
  Recursively define approximate action-value functions $\{\hat Q_t \colon
    \StSp \times \aActSp \to \reals \}$ and value functions $\{\hat
    V_t \colon \StSp_t \to \reals\}$ as follows:
   $\hat V_{T+1}(\st_{T+1}) \coloneqq 0$ and for $t \in \{T, \dots,
  1\}$:
  \begin{subequations}\label{eq:DP-ais-ac-only}
  \begin{align}
    \hat Q_t(\st_t, \aact_t) &\coloneqq \RewFn(\st_t, \aact_t) 
    + \int_{\StSp} \hat V_{t+1}(\st_{t+1}) 
    P(d \st_{t+1} \mid \st_t, \aact_t),
    \\
    \hat V_t(\st_t) &\coloneqq \max_{\aact_t \in \aActSp} \hat Q_t(\st_t, \aact_t).
  \end{align}
  \end{subequations}
  Then, we have the following:
  \begin{enumerate}
    \item \textbf{\textup{Value function approximation:}} For any time~$t$,
      $\st \in \StSp$ and $\act \in \ActSp$, we have
      \begin{equation}\label{eq:value-approx-ac-only}
        \lvert Q_t(\st, \act) - \hat Q_t(\st, \aquant(\act))\rvert 
        \le \alpha_t
        \quad\text{and}\quad
        \lvert V_t(\st) - \hat V_t(\st) \rvert 
        \le \alpha_t,
      \end{equation}
      where $\alpha_t$ is defined as in Theorem~\ref{thm:ais}.
    \item \textbf{\textup{Approximately optimal policy:}} Let $\hat \pol = (\hat \pol_1, \dots,
      \hat \pol_T)$, where $\hat \pol_t \colon \StSp \to \Delta(\aActSp)$,
      be a stochastic policy that satisfies 
      \begin{equation}\label{eq:ais-opt-ac-only}
        \Supp(\hat \pol_t(\st_t)) \subseteq 
        \arg \max_{\aact_t \in \aActSp} \hat Q_t(\st_t, \aact_t).
      \end{equation}
      Since $V^{\hat \pol}_t(\st_t) = \hat V_t(\st_t)$ and $Q^{\hat
      \pi}_t(\st, \aact_t) = \hat Q_t(\st, \aact_t)$, we have
      \begin{equation}\label{eq:policy-approx-ac-only}
        \lvert Q_t(\st_t, \act_t) - Q^{\hat \pol}(\st_t, \aquant(\act_t))\rvert %CHECK subscript t to pol in superscript removed.
        \le \alpha_t
        \quad\text{and}\quad
        \lvert V_t(\st_t) - V^{\hat \pol}(\st_t) \rvert  %CHECK subscript t to pol in superscript removed.
        \le \alpha_t.
      \end{equation}
  \end{enumerate}
\end{corollary}
\begin{proof}
The proof follows in a straightforward manner from the proof of Theorem~\ref{thm:ais-ac}.
\end{proof}

Note that in contrast to Theorem~\ref{thm:ais-ac}, the final approximation
bounds~\eqref{eq:policy-approx-ac-only} in Corollary~\ref{cor:ais-ac-only} do
not have an additional factor of~$2$. This is because the approximate policy
$\hat \pi$ can be directly executed in the original MDP because $\aActSp
\subset \ActSp$. 

Approximation bounds similar to Corollary~\ref{cor:ais-ac-only} are used to
derive bounds for lifelong learning in \cite{chandak2020lifelong}. We show
that similar bounds may be obtained using Corollary~\ref{cor:ais-ac-only} in
Appendix~\ref{app:ais-ac-only}.

\subsection{AIS with observation compression} \label{sec:obsais}

In applications with high-dimensional observations such as video input, it is
desirable to pre-process the video frames into a low-dimensional
representation before passing them on to a planning or learning algorithm. In
this section, we generalize the notion of \ac{AIS} to account for such
observation compression. 

\begin{definition}
  As in the definition of \acs{AIS}, suppose $\{\AisSp_t\}_{t=1}^T$ are a
  pre-specified collection of Banach spaces, $\F$ be a function class for
  \textup{IPMs}, and $\epsilonDelta$ be pre-specified positive real numbers. In
  addition, suppose we have a set $\aObSp$ of compressed observations and a
  compression function $q \colon \ObSp \to \aObSp$. Let $\aHis_t$ denote the
  history $(\aOb_{1:t-1}, \Act_{1:t-1})$ of compressed observations and
  actions and $\aHisSp_t$ denote the space of realizations of such compressed
  histories. Then, a collection $\{\ainfo_t \colon \aHisSp_t \to \AisSp_t
  \}_{t=1}^T$ of history compression functions, along with observation
  compression function $q \colon \ObSp \to \aObSp$, 
  approximate update kernels $\{\nextinfo_t \colon
  \AisSp_t \times \ActSp \to \Delta(\AisSp_{t+1})\}_{t=1}^T$ and reward
  approximation functions $\{\rewinfo_t \colon \AisSp_t \times \ActSp \to
  \reals\}_{t=1}^T$, is called an
  \emph{$\epsilonDelta$-observation-compressed \acs{AIS}
  generator} if the process $\{\Ais_t\}_{t=1}^T$, where
  $\Ais_t = \ainfo_t(\aHis_t)$, satisfies properties \textup{(AP1)} and
  \textup{(AP2)}.
\end{definition}

\begin{figure}[ht]
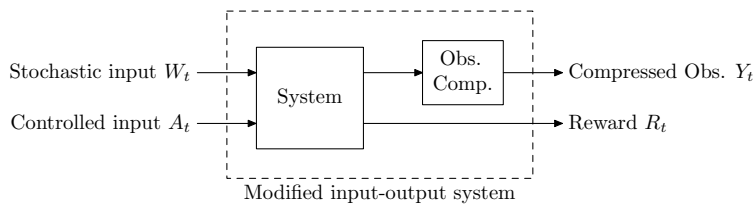

    \centering
    \resizebox{0.65\linewidth}{!}{%
    \begin{mpost}[mpsettings={input boxes;}]
      defaultdx := 10bp;
      defaultdy := 20bp;
      boxit.system(\btex System etex);
      system.c = origin;

      drawboxed(system);

      z1 = 0.5[system.w, system.nw];
      z2 = 0.5[system.w, system.sw];

      z3 = z1 - (1cm,0);
      z4 = z2 - (1cm,0);

      drawarrow z3 -- lft z1;
      drawarrow z4 -- lft z2;

      label.lft(\btex Stochastic input $W_t$ etex, z3);
      label.lft(\btex Controlled input $A_t$ etex, z4);

      z5 = 0.5[system.e, system.ne];
      z6 = 0.5[system.e, system.se];

      z7 = z5 + (1cm, 0);
      z8 = z6 + (1cm, 0);

      boxit.quantizer(\btex \vbox{\hbox{~Obs.} \endgraf \hbox{Comp.}} etex);
      quantizer.dy=5pt;
      quantizer.dx=5pt;
      quantizer.w = z7 ;

      drawarrow z5 -- z7;

      z9 = quantizer.e + (1cm, 0);
      z10 = (x9, y8);

      drawboxed(quantizer);

      drawarrow quantizer.e -- z9;
      drawarrow z6 -- z10;

      z100 = (xpart system.w - 0.5cm, ypart quantizer.n + 0.5cm);
      z101 = (xpart quantizer.e + 0.5cm, ypart quantizer.n + 0.5cm);
      z102 = (xpart quantizer.e + 0.5cm, ypart system.s - 0.5cm);
      z103 = (xpart system.w  - 0.5cm, ypart system.s - 0.5cm);

      draw z100 -- z101 -- z102 -- z103 -- cycle
           dashed evenly;

      label.bot(\btex Modified input-output system etex, 0.5[z102, z103]);

      label.rt(\btex Compressed Obs. $Y_t$ etex, z9);
      label.rt(\btex Reward $R_t$ etex, z10);
    \end{mpost}}
    \caption{A stochastic input-output system with observation compression}
    \label{fig:output-compression}
\end{figure}

In essence, we can view observation compression as a new input-output system
whose outputs are $(\aOb_t, \Rew_t)$ instead of $(\Ob_t, \Rew_t)$ as shown in
Fig.~\ref{fig:output-compression}. A construction similar to
observation-compressed \ac{AIS} is proposed in~\cite{ha2018world}, where it is
shown that such a construction performs well empirically, but there was no
analysis of the approximation guarantees of such a construction.

An immediate implication of the above definition is the following:
\begin{corollary}
  Let $\{ \ainfo_t, q, \nextinfo_t, \rewinfo_t\}_{t \ge 1}$ be an
  $\epsilonDelta$-observation-compression \ac{AIS}. Then, the bounds of
  Theorem~\ref{thm:ais} hold. 
\end{corollary}

\subsection{Discussion and related work}

\acs{AIS} may be viewed as a generalization of state discretization
\citep{Bertsekas_1975} or state aggregation \citep{Whitt_1978} in MDPs. As
illustrated by the examples in Sec.~\ref{ex:ais}, many of the recent results
on approximation bounds for state aggregation and latent state embedding in
MDPs are specific instances of \acs{AIS} and, in some instances, using the
approximation bounds of Theorem~\ref{thm:ais} or its generalization to
infinite horizon (Theorem~\ref{thm:inf-ais}) provide tighter bounds than those
in the literature. A detailed comparison with these results is presented in
the Appendices. We had presented a simpler definition of AIS and the
approximation bounds in the preliminary version of this paper \citep{CDC}.

As mentioned in Sec.~\ref{discuss:info-state} while discussing the related literature
on information states, there are two other methods for identifying ``states''
for POMDPs: bisimulation-based methods and predictive state representations (PSRs). Approximation techniques for both
these methods have been proposed in the literature.

State aggregation techniques based on bisimulation metrics have been proposed
in \cite{FernsPanangadenPrecup_2004, FernsPanangadenPrecup_2011} for MDPs and
\cite{CastroPanangadenPrecup_2009} for POMDPs. The key insight of these
papers is to define a semi-metric called bisimulation metric on the state
space of an MDP or the belief space of a POMDP as the unique fixed point of an
operator on the space of semi-metrics on the state space of the MDP or the
belief space of the POMDP. It is then shown that the value function is
Lipschitz with respect to this metric. Then, they propose state aggregation
based on the bisimulation metric. Although the basic building blocks of
bisimulation metrics are the same as those of an \ac{AIS}, the approximation
philosophies are different. The bisimulation-metric based approximations are
a form of state aggregation, while \ac{AIS} need not be a state
aggregation.

Various methods for learning low dimensional approximations of PSRs have been
proposed in the literature, including approaches which use spectral learning
algorithms \citep{RosencrantzGordonThrun_2004,Boots2011, Hamilton2014,
Kulesza2015,Kulesza2015a,Jiang2016}, and stochastic gradient descent
\citep{Jiang2016}. Error bounds for using an approximate PSR were derived in
\cite{Wolfe2008, Hamilton2014}. These approximation methods for PSRs rely on the specific structure of PSRs and are conceptually different from the approximation
methods used in \ac{AIS}.

\section{Infinite-horizon discounted reward setup}\label{sec:infinite}

So far, we have restricted attention to the finite horizon setup. In this section,
we show how to generalize the notions of information state and \acl{AIS} to the 
infinite horizon discounted reward setup.

\subsection{System model and problem formulation}

We consider the same model as described in Sec.~\ref{sec:model} but assume that
the system runs for an infinite horizon. The performance of any
(history dependent and possibly stochastic) policy $\pol \coloneqq (\pol_1,
\pol_2, \dots)$, where $\pol_t \colon \HisSp_t \to \Delta(\ActSp)$, is given
by
\[
  J(\pol) \coloneqq \liminf_{T \to \infty}
  \EXP^\pol\biggl[ \sum_{t=1}^T \discount^{t-1} R_t \biggr],
\]
where $\discount \in (0,1)$ is the discount factor. As before, we assume that
the agent knows the system dynamics $\{f_t\}_{t \ge 1}$, the reward functions
$\{r_t \}_{t \ge 1}$, and the probability measure $\PR$ on the primitive
random variables $\{W_t\}_{t \ge 1}$. The objective of the agent
is to choose a policy $\pol$ that maximizes the expected discounted total reward
$J(\pol)$. 

Note that we use $\liminf$ rather than $\lim$ in the above definition because
in general the limit might not exist. We later assume that the rewards are
uniformly bounded (see Assumption~\ref{ass:bounded}) which, together with the
finiteness of the action space, implies that the limit is well defined. When the
action space is uncountable, we need to impose appropriate technical conditions
on the model to ensure that an appropriate measurable selection condition holds
\citep{Hernandez-LermaLasserre_2012}.

\subsection{A dynamic programming decomposition}

In the finite-horizon setup, we started with a dynamic program to evaluate the
performance $\{V^\pol_t\}_{t=1}^T$ for any history dependent policy~$\pol$. We
then identified an upper-bound $\{V_t\}_{t=1}^T$ on $\{V^\pol_t\}_{t=1}^T$ and
showed that this upper bound is tight and achieved by any optimal policy. The
subsequent analysis of the information state and the \acl{AIS} based dynamic
programs was based on comparison with $\{V_t\}_{t=1}^T$. 

One conceptual difficulty with the infinite horizon setup is that we cannot
write a general dynamic program to evaluate the performance $\{V^\pol_t\}_{t
\ge 1}$ of an arbitrary history dependent policy $\pol$ and therefore identify
a tight upper-bound $\{V_t\}_{t \ge 1}$. In traditional MDP models, this
conceptual difficulty is resolved by restricting attention to Markov
strategies and then establishing that the performance of a Markov strategy can
be evaluated by solving a fixed point equation. For partially observed MDPs, a
similar resolution works because one can view the belief state as an
information state. However, for general partially observed models as
considered in this paper, there is no general methodology to identify a
time-homogeneous information state. So, we follow a different approach and
identify a dynamic program which bounds the performance of a general history
dependent policy. We impose the following mild
assumption on the model.

\begin{assumption} \label{ass:bounded}
  The reward process $\{\Rew_t\}_{t \ge 1}$ is uniformly bounded and takes
  values inside a finite interval $[\Rew_{\min}, \Rew_{\max}]$. 
\end{assumption}

Given any (history dependent) policy $\pol$, we define the \emph{reward-to-go}
function for any time~$t$ and any realization $\his_t$ of $\His_t$ as
\begin{equation}
  V^\pol_t(\his_t) \coloneqq 
  \EXP^\pol\biggl[ \sum_{s=t}^\infty \discount^{s-t} \Rew_s \biggm| \His_t = \his_t
  \biggr].
\end{equation}
Define the corresponding action value function as:
\begin{equation}
  Q^\pol_t(\his_t, \act_t) \coloneqq
  \EXP^\pol[ R_t + \discount V^\pol_{t+1}(\His_{t+1}) 
  \mid \His_t = \his_t, \Act_t = \act_t ] .
\end{equation}
As stated above, we cannot identify a dynamic program to recursively compute
$\{ V^\pol_t\}_{t \ge 1}$. Nonetheless, we show that under
Assumption~\ref{ass:bounded} we can identify arbitrarily precise upper and
lower bounds for $\{ V^\pol_t\}_{t \ge 1}$ which can be recursively computed. 

\begin{proposition}\label{prop:eval-inf}
  Arbitrarily pick a horizon $T$ and define $\{ J^\pol_{t,T} \colon \HisSp_t
  \to \reals \}_{t=1}^{T}$ as follows: $J^\pol_{T,T}(\his_{T}) = 0$ and
  for $t \in \{T-2, \dots, 1\}$, 
  \begin{equation}\label{eq:pol-eval-inf}
     J^\pol_{t,T}(\his_t) \coloneqq \EXP^\pol[ \Rew_t + \discount
     J^\pol_{t+1,T}(\His_{t+1}) \mid \His_t = \his_t ].
  \end{equation}
  Then, for any time $t \in \{1, \dots, T\}$ and realization $\his_t$ of
  $\His_t$, we have
  \begin{equation}\label{eq:pol-eval-bound}
    J^\pol_{t,T}(\his_t) + \frac{\discount^{T-t}}{1 - \discount} \Rew_{\min}
    \le V^\pol_t(\his_t) \le
    J^\pol_{t,T}(\his_t) + \frac{\discount^{T-t}}{1 - \discount} \Rew_{\max}.
  \end{equation}
\end{proposition}
\begin{proof}
  The proof follows from backward induction. Note that for $t=T$, $\Rew_t
  \in [\Rew_{\min}, \Rew_{\max}]$ implies that 
  \[
    \frac{\Rew_{\min}}{1-\discount} \le V^\pol_{T}(\his_T) \le 
    \frac{\Rew_{\max}}{1-\discount}.
  \]
  This forms the basis of induction. Now asusme that~\eqref{eq:pol-eval-bound}
  holds for time~$t+1$ and consider the model for time~$t$:
  \begin{align*}
    V^\pol_t(\his_t) &= \EXP^\pol\biggl[ \sum_{s=t}^{\infty} \discount^{s-t} \Rew_s
    \biggm| \His_t = \his_t \biggr]
  \\
  &\stackrel{(a)}= \EXP^\pol\biggl[ \Rew_t + 
    \discount \EXP^\pol\biggl[\sum_{s=t+1}^{\infty} \discount^{s-(t+1)} \Rew_s
    \biggm| \His_{t+1}  \biggr] 
    \biggm| \His_t = \his_t \biggr]
  \\
  &\stackrel{(b)}\le \EXP^\pol\biggl[ \Rew_t + \discount 
    \EXP^\pol\biggl[ J^\pol_{t+1,T}(\His_{t+1}) 
    + \frac{\discount^{T-(t+1)}}{1 - \discount} \Rew_{\max} 
    \biggm| \His_{t+1}  \biggr] 
    \biggm| \His_t = \his_t \biggr]
    \\
  &\stackrel{(c)}=
    J^\pol_{t,T}(\his_t) 
    + \frac{\discount^{T-t}}{1 - \discount} \Rew_{\max},
  \end{align*}
  where $(a)$ follows from the smoothing property of conditional expectation,
  $(b)$ follows from the induction hypothesis, and $(c)$ follows from the
  definition of $J^\pol_{t,T}(\cdot)$. This establishes one side
  of~\eqref{eq:pol-eval-bound}. The other side can be established in a similar
  manner. Therefore, the result holds by the principle of induction.
\end{proof}

Note that Proposition~\ref{prop:eval-inf} gives a recursive method to
approximately evaluate the performance of any history dependent policy~$\pol$.
We can modify the recursion in~\eqref{eq:pol-eval-inf} to obtain policy
independent upper bound on performance of an arbitrary policy. For that
matter, define value functions $\{ V_t \colon \HisSp_t \to \reals\}_{t \ge 1}$as follows:
\begin{equation}\label{eq:value-inf}
  V_t(\his_t) = \sup_{\pol} V^\pol_t(\his_t),
\end{equation}
where the supremum is over all history dependent policies. Furthermore, define
action-value functions $\{ Q_t \colon \HisSp_t \times \ActSp \to \reals\}_{t
\ge 1}$ as follows:
\begin{equation}\label{eq:Q-inf}
  Q_t(\his_t, \act_t) = \EXP[ R_t + \discount V_{t+1}(\His_{t+1}) 
  \mid \His_t = \his_t, \Act_t = \act_t ].
\end{equation}
Then, we have the following.

\begin{proposition}\label{prop:optimal-inf}
  Arbitrarily pick a horizon $T$ and define $\{ J_{t,T} \colon \HisSp_t \to
  \reals\}$ as follows: $J_{T,T}(\his_{T}) = 0$ and for $t \in \{T-2, \dots,
  1\}$, 
  \begin{equation}\label{eq:DP-inf}
    J_{t,T}(\his_t) \coloneqq \max_{\act_t \in \ActSp} \EXP[
      \Rew_t + \discount J_{t+1}(\His_{t+1}) \mid 
    \His_t = \his_t, \Act_t = \act_t ].
  \end{equation}
  Then, for any time $t \in \{1, \dots, T\}$ and realization $\his_t$ of
  $\His_t$,
  \begin{equation}\label{eq:pol-eval-bound-2}
    V^\pol_t(\his_t) \le J_{t,T}(\his_t) + 
     \frac{\discount^{T-t}}{1 - \discount} \Rew_{\max}.
  \end{equation}
  Therefore, 
  \begin{equation} \label{eq:opt-bound}
    J_{t,T}(\his_t) +
     \frac{\discount^{T-t}}{1 - \discount} \Rew_{\min}
    \le V_t(\his_t) \le
    J_{t,T}(\his_t) +
    \frac{\discount^{T-t}}{1 - \discount} \Rew_{\max}.
  \end{equation}
\end{proposition}
Note that $J_{t,T}(\his_t)$ is the optimal value function for a finite horizon
system with the discounted reward criterion that runs for horizon $T-1$.
\begin{proof}
  By following almost the same argument as Proposition~\ref{prop:optimal}, we
  can establish that for any history dependent policy~$\pol$, $J^\pol_{t,T}(\his_t) \le
  J_{t,T}(\his_t)$, which immediately implies~\eqref{eq:pol-eval-bound-2}.

  Maximizing the left hand side of~\eqref{eq:pol-eval-bound-2} gives us the
  upper bound in~\eqref{eq:opt-bound}. For the lower bound
  in~\eqref{eq:opt-bound}, observe that
  \begin{align*}
    V_t(\his_t) &= 
    \sup_{\pol} \EXP^\pol\biggl[
      \sum_{s=t}^\infty \discount^{s-t} \Rew_s 
    \biggm| \His_t = \his_t \biggr]
    \\
    &\stackrel{(a)}\ge 
    \sup_{\pol} \EXP^\pol\biggl[
      \sum_{s=t}^{T-1} \discount^{s-t} \Rew_s + 
      \sum_{s=T}^\infty \discount^{s-t} \Rew_{\min} 
    \biggm| \His_t = \his_t \biggr]
    \\
    &=
    \sup_{\pol} \EXP^\pol\biggl[
      \sum_{s=t}^{T-1} \discount^{s-t} \Rew_s 
    \biggm| \His_t = \his_t \biggr]
    + \frac{\discount^{T-t}}{1-\discount} \Rew_{\min} 
    \\
    &\stackrel{(b)}= J_{t,T}(\his_t) 
    + \frac{\discount^{T-t}}{1-\discount} \Rew_{\min} .
  \end{align*}
  where $(a)$ follows from the fact that $\Rew_s \ge \Rew_{\min}$ and $(b)$
  follows from the definition of $J_{t,T}(\his_t)$. This complete the proof
  of~\eqref{eq:opt-bound}.
\end{proof}

\subsection{Time-homogeneous information state and simplified dynamic program}

\begin{definition}
  Given a Banach space $\IsSp$, an information state generator $\{\info_t \colon
  \HisSp_t \to \IsSp\}$ is said to be
  \emph{time-homogeneous} if, in addition to \textup{(P1)} and \textup{(P2)},
  it satisfies the following:
  \begin{description}
    \item[(S)] The expectation
      \(
        \EXP[\Rew_t | \Is_t = \info_t(\His_t), \Act_t = \act_t ]
      \)
      and the transition kernel
      \(
        \PR(\Is_{t+1} \in B | \Is_t = \info_t(\His_t), \Act_t = \act_t)
      \)
      are time-homogeneous.
  \end{description}
\end{definition}

Note that all except the first example of information state presented in
Sec.~\ref{ex:info-state} are time-homogeneous. However, in general, a
time-homogeneous information state may not exist for all partially observed
models and it is important to understand conditions under which such an
information state exists. However, we do not pursue that direction in this
paper.

For any time-homogeneous information state, define the Bellman operator
$\mathcal{B} \colon [ \IsSp \to \reals]  \to [\IsSp \to \reals]$ as
follows: for any uniformly bounded function $\bar V \colon \IsSp \to \reals$
\begin{equation} \label{eq:bellman}
  [\mathcal{B} \bar V](\is) = \max_{\act \in \ActSp}
  \EXP[ \Rew_t + \discount \bar V(\Is_{t+1}) \mid \Is_t = \is, \Act_t = \act ],
\end{equation}
where $\discount \in (0,1)$ is the discount factor. Because of (S), the expectation on the right hand side does not depend on
time. Due to discounting, the operator $\mathcal B$ is a contraction and
therefore, under Assumption~\ref{ass:bounded}, the fixed point
equation 
\begin{equation} \label{eq:fixed-point}
  \bar V = \mathcal{B} \bar V
\end{equation}
has a unique bounded solution (due to the Banach fixed point theorem).
Let $\bar V^*$ be the fixed point and $\pol^*$ be any policy such that $\pol^*(\is)$
achieves the arg max in the right hand side of~\eqref{eq:bellman} for
$[\mathcal{B} \bar V^*](\is)$. It is easy to see that $\bar V^*$ is the performance of the
time homogeneous policy $(\pol^*, \pol^*, \dots)$. However, it is not obvious
that $\bar V^*$ equals to the optimal performance $V_1$ defined
in~\eqref{eq:value-inf},
because the proof of Theorem~\ref{thm:info-state} relies on backward induction
and is not applicable to infinite horizon models. So, we present an
alternative proof below which uses the performance bounds of
Proposition~\ref{prop:optimal-inf}.

\begin{theorem}\label{thm:inf-is}
  Let $\{\Is_t\}_{t \ge 1}$ be a time-homogeneous information state process
  with generator $\{ \info_t \colon \HisSp_t \to \IsSp \}_{t \ge 1}$.
  Suppose Assumption~\ref{ass:bounded} holds and let $\bar V^*$ be the unique
  bounded fixed point of~\eqref{eq:bellman}. Then, for any time~$t$ and
  realization $\his_t$ of $\His_t$, we have
  \[
    V_t(\his_t) = \bar V^*(\info_t(\his_t)).
  \]
  Furthermore, let $\pol^* \colon \IsSp \to \Delta(\ActSp)$ be a
  time-homogeneous (stochastic) policy such that $\Supp(\pol^*(\is))$ is a
  subset of the arg max of the right hand side of~\eqref{eq:bellman}. Then,
  the time-homogeneous policy $\pol^* \coloneqq (\pol^*, \pol^*, \dots)$ is
  optimal.
\end{theorem}
\begin{proof}
  Consider the following sequence of value functions: $\bar V^{(0)}(\is) = 0$ and
  for $n \ge 0$, define $\bar V^{(n+1)} = \mathcal{B} \bar V^{(n)}$. Now fix a
  horizon~$T$ and consider the finite-horizon discounted reward problem of
  horizon~$T-1$. As argued earlier, $J_{t,T}(\his_t)$ is the optimal
  value-function for this finite horizon discounted problem. Moreover, note
  that $\{\Is_t\}_{t=1}^T$ is an information state for this finite horizon
  discounted problem. Therefore, from using the result of
  Theorem~\ref{thm:info-state}, we get that for
    any time $t \in \{1, \dots, T\}$, and
      realization $\his_t$ of $\His_t$, 
      \[
        J_{t,T}(\his_t) = \bar V^{(T-t)}(\info_t(\his_t)).
      \]

  Substituting~\eqref{eq:opt-bound} from Proposition~\ref{prop:optimal-inf} in
  the above, we get
  \[
    \bar V^{(T-t)}(\info_t(\his_t)) +
     \frac{\discount^{T-t}}{1 - \discount} \Rew_{\min}
    \le V_t(\his_t) \le
    \bar V^{(T-t)}(\info_t(\his_t)) +
    \frac{\discount^{T-t}}{1 - \discount} \Rew_{\max}.
  \]
  The result follows from taking limit $T \to \infty$ and observing that
  $\bar V^{(T-t)}(\is)$ converges to $\bar V^*(\is)$.
\end{proof}

\subsection{Time-homogeneous AIS and approximate dynamic programming}

\begin{definition}
  Given a Banach space $\AisSp$, a function class $\F$ for IPMs, and positive
  real numbers $(\varepsilon, \delta)$, we say
  that a collection $\{\ainfo_t \colon \HisSp_t \to \Ais \}_{t \ge 1}$ along
  with a time-homogeneous update kernel $\nextinfo \colon \AisSp \times \ActSp \to
  \Delta(\AisSp)$ and a time-homogeneous reward approximation function
  $\rewinfo \colon \AisSp \times \ActSp \to \reals$ is a
  \emph{$(\varepsilon,\delta)$ time
  homogeneous \ac{AIS} generator} if the process $\{\Ais_t\}_{t \ge 1}$,
  where $\Ais_t = \ainfo_t(\His_t)$, satisfies \textup{(AP1)} and
  \textup{(AP2)} where $\rewinfo_t$, $\nextinfo_t$, $\varepsilon_t$ and
  $\delta_t$ in the definition of \textup{(AP1)} and \textup{(AP2)} are replaced by their
  time-homogeneous counterparts.
\end{definition}

For any time-homogeneous \ac{AIS}, define the approximate Bellman operator
$\hat{\mathcal{B}} \colon [\AisSp \to \reals] \to [\AisSp \to \reals]$ as
follows: for any uniformly bounded function $\hat V \colon \AisSp \to \reals$, 
\begin{equation}\label{eq:bellman-operator-ais}
  [\hat{\mathcal B} \hat V](\ais) = \max_{\act \in \ActSp}
  \biggl\{ 
      \rewinfo(\ais, \act) + \discount \int_{\AisSp} \hat V(\ais') \nextinfo(d\ais' |
      \ais, \act)
  \biggr\}.
\end{equation}
Note that the expectation on the right hand side does not depend on
time. Due to discounting, the operator $\hat{\mathcal{B}}$ is a contraction,
and therefore, under Assumption~\ref{ass:bounded}, the fixed point equation
\begin{equation}\label{eq:bellman-ais}
  \hat V = \hat{\mathcal{B}}\hat V
\end{equation}
has a unique bounded solution (due to the Banach fixed point theorem). Let $\hat
V^*$ be the fixed point and $\hat \pol^*$ be any policy such that $\hat
\pol^*(\ais)$ achieves the arg max in the right hand side
of~\eqref{eq:bellman-operator-ais} for $[\hat{\mathcal{B}} \hat V^*](\ais)$. It is not
immediately clear if $\hat V^*$ is close to the performance of
policy $\pol = (\pol_1, \pol_2, \dots)$, where $\pol_t = \pol^* \circ
\ainfo_t$, or if $\hat V^*$ is close to the optimal performance. The proof of
Theorem~\ref{thm:ais} relies on backward induction and is not immediately
applicable to the infinite horizon setup. Nonetheless, we establish results
similar to Theorem~\ref{thm:ais} by following the proof idea of
Theorem~\ref{thm:inf-is}.

\begin{theorem}\label{thm:inf-ais}
  Suppose $(\{\ainfo_t\}_{t \ge 1}, \nextinfo, \rewinfo)$ is a time-homogeneous
  $(\varepsilon,\delta)$-\ac{AIS} generator. Consider the fixed point
  equation~\eqref{eq:bellman-ais}, which we rewrite as follows:
  \begin{subequations}\label{eq:DP-ais-inf}
  \begin{align}
    \hat Q(\ais, \act) &\coloneqq \rewinfo(\ais, \act) 
    + \discount \int_{\AisSp} \hat V(\ais') 
    \nextinfo(d \ais' | \ais, \act),
    \\
    \hat V(\ais) &\coloneqq \max_{\act \in \ActSp} \hat Q(\ais, \act).
  \end{align}
  \end{subequations}
  Let $\hat V^*$ denote the fixed point of~\eqref{eq:DP-ais-inf} and $\hat
  Q^*$ denote the corresponding action-value function. 
  Then, we have the following:
  \begin{enumerate}
    \item \textbf{\textup{Value function approximation:}} For any time~$t$,
      realization~$\his_t$ of $\His_t$, and choice $\act_t$ of $\Act_t$, we have
      \begin{equation}\label{eq:value-approx-inf}
        \lvert Q_t(\his_t, \act_t) - \hat Q^*(\ainfo_t(\his_t), \act_t)\rvert 
        \le \alpha
        \quad\text{and}\quad
        \lvert V_t(\his_t) - \hat V^*(\ainfo_t(\his_t)) \rvert 
        \le \alpha,
      \end{equation}
      where 
      \[
        \alpha = \frac{ \varepsilon + \discount \Minkowski_\F(\hat V^*) \delta}{1 - \discount}
      \]
    \item \textbf{\textup{Approximately optimal policy:}} Let 
      $\hat \pol^* \colon \AisSp \to \Delta(\ActSp)$
      be a stochastic policy that satisfies 
      \begin{equation}\label{eq:ais-opt-inf}
        \Supp(\hat \pol^*(\ais)) \subseteq 
        \arg \max_{\act \in \ActSp} \hat Q^*(\ais, \act).
      \end{equation}
      Define policy $\pol = (\pol_1, \pol_2, \dots)$, where $\pol_t
      \colon \HisSp_t \to \Delta(\ActSp)$ is defined by $\pol_t \coloneqq \hat \pol^* \circ
      \ainfo_t$. Then, for any time~$t$, realization~$\his_t$ of $\His_t$, and
      choice $\act_t$ of $\Act_t$, we
      have
      \begin{equation}\label{eq:policy-approx-inf}
        \lvert Q_t(\his_t, \act_t) - Q^\pol_t(\his_t, \act_t)\rvert 
        \le 2\alpha
        \quad\text{and}\quad
        \lvert V_t(\his_t) - V^\pol_t(\his_t) \rvert 
        \le 2\alpha.
      \end{equation}
  \end{enumerate}
\end{theorem}
\begin{proof}
  The proof follows by combining ideas from Theorem~\ref{thm:ais}
  and~\ref{thm:inf-is}. We provide a detailed proof of the value
  approximation. The proof argument for policy approximation is similar.

  Consider the following sequence of value functions:
  $\hat V^{(0)}(\ais) = 0$ and for $n \ge 0$, define $\hat V^{(n+1)} =
  \hat{\mathcal{B}} \hat V^{(n)}$. Now fix a
  horizon~$T$ and consider the finite-horizon discounted reward problem of
  horizon~$T-1$. As argued earlier, $J_{t,T}(\his_t)$ is the optimal
  value-function for this finite horizon discounted problem. Moreover, note
  that $\{\Ais_t\}_{t=1}^T$ is an $(\varepsilon,\delta)$-\ac{AIS} for this
  finite horizon discounted problem. Therefore, from using the result of
  Theorem~\ref{thm:ais}, we get that for any time $t \in \{1, \dots, T\}$, and
  realization $\his_t$ of $\His_t$, 
  \[
    \lvert J_{t,T}(\his_t) - \hat V^{(T-t)}(\ainfo_t(\his_t)) \rvert
    \le \alpha_t,
  \]
  where 
  \[
    \alpha_t = \varepsilon + \sum_{\tau = t+1}^{T-1} 
    \discount^{\tau - t}\bigl[ \Minkowski_\F(\hat V^{(T-\tau)}) \delta + \varepsilon \bigr].
  \]
  Substituting~\eqref{eq:opt-bound} from Proposition~\ref{prop:optimal-inf} in
  the above, we get that
  \[
    \hat V^{(T-t)}(\ainfo_t(\his_t)) - \alpha_t + 
    \frac{\discount^{T-t}}{1 - \discount} R_{\min}
    \le V_t(\his_t)  \le
    \hat V^{(T-t)}(\ainfo_t(\his_t)) + \alpha_t + 
    \frac{\discount^{T-t}}{1 - \discount} R_{\max}.
  \]
  Since $\hat{\mathcal{B}}$ is a contraction, from the Banach fixed point theorem
  we know that $\lim_{T \to \infty} \hat V^{(T-t)} = \hat V^*$. Therefore, by
  continuity of $\Minkowski_\F(\cdot)$, we have
  $\lim_{T \to \infty} \Minkowski_\F(\hat V^{T-t}) = \Minkowski_\F(\hat V^*)$.
  Consequently, $\lim_{T \to \infty} \alpha_t = \alpha$. Therefore, taking the
  limit $T \to \infty$ in the above equation, we get
  \[
    \hat V^*(\ainfo_t(\his_t)) - \alpha
    \le V_t(\his_t)  \le
    \hat V^*(\ainfo_t(\his_t)) + \alpha,
  \]
  which establishes the bound on the value function
  in~\eqref{eq:value-approx-inf}. The bound on the action-value function
  in~\eqref{eq:value-approx-inf} follows from a similar argument. 
\end{proof}

\blue{Theorem~\ref{thm:inf-ais} shows how the result of Theorem~\ref{thm:ais}
  generalizes to infinite horizon. We can similarly extend the results for
  approximate policy evaluation (as in Sec.~\ref{sec:ais-poleval}), the stochastic AIS case (as in
  Sec.~\ref{sec:stochais}), the action compression case (as in
  Sec.~\ref{sec:actais}), and the observation
compression case (as in Sec.~\ref{sec:obsais}).}

  %   \blue{
  %    \textbf{\textup{Policy value function approximation:}} Given any stochastic policy
  %    $\hat \pol = (\hat \pol_1, \dots,
  %     \hat \pol_T)$, where $\hat \pol_t \colon \AisSp_t \to \Delta(\ActSp)$,
  %   define $\pol = (\pol_1, \dots, \pol_T)$, where $\pol_t
  %     \colon \HisSp_t \to \Delta(\ActSp)$ by $\pol_t \coloneqq \hat \pol_t \circ
  %     \ainfo_t$. Then, for any time~$t$,
  %     realization~$\his_t$ of $\His_t$, and choice $\act_t$ of $\Act_t$, we have
  %     \begin{equation}
  %       \lvert Q^\pol_t(\his_t, \act_t) - \hat Q^{\hat \pol}_t(\ainfo_t(\his_t), \act_t)\rvert 
  %       \le \alpha_t
  %       \quad\text{and}\quad
  %       \lvert V^\pol_t(\his_t) - \hat V^{\hat \pol}_t(\ainfo_t(\his_t)) \rvert 
  %       \le \alpha_t,
  %     \end{equation}
  %     where $\alpha_t$ satisfies the following recursion: $\alpha_{T+1} =
  %     0$ and for $t \in \{T, \dots, 1 \}$,
  %     \[
  %       \alpha_t = \varepsilon_t + \Minkowski_{\F}(\hat V_{t+1})
  %       \delta_{t} + \alpha_{t+1}.
  %     \]
  %   }
  % \blue{The policy value function approximation can be proved easily by noting that~\eqref{eq:policy-approx-2} 
  % is valid for any policy $\hat \pol$ and not necessarily an optimal policy.}

\section{An \acs{AIS}-based approximate dynamic programming for Dec-POMDPs}
\label{sec:dec}

The theory of approximation for partially observed systems presented in the previous section is fairly general and is applicable to other models of decision making as well. As an example, in this section we show how to use the same ideas to obtain approximation results for decentralized (i.e., multi-agent) partially observed models.

There is a rich history of research on these models in multiple research
disciplines. Decentralized multi-agent systems have been studied in Economics
and Organizational Behavior since the mid
1950s~\citep{Marschak1954,Radner1962,MarschakRadner_1972} under the heading
of team theory. Such models have been studied in systems and control since the
mid 1960s under the heading of decentralized stochastic
control~\citep{witsenhausen1968counterexample,Witsenhausen_1971,sandell1978survey}.
Such models have also been studied in Artificial Intelligence since the
2000s~\citep{bernstein2005bounded,
SzerCZ05,seuken2007memory,carlin2008observation} under the heading of
Dec-POMDPs. In the interest of space, we do not provide a detailed overview of
this rich area; instead we refer the reader to the comprehensive survey
articles of~\cite{MMRY:tutorial-CDC,liu2016learning} for a detailed overview
from the perspective of Systems and Control and Artificial Intelligence,
respectively.

We briefly state the facts about this literature which are pertinent to the
discussion below. The general Dec-POMDP problem is NEXP
complete~\citep{bernstein2002complexity}, so it is not possible to derive an
efficient algorithm to compute the optimal solution. Nonetheless, considerable
progress has been made in identifying special cases where a dynamic
programming decomposition is
possible~\citep{WalrandVaraiya:1983, AicardiDavoliMinciardi:1987,
  OoiVerboutLudwigWornell:1997, MT:NCS, MT:real-time, MNT:tractable-allerton,   Nayyar_2011,NayyarMahajanTeneketzis_2013, M:control-sharing, 
  ArabneydiMahajan_2014, OliehoekAmato_2015, Dibangoye2016,
boularias2008exact,KumarZ09}. A high level approach which encapsulates many of
these special cases is the common information approach
of~\cite{NayyarMahajanTeneketzis_2013} which shows that the Dec-POMDP problem
with a specific but relatively general information
structure  can be converted into a single agent, partially observed problem from the
point of view of a virtual agent which knows the information commonly known to
all agents and chooses prescriptions (or partially evaluated policies) which
map the local information at each agent to their respective actions. We
summarize these results in the next subsection and then show how we can
identify an AIS for such models. 

\subsection{Model of a Dec-POMDP}

A Dec-POMDP is a tuple $\langle \ALPHABET K, \StSp, (\ObSp^k)_{k \in \ALPHABET
K}, (\ActSp^k_t)_{k \in \ALPHABET K}, P_1, P, P^y, r \rangle$ where
\begin{itemize}
  \item $\ALPHABET K = \{1, \dots, K\}$ is the set of agents.

  \item $\StSp$ is the state space. $\ObSp^k$, $\ActSp^k$, $k
    \in \ALPHABET K$, are the observation and action spaces of agent~$k$. Let
    $\ObSp = \prod_{k \in \ALPHABET K} \ObSp^k$ and $\ActSp =
    \prod_{k \in \ALPHABET K} \ActSp^k$. We use $S_t \in \StSp$,
    $\Ob_t \coloneqq (\Ob^k_t)_{k \in \ALPHABET K} \in \ObSp$, and $A_t
    \coloneqq (A^k_t)_{k \in \ALPHABET K} \in \ActSp$, to denote the
    system state, observations, and actions at time~$t$.

  \item $P_1 \in \Delta(\StSp)$ is the initial distribution of the
    initial state $\StSp_1$.

  \item $P \colon \StSp \times \ActSp \to \Delta(\StSp)$
    denotes the transition probability of the system, i.e., 
    \begin{align*}
      \PR(S_{t+1} = s_{t+1} \mid S_{1:t} = s_{1:t}, A_{1:t} = a_{1:t}) &= \PR(S_{t+1} = s_{t+1} \mid S_t = s_t, A_t = a_t)\\
       &= P(s_{t+1} | s_t, a_t).
    \end{align*}

  \item $P^y \colon \StSp \times \ActSp \to \Delta(\ObSp)$
    denotes the observation probability of the system, i..e, 
    \begin{align*}
      \PR(\Ob_{t} = \ob_{t} \mid S_{1:t} = s_{1:t}, A_{1:t-1} = a_{1:t-1}) &= \PR(\Ob_{t} = \ob_{t} \mid S_t = s_t, A_{t-1} = a_{t-1})  \\
      &= P^y(\ob_{t} | s_t, a_{t-1}).
    \end{align*}

  \item $r \colon \StSp \times \ActSp \times \StSp \to
    \reals$ denotes the per-step reward function. The team receives a reward
    $R_t = r(S_t, A_t, S_{t+1})$ at time~$t$.
\end{itemize}

\paragraph{Information structure:} A critical feature of a Dec-POMDP
is the \emph{information structure} which captures the knowledge of who knows
what about the system and when. We use $\Istruct^k_t$ to denote the information
known to agent~$k$ at time~$t$. In general, $\Istruct^k_t$ is a subset of the total
information $(\Ob_{1:t}, A_{1:t-1}, R_{1:t-1})$ known to all agents in the system. 
We use $\IstructSp^k_t$ to denote the space of
the information available to agent $k$ at time~$t$. Note that, in general, the
information available to agent~$k$ increases with time. So, $\IstructSp^k_t$
are sets that are increasing with time. Some examples of information structures
are:
\begin{itemize}
  \item \textbf{Delayed sharing:}
    \(
      I^k_t = \{\Ob_{1:t-d}, A_{1:t-d}, \allowbreak \Ob^{k}_{t-d+1:t}, A^k_{t-d+1:t-1} \}.
    \) This models systems where agents broadcast their
    information and communication has delay of~$d$. Planning for models where
    $d=1$ has been considered in~\cite{SandellAthans:1974, Yoshikawa:1975} and
    for general~$d$ has been considered in~\cite{NMT:delay-sharing}.

  \item \textbf{Periodic sharing:} 
    \(
      I^k_t = \{\Ob_{1:t-\tau}, A_{1:t-\tau}, \Ob^{k}_{t-\tau+1:t}, A^k_{t-\tau+1:t-1} \},
    \)
      where 
    \( 
      \tau = p \bigl\lfloor \tfrac tp \bigr\rfloor
    \). This models systems where agents periodically broadcast their
    information every~$p$ steps. Planning for this model has been
    considered in~\cite{OoiVerboutLudwigWornell:1997}.

  \item \textbf{Control sharing:} 
    \(
      I^k_t = \{\Ob^k_{1:t}, A_{1:t-1} \}.
    \)
    This models systems where control actions are observed by everyone
    (which is the case for certain communication and economic applications).
    Planning for variations of this model has been considered
    in~\cite{Bismut:1972, SandellAthans:1974, M:control-sharing}. 

  \item \textbf{Mean-field sharing:}
  \(
    I^k_t = \{S^k_{1:t}, A^k_{1:t-1}, M_{1:t} \},
  \)
  where the state $S_t$ is $(S^1_t, \dots, S^K_t)$, the observation of agent~$k$ is
  $S^k_t$, and $M_t = \bigl( \sum_{k \in \ALPHABET K} \delta_{S^k_t} \bigr)/K$
  denotes the empirical distribution of the states. This models
  systems where mean-field is observed by all agents (which is the case for
  smart grid  and other large-scale systems).
  Planning for variations of this model has been considered
  in~\cite{ArabneydiMahajan_2014}.
\end{itemize}

\paragraph{Policy:} The policy of agent~$k$ is a collection
$\pi^k = (\pi^k_1, \pi^k_2, \dots)$, where $\pi^k_t \colon \IstructSp^k_t \to
\Delta(\ALPHABET A^i)$. We use $\pi =
(\pi^k)_{k \in \ALPHABET K}$ to denote the policy for all agents. The
performance of a policy $\pi$ is given by 
\begin{equation} \label{eq:performance}
  J(\pi) = \EXP^{\pi}\bigg[ \sum_{t=1}^T R_t \bigg].
\end{equation}
The objective is to find a
(possibly time-varying) policy $\pi$ that maximizes the performance $J(\pi)$
defined in~\eqref{eq:performance}.

\subsection{Common information based planning for Dec-POMDPs}

As mentioned earlier, in general, finding the optimal plan for multi-agent teams is
NEXP-complete~\citep{bernstein2002complexity}. However, it is shown
in~\cite{NayyarMahajanTeneketzis_2013} that when the
information structure is of a particular form (known as partial history
sharing), it is possible to reduce the multi-agent planning problem to a
single agent planning problem from the point of view of a virtual
agent called the coordinator. We summarize this approach below. 

\paragraph{Common and local information:} Define
\[
  C_t = \bigcap_{s \ge t} \bigcap_{k \in \ALPHABET K} \Istruct^k_s
  \qquad\text{and}\qquad
  L^k_t = \Istruct^k_t \setminus C_t, \quad k \in \ALPHABET K.
\]
$C_t$ denotes the \emph{common information}, i.e., the information that
is common to all agents all the time in the future and $L^k_t$ denotes
the \emph{local information} at agent~$k$. By construction, $\Istruct^k_t = \{C_t,
L^k_t \}$. Let $\ALPHABET C_t$ and $\ALPHABET L^k_t$ denote the space of
realizations of $C_t$ and $L^k_t$ and let $L_t = (L^k_t)_{k \in \ALPHABET K}$
and $\ALPHABET L_t = \prod_{k \in \ALPHABET K} \ALPHABET L^k_t$. By
construction, $C_t \subseteq C_{t+1}$. Let $C^{\new}_{t+1} = C_{t+1} \setminus
C_t$ denote the new common information at time~$t$. Then, $C_t$ may be written
as $C^{\new}_{1:t}$.

\begin{definition}
The information structure is called \emph{partial history sharing} if
  for any Borel subset $\ALPHABET B$ of $\ALPHABET L^k_{t+1}$ and any
    realization $c_t$ of $C_t$, $\ell^k_t$ of $L^k_t$, $\act^k_t$ of
    $\Act^k_t$ and $\ob^k_{t+1}$ of $\Ob^k_{t+1}$, we have
    \begin{multline*}
      \PR(L^k_{t+1} \in \ALPHABET B \mid C_t = c_t, L^k_t = \ell^k_t, 
      \Act^k_t = \act^k_t, \Ob^k_{t+1} = \ob^k_{t+1}) 
      \\=
      \PR(L^k_{t+1} \in \ALPHABET B \mid L^k_t = \ell^k_t, 
      \Act^k_t = \act^k_t, \Ob^k_{t+1} = \ob^k_{t+1}).
    \end{multline*}
\end{definition}

\blue{The main intuition behind this definition is as follows. For any system,
  the information available to the agents can always be split into common and
  local information such that $I^k_t = \{ C_t, L^k_t\}$. A partial history
  sharing information structure satisfies the property that at any time~$t$
  and for any agent~$k$, the updated value $L^k_{t+1}$ of the local
  information is a function of only the current local information $L^k_t$, the
  current local action $A^k_t$ and the next local observation $Y^k_{t+1}$.
  Consequently, the common information $C_t$ is not needed to keep
track of the update of the local information. This ensures that compressing the
common information into an information state or an approximate information
state does not impact the update of the local information.}

\paragraph{Prescriptions:}
Given a policy $\pi = (\pi^k)_{k \in \ALPHABET K}$ and a realized
trajectory $(c_1, c_2, \dots)$ of the common information, the prescription
$\Psc^k_t$ is the partial application of $c_t$ to $\pi^k_t$, i.e.,
\(
  \Psc^k_t = \pi^k_t(c_t, \cdot)
\), 
\(
  k \in \ALPHABET K.
\)
Note that $\Psc^k_t$ is a function from $\ALPHABET L^k_t$ to $\Delta(\ALPHABET
A^k_t)$. Let $\Psc_t$ denote $(\Psc^k_t)_{k \in \ALPHABET K}$ and let $\PscSp$ 
denote the space of all such prescriptions for time $t$. 

\blue{The reason for constructing prescriptions is as follows. Prescriptions
  encode the information about the policies of all agents needed to evaluate
  the conditional expected per-step reward given the common information, i.e.,
  $\EXP[ R_t | C_t, (\pi^k)_{k \in \ALPHABET K}]$ can be written as a function
  of $C_t$ and $(\Psc^k_t)_{k \in \ALPHABET K}$, say $\hat r_t(C_t,
(\Psc^k_t)_{k \in \ALPHABET K})$. This allows us to construct a virtual single-agent 
optimization problem where a decision maker (which we call the virtual
coordinator) observes the common information $C_t$ and chooses the
prescriptions $(\Psc^k_t)_{k \in \ALPHABET K}$ to maximize the sum of rewards
$\hat r_t(C_t, (\Psc^k_t)_{k \in \ALPHABET K})$. The details of this virtual
coordinated system are presented next.}

\paragraph{A virtual coordinated system:}

The key idea of \cite{NayyarMahajanTeneketzis_2013} is to construct a
virtual single agent planning problem which they call a coordinated system.
The environment of the virtual coordinated system consists of two components:
the first component is the same as the environment of the original multi-agent
system which evolves according to dynamics~$P$; the second component consists
of $K$ \emph{passive agents}, whose operation we will describe later. There is
a virtual coordinator who observes the common information $C_t$ and chooses
\emph{prescriptions} $\PSC_t = (\PSC^k_t)_{k \in \ALPHABET K}$, where
$\PSC^k_t \colon \ALPHABET L^k \to \Delta(\ALPHABET A^k)$ using a
\emph{coordination rule}~$\psi_t$, i.e., $\PSC_t \sim \psi_t(C_t)$. In general, the 
coordination rule can be stochastic. Let $\Psc_t$ denote the realization of
$\PSC_t$. Each agent in the virtual coordinated system is a passive agent and
agent~$k$ uses the prescription $\PSC^k_t$ to sample an action $A^k_t \sim
\PSC^k_t(L^k_t)$. 

A key insight of \cite{NayyarMahajanTeneketzis_2013} is that the virtual
coordinated system is equivalent to the original multi-agent system in the
following sense.
\begin{theorem}[{\cite{NayyarMahajanTeneketzis_2013}}] \label{thm:equiv}
  Consider a \textup{Dec-POMDP} with a partial history sharing information structure.
  Then, for any policy $\pol = (\pol^k)_{k \in \ALPHABET K}$, where $\pol^k =
  (\pol^k_1, \dots, \pol^k_T)$ for the \textup{Dec-POMDP}, define a coordination policy
  $\psi = (\psi_1, \dots, \psi_T)$ for the virtual coordinated system given by
  \( \psi_t(c_t) = \bigl( \pi^k_t(c_t, \cdot) \bigr)_{k \in \ALPHABET K} \).
  Then, the performance of the virtual coordinated system with policy $\psi$
  is the same as the performance of the \textup{Dec-POMDP} with policy $\pol$.

  Conversely, for any coordination policy $\psi = (\psi_1, \dots, \psi_T)$ for
  the virtual coordinated system, define a policy $\pol = (\pol^k)_{k \in
  \ALPHABET K}$ with $\pol^k = (\pol^k_1, \dots, \pol^k_T)$ for the \textup{Dec-POMDP}
  given by 
  \(
    \pol^k_t(c_t, \ell^k_t) = \psi^k_t(c_t)(\ell^k_t).
  \)
  Then, the performance of the \textup{Dec-POMDP} with policy $\pol$ is the same as
  that of the virtual coordinated system with policy $\psi$.
\end{theorem}

\paragraph{Dynamic program:}
Theorem~\ref{thm:equiv} implies that the problem of finding optimal
decentralized policies in a Dec-POMDP is equivalent to a centralized
(single-agent) problem of finding the optimal coordination policy for the
virtual coordinated system. The virtual coordinated system is a POMDP with
unobserved state $(S_t, L^1_t, \dots, L^K_t)$, observation $C^\new_t$,
and actions $\PSC_t$. The corresponding history of observations is
$(C^\new_{1:t}, \PSC_{1:t-1})$ and therefore we can write a history dependent
dynamic program similar to the one presented in
Proposition~\ref{prop:optimal}. \cite{NayyarMahajanTeneketzis_2013} presented
a simplified dynamic program which used the belief state as an information
state; however, it is clear from the above discussion that any other choice of
information state will also lead to a dynamic programming decomposition.

\subsection{Common-information based AIS and approximate dynamic programming}

Since the coordinated system is a POMDP, we can simply adapt the definition of
AIS Dec-POMDPs and obtain an approximate dynamic program with approximation
guarantees. Let $\PscSp_t$ denote the space of realization of $\PSC_t$.
Then, we have the following.

\begin{definition} \label{def:decais}
  Let $\{\AisSp_t\}_{t=1}^T$ be a pre-specified collection of Banach spaces,
  $\F$ be a function class for \textup{IPMs}, and $\epsilonDelta$ be
  pre-specified positive real numbers. A collection $\{ \ainfo_t \colon (C_t,
    \PSC_{1:t-1}) \mapsto \AisSp_t \}_{t=1}^T$ of history compression functions,
    along with approximate update kernels $\{\nextinfo_t \colon \AisSp_t
      \times \PscSp_t \to
    \Delta(\AisSp_{t+1})\}_{t=1}^T$ and reward approximation functions
    $\{\rewinfo_t \colon \AisSp_t \times \PscSp_t \to \reals\}_{t=1}^T$, is
    called an \emph{$\epsilonDelta$-\ac{AIS} generator} if the process
    $\{\Ais_t\}_{t=1}^T$, where $\Ais_t = \ainfo_t(C_t, \PSC_{1:t-1})$, satisfies the
    following properties:
  \begin{description}
    \item[(DP1)] \textbf{\textup{Sufficient for approximate performance
      evaluation}}, i.e., for any time~$t$, any realization $c_t$ of
      $C_t$ and any choice $\Psc_{1:t}$ of $\PSC_{1:t}$, we have
      \[
        \bigl\lvert \EXP[ \Rew_t \mid C_t = c_t, \PSC_{1:t} = \Psc_{1:t} ] - 
        \rewinfo_t(\ainfo_t(c_t, \Psc_{1:t-1}), \Psc_t) \bigr\rvert 
        \le \varepsilon_t.
      \]
    \item[(DP2)] \textbf{\textup{Sufficient to predict itself approximately}}.
      i.e., for any time~$t$, any realization $c_t$ of $C_t$, any choice
      $\Psc_{1:t}$ of $\PSC_{1:t}$, and for any Borel subset $\ALPHABET B$ of
      $\AisSp_{t+1}$, define
      \(
        \mu_t(\ALPHABET B) \coloneqq \PR(\Ais_{t+1} \in B \mid C_t = c_t,
        \PSC_{1:t} = \Psc_{1:t})
      \)
      and
      \(
        \nu_t(\ALPHABET B) \coloneqq \nextinfo_t(B \mid \ainfo_t(c_t,
        \Psc_{1:t-1}), \Psc_t);
      \)
      then, 
      \[
        d_\F( \mu_t, \nu_t) \le \delta_t.
      \]
  \end{description}
\end{definition}

Similar to Proposition~\ref{prop:alt-info-state}, we can provide an
alternative characterization of an \acs{AIS} where we replace (DP2) with 
approximations of (P2a) and (P2b) and we can prove a proposition similar to 
Proposition~\ref{prop:alt-ais} for the virtual coordinated system.

We can now establish a result similar to Theorem~\ref{thm:ais} that any
\acs{AIS} gives rise to an approximate dynamic program. In this discussion, 
$h_t$ denotes $(c_t, \Psc_{1:t-1})$ and $\HisSp_t$ denotes the space of
realization of $h_t$. 
\begin{theorem}\label{thm:dec_ais}
  Suppose $\{\ainfo_t, \nextinfo_t, \rewinfo_t\}_{t=1}^T$ is an
  $\epsilonDelta$-\acs{AIS} generator.
  Recursively define approximate action-value functions $\{\hat Q_t \colon
    \AisSp_t \times \PscSp_t \to \reals \}_{t=1}^T$ and value functions $\{\hat
    V_t \colon \AisSp_t \to \reals\}_{t=1}^T$ as follows:
   $\hat V_{T+1}(\ais_{T+1}) \coloneqq 0$ and for $t \in \{T, \dots,
  1\}$:
  \begin{subequations}\label{eq:decDP-ais}
  \begin{align}
    \hat Q_t(\ais_t, \Psc_t) &\coloneqq \rewinfo_t(\ais_t, \Psc_t) 
    + \int_{\AisSp_{t+1}} \hat V_{t+1}(\ais_{t+1}) 
    \nextinfo_t(d \ais_{t+1} \mid \ais_t, \Psc_t),
    \\
    \hat V_t(\ais_t) &\coloneqq \max_{\Psc_t \in \PscSp_t} \hat Q_t(\ais_t, \Psc_t).
  \end{align}
  \end{subequations}
  Then, we have the following:
  \begin{enumerate}
    \item \textbf{\textup{Value function approximation:}} For any time~$t$,
      realization~$\his_t$ of $\His_t$, and choice $\Psc_t$ of $\PSC_t$, we have
      \begin{equation}\label{eq:value-approx}
        \lvert Q_t(\his_t, \Psc_t) - \hat Q_t(\ainfo_t(\his_t), \Psc_t)\rvert 
        \le \alpha_t
        \quad\text{and}\quad
        \lvert V_t(\his_t) - \hat V_t(\ainfo_t(\his_t)) \rvert 
        \le \alpha_t,
      \end{equation}
      where
      \[
        \alpha_t = \varepsilon_t + \sum_{\tau=t+1}^{T}\big[  
        \Minkowski_\F(\hat V_{\tau}) \delta_{\tau-1} + \varepsilon_\tau \bigr].
      \]
    \item \textbf{\textup{Approximately optimal policy:}} Let $\hat \psi = (\hat \psi_1, \dots,
      \hat \psi_T)$, where $\hat \psi_t \colon \AisSp_t \to \Delta(\PscSp_t)$,
      be a coordination rule that satisfies 
      \begin{equation}\label{eq:decais-opt}
        \Supp(\hat \psi(\ais_t)) \subseteq 
        \arg \max_{\Psc_t \in \PscSp_t} \hat Q_t(\ais_t, \Psc_t).
      \end{equation}
      Define coordination rule $\psi = (\psi_1, \dots, \psi_T)$, where $\psi_t
      \coloneqq \hat \psi_t \circ \ainfo_t$. Then, for any time~$t$,
      realization~$\his_t$ of $\His_t$, and choice $\Psc_t$ of $\PSC_t$, we
      have
      \begin{equation}\label{eq:coordinationrule-approx}
        \lvert Q_t(\his_t, \Psc_t) - Q^\psi_t(\his_t, \Psc_t)\rvert 
        \le 2\alpha_t
        \quad\text{and}\quad
        \lvert V_t(\his_t) - V^\psi_t(\his_t) \rvert 
        \le 2\alpha_t.
      \end{equation}
  \end{enumerate}
\end{theorem}
\begin{proof}
The proof is similar to the proof of Theorem~\ref{thm:ais}.
\end{proof}

We can extend the approximation result{}s for the virtual coordinated system 
to {\color{black}the approximate policy evaluation case (as in Sec.~\ref{sec:ais-poleval})}, infinite horizon case (as in Sec.~\ref{sec:infinite}), the stochastic \ac{AIS} 
case (as in Sec.~\ref{sec:stochais}), the action compression case (as in Sec.~\ref{sec:actais}), 
and the observation compression case (as in Sec.~\ref{sec:obsais}) in a straightforward manner.

\section{Reinforcement learning for partially observed systems using AIS}
\label{sec:RL}

In this section, we present a policy gradient based reinforcement learning
(RL) algorithm for infinite horizon partially observed systems. The algorithm
learns a time-homogeneous AIS generator $(\ainfo_t, \rewinfo, \nextinfo)$
which satisfies (AP1) and (AP2) or a time-homogeneous AIS generator
$(\ainfo_t, \rewinfo, \aupdate, \nextobs)$ which satisfies (AP1), (AP2a), and
(AP2b). The key idea is to represent each component of the AIS generator using
a parametric family of functions/distributions and use a multi time-scale
stochastic gradient descent algorithm \citep{Borkar_1997} which learns AIS
generator at a faster time-scale than the policy and/or the action-value
function.

Then, for the ease of exposition, we first assume that the policy is fixed and
describe how to learn the AIS generator using stochastic gradient descent.
To specify an \acs{AIS}, we must pick an IPM $\F$ as well. Although, in
principle, we can choose any IPM, in practice, we want to choose an IPM such
that the distance $d_\F(\mu_t, \nu_t)$ in (AP2) or (AP2b) can be computed
efficiently. We discuss the choice of IPMs in Sec.~\ref{sec:learn-IPM} and
then discuss the stochastic gradient descent algorithm to learn the
AIS-generator for a fixed policy in Sec.~\ref{sec:grad-ascent}.
Then we describe how to simultaneously learn the AIS generator and the policy
using a multi-time scale algorithm, first for an actor only framework and then for an actor-critic framework in Sec.~\ref{sec:PORL}.

\subsection{The choice of an IPM} \label{sec:learn-IPM}

As we will explain in the next section in detail, our general modus operandi
is to assume that the stochastic kernel $\nextinfo$ or $\nextobs$ that we are
trying to learn belongs to a parametric family and then update the parameters
of the distribution to either minimize $d_\F(\mu, \nu)$ defined in (AP2) or
minimize $d_\F(\mu^\ob, \nu^\ob)$ defined in (AP2b). Just to keep the
discussion concrete, we focus on (AP2). Similar arguments apply to (AP2b)
as well. First note that for a particular choice of parameters, we know the
distribution $\nu$ in closed form, but we do not know the distribution $\mu$
in closed form and only have samples from that distribution. One way to
estimate the IPM between a distribution and samples from another distribution
is to use duality and minimize $\bigl| \int_{\AisSp} f d\mu - \int_{\AisSp} f
d\nu \bigr|$ over the choice of function $f$ such that $f \in \F$. When $d_\F$
is equal to the total variation distance or the Wasserstein distance, this
optimization problem may be solved using a linear program
\citep{Sriperumbudur2012}. However, solving a linear program at each step of
the stochastic gradient descent algorithm can become a computational
bottleneck. We propose two alternatives here. The first is to use the total
variation distance or the Wasserstein distance but instead of directly working
with them, we use a KL divergence based upper bound as a surrogate loss. The
other alternative is to work with RKHS-based MMD (maximum mean
discrepancy) distance, which can be computed from samples without solving an
optimization problem \citep{Sriperumbudur2012}. It turns out that for the
\acs{AIS}-setup, a specific form of MMD known as distance-based MMD is particularly convenient as we explain below.

\paragraph{KL-divergence based upper bound for total variation or Wasserstein
distance.}

Recall that the KL-divergence between two densities $\mu$ and $\nu$  on
$\Delta(\ALPHABET X)$ is defined as
\[
  D_{\textup{KL}}(\mu \| \nu) =
  \int_{\ALPHABET X} \log \mu(x)\mu(dx)  -
  \int_{\ALPHABET X}  \log \nu(x) \mu(dx).
\]
The total variation distance can be upper bounded by the KL-divergence using
Pinsker's inequality \citep{csiszar2011} (see footnote~\ref{fnt:TV}
  for the difference in constant factor from the standard Pinsker's
inequality):
\begin{equation}\label{eq:pinsker}
  d_{\textup{TV}}(\mu, \nu) \le
  \sqrt{ 2 D_{\textup{KL}}(\mu \| \nu) }.
\end{equation}

As we will explain in the next section, we consider the setup where
we know the distribution $\nu$ but only obtain samples from the distribution
$\mu$. Since there are two losses---the reward prediction loss $\varepsilon$
and the \acs{AIS}/observation prediction loss $\delta$, we work with
minimizing the weighted square average $\lambda \varepsilon^2 + (1-\lambda)
\delta^2$, where $\lambda \in [0,1]$ is a hyper-parameter. Pinsker's
inequality~\eqref{eq:pinsker} suggests that instead of
$d_{\textup{TV}}(\mu,\nu)^2$, we can use the surrogate loss function
\[
  \int_{\ALPHABET X} \log \nu(x) \mu(dx)
\]
where we have dropped the term that does not depend on $\nu$. Note that the
above expression is the same as the cross-entropy between $\mu$ and $\nu$
which can be efficiently computed from samples. In particular, if we get $T$
i.i.d.\@ samples $X_1, \dots, X_T$ from $\mu$, then
\begin{equation}\label{eq:surrogate-KL}
  \frac 1T \sum_{t=1}^T \log \nu(X_t)
\end{equation}
is an unbiased estimator of $\int_{\ALPHABET X} \log \nu(x) \mu(dx)$.

Finally, if $\ALPHABET X$ is a bounded space with diameter~$D$, then 
\[
  d_{\textup{Wass}}(\mu, \nu) \le D d_{\textup{TV}}(\mu,\nu).
\]
So, using cross-entropy as a surrogate loss also works for Wasserstein
distance.

\paragraph{Distance-based MMD.}

The key idea behind using a distance-based MMD is the following results.

\begin{proposition}[Theorem 22 of \cite{Sejdinovic2013}] 
  \label{prop:RKHS}
  Let $\ALPHABET X \subseteq \reals^m$ and $d_{\ALPHABET X,p} \colon \ALPHABET
  X \times \ALPHABET X \to \reals_{\ge 0}$ be a metric given by $d_{\ALPHABET
  X, p}(x,x') = \| x - x'\|_2^{p}$, for $p \in (0, 2]$. Let $k_p \colon
  \ALPHABET X \times \ALPHABET X \to \reals$ be any kernel given
  \[
    k_p(x,x') = \tfrac 12\bigl[ 
      d_{\ALPHABET X, p}(x,x_0) + 
      d_{\ALPHABET X, p}(x',x_0) -
      d_{\ALPHABET X, p}(x, x')
    \bigr],
  \]
  where $x_0 \in \ALPHABET X$ is arbitrary, 
  and let $\mathcal H_p$ be a \textup{RKHS} with kernel $k_p$ 
  and $\F_p = \{ f \in \mathcal H_p : \| f \|_{\mathcal H_p} \le
  1\}$. Then, for any distributions $\mu, \nu \in \Delta(\ALPHABET X)$, the
  \textup{IPM} $d_{\F_p}(\mu,\nu)$ can be expressed as follows:
  \begin{equation}\label{eq:RKHS}
    d_{\F_p}(\mu, \nu) = \sqrt{\EXP[ d_{\ALPHABET X,p}(X,W) ] - 
      \tfrac{1}{2} \EXP[ d_{\ALPHABET X,p}(X,X') ] -
    \tfrac{1}{2} \EXP[ d_{\ALPHABET X,p}(W,W') ]},
  \end{equation}
  where $X,X' \sim \mu$, $W, W' \sim \nu$ and $(X,X',W,W')$ are all
  independent. 
\end{proposition}
We call $d_p$ defined above as a \emph{distance-based} MMD. For $p=1$
(for which $d_{\ALPHABET X}$ corresponds to the $L_2$ distance), the
expression inside the square root in~\eqref{eq:RKHS} is called the Energy
distance in the statistics literature~\citep{Szekely2004}. In
\cite{Sejdinovic2013}, the above result is stated for a general semimetric of
a negative type. Our statement of the above result is specialized to the
semimetric $d_{\ALPHABET X, p}$. See Proposition~3 and Example~15 of
\cite{Sejdinovic2013} for details.

As explained in the previous section, we work with
minimizing the weighted square average $\lambda \varepsilon^2 + (1-\lambda)
\delta^2$, where $\lambda$ is a hyper-parameter. Proposition~\ref{prop:RKHS}
suggests that instead of $d_{\F_p}(\mu, \nu)^2$, we can use a surrogate loss
function
\begin{equation}\label{eq:surrogate-general-0}
  \int_{\ALPHABET X} \int_{\ALPHABET X} \| x - w \|_2^p\; \mu(dx) \nu(dw) 
  - \frac12
  \int_{\ALPHABET X} \int_{\ALPHABET X} \| w - w' \|_{2}^p\; \nu(dw) \nu(dw')
\end{equation}
for $p \in (0, 2]$, where we have dropped the term that does not depend on
$\nu$. It is possible to compute the surrogate loss efficiently from samples
as described in \cite{Sriperumbudur2012}. In particular, if we get $T$
i.i.d.\@ samples $X_1, \dots, X_T$ from $\mu$, then
\begin{equation}\label{eq:surrogate-general}
  \frac 1T \sum_{t=1}^T \int_{\ALPHABET X} \| X_t - w \|_2^p\; \nu(dw) 
  - \frac12
  \int_{\ALPHABET X} \int_{\ALPHABET X} \| w - w' \|_{2}^p\; \nu(dw) \nu(dw')
\end{equation}
is an unbiased estimator of~\eqref{eq:surrogate-general-0}.

In our numerical experiments, we use the surrogate
loss~\eqref{eq:surrogate-general} for $p=2$, which simplifies as follows.
\begin{proposition}\label{prop:surrogate}
  Consider the setup of \textup{Proposition~\ref{prop:RKHS}} for $p=2$.
  Suppose $\nu_\xi$ is a known parameterized distribution with mean $M_\xi$
  and $X$ is a sample from $\mu$. Then, the gradient of 
  \begin{equation}\label{eq:surrogate}
    (M_\xi - 2 X) ^\TRANS M_\xi
  \end{equation}
  with respect to $\xi$ 
  in an unbiased estimator of $\GRAD_{\xi} d_{\F_2}(\mu, \nu_\xi)^2$. 
\end{proposition}
\begin{proof}
  For $p=2$, we have that
  \[
    d_{\F_2}(\mu, \nu_\xi)^2 = \EXP[\| X - W \|_2^2 ] - 
      \tfrac{1}{2} \EXP[\| X - X'\|_2^2 ] -
    \tfrac{1}{2} \EXP[ \| W - W'\|_2^2 ],
  \]
  where $X,X' \sim \mu$ and $W,W' \sim \nu_\xi$. Simplifying the right hand
  side, we get that
  \[
    d_{\F_2}(\mu, \nu_\xi)^2 = 
    \|\EXP[  X ]\|_2^2  - 2 \EXP [ X ]^\TRANS \EXP [ W ] +
    \| \EXP[  W  ] \|_2^2.
  \]
  Note that the term $\|\EXP[ X ]\|_2^2 $ does not depend on the
  distribution~$\nu_\xi$. Thus, the expression~\eqref{eq:surrogate} captures
  all the terms which depend on $\xi$.
\end{proof}

The implication of Proposition~\ref{prop:surrogate} is if we use MMD with the
RKHS $\mathcal H_2$ defined in Proposition~\ref{prop:RKHS}, then we can can
use the expression in~\eqref{eq:surrogate} as a surrogate loss function for
$d_{\F_2}(\mu, \nu_\xi)^2$. 

Now we show how to compute the surrogate loss~\eqref{eq:surrogate} for two
types of parameterized distributions $\nu_\xi$.
\begin{enumerate}
  \item \textsc{Surrogate loss for predicting discrete variables:}
     When predicting a discrete-valued
    random variable, say a discrete-valued \acs{AIS} $\Ais_{t+1}$ in (AP2) or
    a discrete-valued observation $\Ob_t$ in (AP2b), we view the discrete
    random variable as a continuous-valued random vector by representing it as
    a one-hot encoded vector. In particular, if the discrete random variable,
    which we denote by $V$, takes $m$ values, then its one-hot encoded
    representation, which we denote by $X$, takes values in the corner points
    of the simplex on $\reals^m$. Now, suppose $\nu_\xi$ is any parameterized
    distribution on the discrete set $\{1, \dots, m\}$ (e.g., the softmax
    distribution). Then, in the one-hot encoded representation, the mean
    $M_\xi$ is given by
    \[
      M_\xi = \sum_{i=1}^m \nu_\xi(i) e_i = 
      \begin{bmatrix}
        \nu_\xi(1) \\ \vdots \\ \nu_\xi(m) 
      \end{bmatrix},
    \]
    where $e_i$ denotes the $m$-dimensional unit vector with $1$ in the
    $i$-th location. Thus, when we one-hot encode discrete \acs{AIS} or
    discrete observations, the ``mean'' $M_\xi$ is same as the probability
    mass function (PMF) $\nu_\xi$. Thus, effectively, $d_{\F_2}(\mu,\nu)^2$ is
    equivalent to $\| \mu - \nu \|_2^2$ and~\eqref{eq:surrogate} is an
    unbiased estimator where we have removed the terms that do not depend on~$\nu$. 

  \item \textsc{Surrogate loss for predicting continuous variables:}
    When predicting a continuous-valued random variable, say a
    continuous-valued \acs{AIS} $\Ais_{t+1}$ in (AP2) or a continuous-valued
    observation $\Ob_t$ in (AP2b), we can immediately use the surrogate
    loss~\eqref{eq:surrogate} as long as the parameterized distribution
    $\nu_\xi$ is such that its mean $M_\xi$ is given in closed form. Note that
    the surrogate loss~\eqref{eq:surrogate} only depends on the mean of the
    distribution and not one any other moment. So, any two distributions $\nu$
    and $\nu'$ that have the same mean, the surrogate loss between any
    distribution $\mu$ and $\nu$ is same as the surrogate loss between $\mu$
    and $\nu'$.  Thus, using the surrogate loss~\eqref{eq:surrogate} for
    predicting continuous variables only makes sense when we expect the true
    distribution to be close to a deterministic function.
\end{enumerate}

\subsection{Learning an AIS for a fixed policy} \label{sec:grad-ascent}

The definition of \acs{AIS} suggests that there are two ways to construct an
information state from data: we either learn a time-homogeneous
\acs{AIS}-generator $(\ainfo, \rewinfo, \nextinfo)$ that satisfies (AP1) and
(AP2) or we learn a time-homogeneous \acs{AIS}-generator $(\ainfo, \rewinfo,
\aupdate, \nextobs)$ that satisfies (AP1), (AP2a), and (AP2b). In either case,
there are three types of components of \acs{AIS}-generators: (i)~regular
functions such as $\rewinfo$ and $\aupdate$; (ii)~history compression
functions $\{ \ainfo_t \}_{t \ge 1}$; and (iii)~stochastic kernels $\nextinfo$
and $\nextobs$. To learn these components from data, we must choose
parametric class of functions for all of these. In this section, we do not
make any assumption about how these components are chosen. In particular,
$\rewinfo$ and $\aupdate$ could be represented by any class of function
approximators (such as a multi-layer preceptron); $\ainfo$ 
could be represented by any class of time-series approximators (such as a
RNN or its refinements such as LSTM or GRU); and $\nextinfo$ and $\nextobs$
could be represented by any class of stochastic kernel approximators (such as
softmax distribution or mixture of Gaussians).  We use $\xi_t$ to denote the
corresponding parameters. 

There are two losses in the definition of an \acs{AIS}: the reward loss $|R_t
- \rewinfo(\ais_t, \act_t) |$ and the prediction loss $d_\F(\mu_t, \nu_t)$ or
$d_\F(\mu^\ob_t, \nu^\ob_t)$. We combine these into a single criterion and
minimize the combined loss function
\[
  \frac{1}{T} \sum_{t=1}^T \Bigl[ \lambda | R_t - \rewinfo(\Ais_t, \Act_t) |^2 
    + (1-\lambda) d_\F(\mu_t, \nu_t)^2
\Bigr]
\]
where $T$ is the length of the episode or the rollout horizon and $\lambda \in
[0, 1]$ may be viewed as a hyper-parameter. 

As described in Section~\ref{sec:learn-IPM}, there are two possibilities to
efficiently compute $d_\F(\mu_t, \nu_t)^2$: use total-variation distance or
Wasserstein distance as the IPM and use surrogate
loss~\eqref{eq:surrogate-KL}; or use distance-based MMD as the IPM and use
the surrogate loss~\eqref{eq:surrogate}.

In particular, to choose an AIS that satisfies (AP1) and (AP2), we either
minimize the surrogate loss
\begin{equation}\label{eq:loss-KL-1}
  \frac 1T \sum_{t=1}^T \big[ \lambda | R_t - \hat r(\Ais_t, \Act_t)|^2 + 
  (1-\lambda) \log(\nu_t(\Ais_{t+1})) \bigr]
\end{equation}
or we minimize the surrogate loss (specialized for $p=2$)
\begin{equation} \label{eq:loss-1}
  \frac{1}{T} \sum_{t=1}^T \Bigl[ \lambda | R_t - \rewinfo(\Ais_t, \Act_t) |^2
    + (1-\lambda) (M_{t} - 2 \Ais_{t+1})^\TRANS M_{t}
  \Bigr]
\end{equation}
where $M_t$ is the mean of the distribution $\nu_t$.

Similarly, in order to choose an \acs{AIS} that satisfies (AP1), (AP2a) and
(AP2b), we minimize the surrogate loss
\begin{equation}\label{eq:loss-KL-2}
  \frac 1T \sum_{t=1}^T \big[ \lambda | R_t - \hat r(\Ais_t, \Act_t)|^2 + 
  (1-\lambda) \log(\nu^\ob_t(\Ob_{t})) \bigr]
\end{equation}
or we minimize the surrogate loss (specialized for $p=2$)
\begin{equation}\label{eq:loss-2}
  \frac{1}{T} \sum_{t=1}^T \Bigl[ \lambda | R_t - \rewinfo(\Ais_t, \Act_t) |^2
    + (1-\lambda) (M^\ob_{t} - 2 \Ob_t)^\TRANS M^\ob_{t}
  \Bigr]
\end{equation}
where $M^\ob_{t}$ is the mean of the distribution $\nu^\ob_{t}$. 

We use $\bar \xi$ to denote the parameters of the \acs{AIS}-generator, i.e.,
the parameters of $(\ainfo, \nextinfo, \rewinfo)$ when using (AP1) and (AP2)
or the parameters of $(\ainfo, \aupdate, \nextobs, \rewinfo)$ when using
(AP1), (AP2a), (AP2b). We then use $\mathcal L(\bar \xi)$ to the denote the
corresponding loss~\eqref{eq:loss-KL-1}, \eqref{eq:loss-1},
\eqref{eq:loss-KL-2}, or~\eqref{eq:loss-2}. Then, we can learn the
parameters $\bar \xi$ using stochastic gradient descent:
\begin{equation}\label{sgd:ais}
  \bar \xi_{k+1} = \bar \xi_k - a_k \GRAD_{\bar \xi} \mathcal L(\bar \xi_k),
\end{equation}
where the learning rates $\{a_k\}_{k \ge 0}$ satisfy the standard conditions
$\sum a_k = \infty$ and $\sum a_k^2 < \infty$. 

\subsection{AIS-based PORL}\label{sec:PORL}

Given the stochastic gradient descent algorithm to learn an
\acs{AIS}-generator for a fixed policy, we can simultaneously learn a policy
and \acs{AIS}-generator by following a multi time-scale stochastic gradient
descent \citep{Borkar_1997}, where we learn the \acs{AIS}-generator at a faster
learning rate than the policy. 

In particular, let $\pol_\theta \colon \AisSp \to \Delta(\ActSp)$ be a
parameterized stochastic policy with parameters~$\theta$. Let $J(\bar \xi,
\theta)$ denote the performance of policy $\pol_\theta$. From the policy
gradient theorem \citep{Sutton2000, Baxter2001}, we know that 
\begin{equation}
  \GRAD_{\theta} J(\bar \xi, \theta) = 
  \EXP \biggl[ \sum_{t=1}^\infty \biggl( \sum_{\tau = 1}^t 
    \GRAD_{\theta} \log \pol_\theta(\Act_t \mid \Ais_t) \biggr)
    \discount^{t-1} \Rew_t \biggr]
\end{equation}
which can be estimated from a sampled trajectory with a rollout horizon of $T$ using the G(PO)MDP gradient \citep{Baxter2001}
\begin{equation} \label{eq:GPOMDP_update}
  \widehat \GRAD_{\theta} J(\bar \xi, \theta) = 
   \sum_{t=1}^T \biggl( \sum_{\tau = 1}^t 
    \GRAD_{\theta} \log \pol_\theta(\Act_t \mid \Ais_t) \biggr)
    \discount^{t-1} \Rew_t .
\end{equation}

We can iteratively update the parameters $\{ (\bar \xi_k, \theta_k) \}_{k \ge
1}$ of both the \acs{AIS}-generator and policy as follows. We start with an
initial choice $(\bar \xi_1, \theta_1)$, update both parameters after a
rollout of $T$ as follows
\begin{equation}\label{eq:ais_a}
  \bar \xi_{k+1} = \bar \xi_k - a_k \GRAD_{\bar \xi} \mathcal L(\bar \xi_k)
  \quad\text{and}\quad
  \theta_{k+1} = \theta_k + b_k   \widehat \GRAD_{\theta} J(\bar \xi_k, \theta_k)
\end{equation}
where the learning rates $\{a_k\}_{k \ge 1}$ and $\{b_k\}_{k \ge 1}$ satisfy
the standard conditions on multi time-scale learning: $\sum_{k} a_k = \infty$,
$\sum_k b_k = \infty$, $\sum_{k} a_k^2 < \infty$, $\sum_{k} b_k^2 < \infty$,
and $\lim_{k \to \infty} b_k/a_k = 0$, which ensures that \ac{AIS}-generator
learns at a faster rate than the policy. 

A similar idea can be used for an actor-critic algorithm. Suppose we have a
parameterized policy $\pol_\theta \colon \AisSp \to \Delta(\ActSp)$ and a
parameterized critic $\hat Q_{\zeta} \colon \AisSp \times \ActSp \to \reals$,
where $\theta$ denotes the parameters of the policy and $\zeta$ denotes the
parameters of the critic. Let $J(\bar \xi, \theta, \zeta)$ denote the
performance of the policy. From the policy gradient theorem \citep{Sutton2000,
KondaTsitsiklis2003}, we know that 
\begin{equation}
  \GRAD_\theta J(\bar \xi, \theta, \zeta) = \frac{1}{1-\discount}
  \EXP\bigl[ \GRAD_\theta \log \pol_\theta(\Act_t \mid \Ais_t) 
  Q_\zeta(\Ais_t, \Act_t) \bigr]
\end{equation}
which can be estimated from a sampled trajectory with a rollout horizon of $T$
by
\begin{equation}
  \widehat \GRAD_\theta J(\bar \xi, \theta, \zeta) = \frac{1}{(1-\discount)T}
  \sum_{t=1}^T \GRAD_\theta \log \pol_\theta(\Act_t \mid \Ais_t) 
  \hat Q_\zeta(\Ais_t, \Act_t).
\end{equation}

For the critic, we use the temporal difference loss 
\begin{equation}
  \mathcal L_{\textup{TD}}(\bar \xi, \theta, \zeta) = \frac{1}{T}
  \sum_{t=1}^T \texttt{smoothL1}\bigl( 
    \hat Q_\zeta(\Ais_t, \Act_t) - R_t - \discount
    \hat Q_\zeta(\Ais_{t+1}, \Act_{t+1}) \bigr)
\end{equation}
where \texttt{smoothL1} is the smooth $L_1$ distance given by
\[
\texttt{smoothL1}(x) = \begin{cases}
  \frac12 x^2 \quad &\text{if } |x| < 1 \\
  |x| - \frac12 &\text{otherwise}.
\end{cases}
\]

We can iteratively update the parameters $\{ (\bar \xi_k, \theta_k, \zeta_k)
\}_{k \ge 1}$ of the \acs{AIS}-generator, policy, and critic as follows. We start
with an initial choice $(\bar \xi_1, \theta_1, \zeta_1)$, and update all the parameters after a rollout of $T$ steps as follows
\begin{equation}\label{eq:ais_ac}
  \bar \xi_{k+1} = \bar \xi_k - a_k \GRAD_{\bar \xi} \mathcal L(\bar \xi_k)
  , \quad
  \theta_{k+1} = \theta_k + b_k   \widehat \GRAD_{\theta} J(\bar \xi_k, \theta_k, \zeta_k)
  \quad\text{and}\quad
  \zeta_{k+1} = \zeta_k - c_k \GRAD_{\zeta} \mathcal L_{\textup{TD}}(\bar \xi_k,
  \theta_k, \zeta_k)
\end{equation}
where the learning rates $\{a_k\}_{k \ge 1}$,  $\{b_k\}_{k \ge 1}$,
$\{c_k\}_{k \ge 1}$ satisfy
the standard conditions on multi time-scale learning: $\sum_{k} a_k = \infty$,
$\sum_k b_k = \infty$, $\sum_k c_k = \infty$, $\sum_{k} a_k^2 < \infty$,
$\sum_{k} b_k^2 < \infty$, $\sum_k c_k^2 < \infty$, 
$\lim_{k \to \infty} c_k/a_k = 0$, and $\lim_{k \to \infty} b_k/c_k = 0$,
which ensures that \ac{AIS}-generator learns at a faster rate than the critic,
and the critic learns at a faster rate than the policy. \blue{The complete algorithm
is shown in Algorithm~\ref{alg:PORL}.}

\begin{algorithm}[!t]
  {\color{black}
    \caption{AIS-based PORL algorithm}\label{alg:PORL}
    \textbf{Input:} Initial AIS-Generator: $(\ainfo, \nextinfo, \rewinfo)_{\bar \xi_0}$, 
    Initial Policy: $\pol_{\polPars_0}$, 
    Discount factor: $\discount$, 
    \\
    \phantom{\textbf{Input:}}
    Reward weight: $\lambda$, 
    Number of episodes: $K$, 
    AIS-LR: $a_{k=1}^K$,
    Policy-LR: $b_{k=1}^K$.
    \\
    \textbf{Output:} 
    Learned policy: $\pol_{\polPars_K}$, 
    Learned AIS-generator: $(\ainfo, \nextinfo, \rewinfo)_{\bar \xi_K}$
    \begin{algorithmic}[1]
    \Procedure{AIS-based PORL}{}
    \ForAll {$k \in \{1, \dots, K\}$}
	    \State Reset  environment and perform an episode using 
        $\pol_{\polPars_{k-1}}, (\ainfo, \nextinfo, \rewinfo)_{\bar \xi_{k-1}}$.
        \State $\Act_{1:T}, \Ob_{1:T}, \Rew_{1:T} \gets$ Actions, observations, and
        rewards for episode~$k$. 
	    \State Compute AIS loss using $\Act_{1:T}, \Ob_{1:T}, \Rew_{1:T},
        \lambda, (\ainfo, \nextinfo, \rewinfo)_{\bar \xi_{k-1}}$ 
        using Eq.~\eqref{eq:loss-KL-2} or~\eqref{eq:loss-2}
        % (depending on the IPM)
	    \State Compute policy loss using $\Act_{1:T}, \Ob_{1:T}, \Rew_{1:T},
        \discount, \pol_{\polPars_{k-1}}, (\ainfo)_{\bar \xi_{k-1}}$ using
        Eq.~\eqref{eq:GPOMDP_update}
	    \State Update AIS parameters $\bar \xi_{k-1}$ and policy
        parameters $\pol_{\polPars_{k-1}}$ using Eq.~\eqref{eq:ais_a}
    \EndFor \EndProcedure
    \end{algorithmic}
}
\end{algorithm}

Under standard technical conditions (see Theorem~23 of \cite{Borkar_1997} or
Page~35 of \cite{Leslie_2004}), we can show that
iterations~\eqref{eq:ais_a} and~\eqref{eq:ais_ac} will converge to a
stationary point of the corresponding ODE limits. At convergence, depending on
$\varepsilon$ and $\delta$ for the quality of \acs{AIS} approximation, we can
obtain approximation guarantees corresponding to Theorem~\ref{thm:inf-ais}.
\blue{For a more detailed discussion on convergence, please refer to
Appendix~\ref{sec:porl-convergence}.}

We conclude this discussion by mentioning that algorithms similar to the
\acs{AIS}-based PORL have been proposed in the literature including
\cite{bakker2002reinforcement,WierstraFoersterPetersSchmidhuber_2007,
WierstraFoersterPetersSchmidhuber_2010,HausknechtStone_2015,HeessHuntLillicrapSilver_2015,Zhu2017,ha2018world,BaiseroAmato_2018,Igl2018,Zhang2019}.
However, these papers only discuss the empirical performance of the proposed
algorithms but do not derive performance bounds.

\section{Experiments}\label{sec:experiments}

We perform numerical experiments to check the effectiveness of AIS-based PORL
algorithms proposed in the previous section. The code for all AIS experiments is available in \cite{ais_git_repo}. We consider three classes of
POMDP environments, which have increasing difficulty in terms of the dimension
of their state and observation spaces:
\begin{enumerate}
  \item Low-dimensional environments (Tiger, Voicemail, and Cheese Maze)
  \item Moderate-dimensional environments (Rock Sampling and Drone Surveillance)
  \item High-dimensional environments (different variations of MiniGrid)
\end{enumerate}
For each environment, we use the actor only framework and learn an AIS based
on (AP1), (AP2a), and (AP2b). There are four components
of the corresponding \acs{AIS}-generator: the history compression function
$\ainfo$, the \acs{AIS} update function $\aupdate$, the reward prediction
function $\rewinfo$, and the observation prediction kernel $\nextobs$. 
We model the $\ainfo$ as an LSTM, where the memory update unit of LSTM acts as
$\aupdate$. We model $\rewinfo$, $\nextobs$, and the policy $\apol$ as
feed-forward neural networks. A block diagram of the network architecture is
shown in Fig.~\ref{fig:network} and the details of the networks and the
hyperparameters are presented in
Appendix~\ref{sec:network}. To avoid over-fitting, we use the same network
architecture and hyperparameters for all environments in the same difficulty
class. 

\begin{figure}[!t!b]
  \centering
  \includegraphics{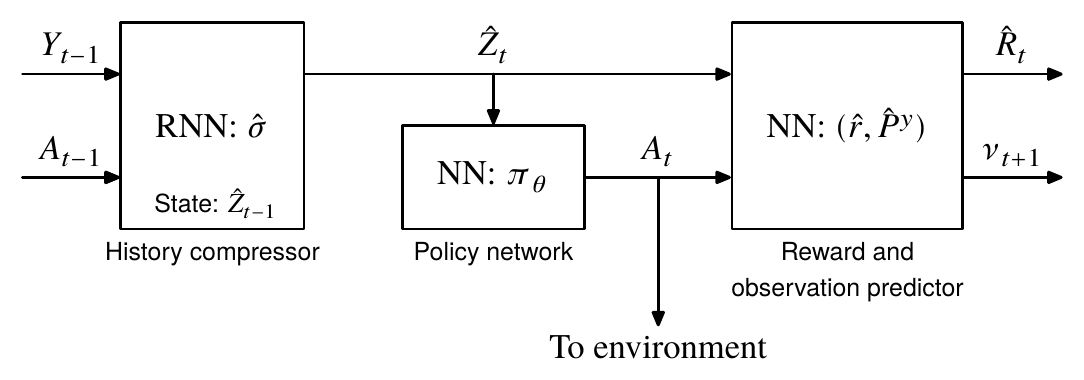}
  \caption{Network architecture for PORL using AIS.}
  \label{fig:network}
\end{figure}

We repeat each experiment for multiple random seeds and plot the median value
along with the uncertainty band from the first to the third quartile. For all
environments, we compare our performance with a baseline which uses an
actor-critic algorithm where both the actor and critic are modeled using LSTM
and the policy parameters are updated using PPO. This architecture was
proposed as a baseline for the Minigrid environments in
\cite{chevalier2018babyai}. The details of the baseline architecture are
presented in Appendix~\ref{sec:network}. 

To evaluate the performance of the policy while training for AIS-based PORL, a
separate set of rollouts is carried out at fixed intervals of time steps and
the mean of these rollouts is considered. For the PPO baseline a number of
parallel actors are used during training, and once the episodes are completed,
their returns are stored in a list. A fixed number (based on the number of parallel actors) of past episodes are
considered to evaluate the mean performance of the current policy during
training. See Appendix~\ref{sec:network} for details.

For the low and moderate dimensional environments, we compare the performance with
the best performing planning solution obtained from the JuliaPOMDP repository
\citep{JuliaPOMDPs}. For the high-dimensional environments, finding a planning solution is intractable, so we only compare with the PPO baseline mentioned previously.

\subsection{Low-dimensional environments}
In these POMDP environments, the size of the unobserved state space is less
than about 10 and the planning solution can be easily obtained using standard
POMDP solvers. 

\begin{figure}[!thb]
    \centering
    \begin{subfigure}[t]{0.475\textwidth}
      \includegraphics[width=\textwidth]{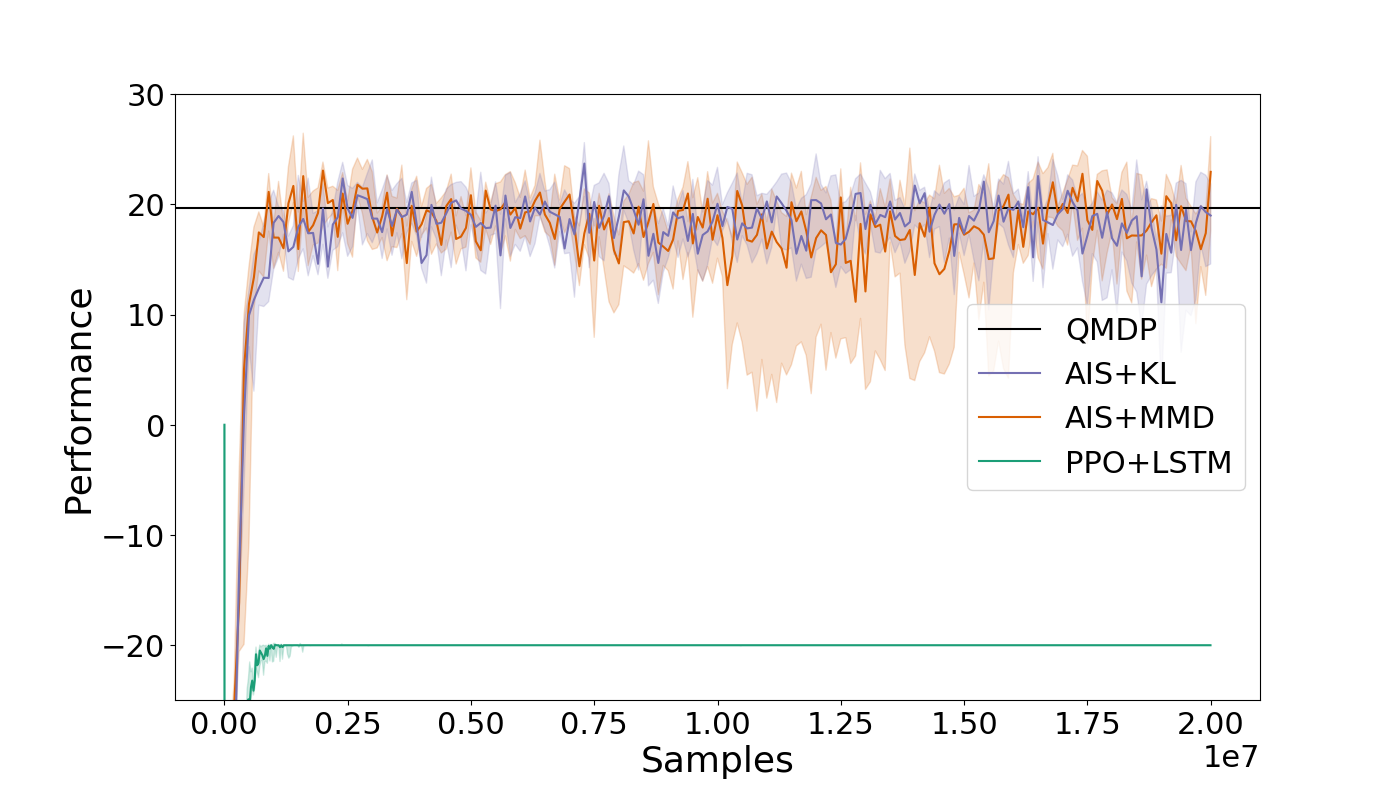}
        \caption{Tiger}
    \end{subfigure}%
    \hfill
    \begin{subfigure}[t]{0.475\textwidth}
      \includegraphics[width=\textwidth]{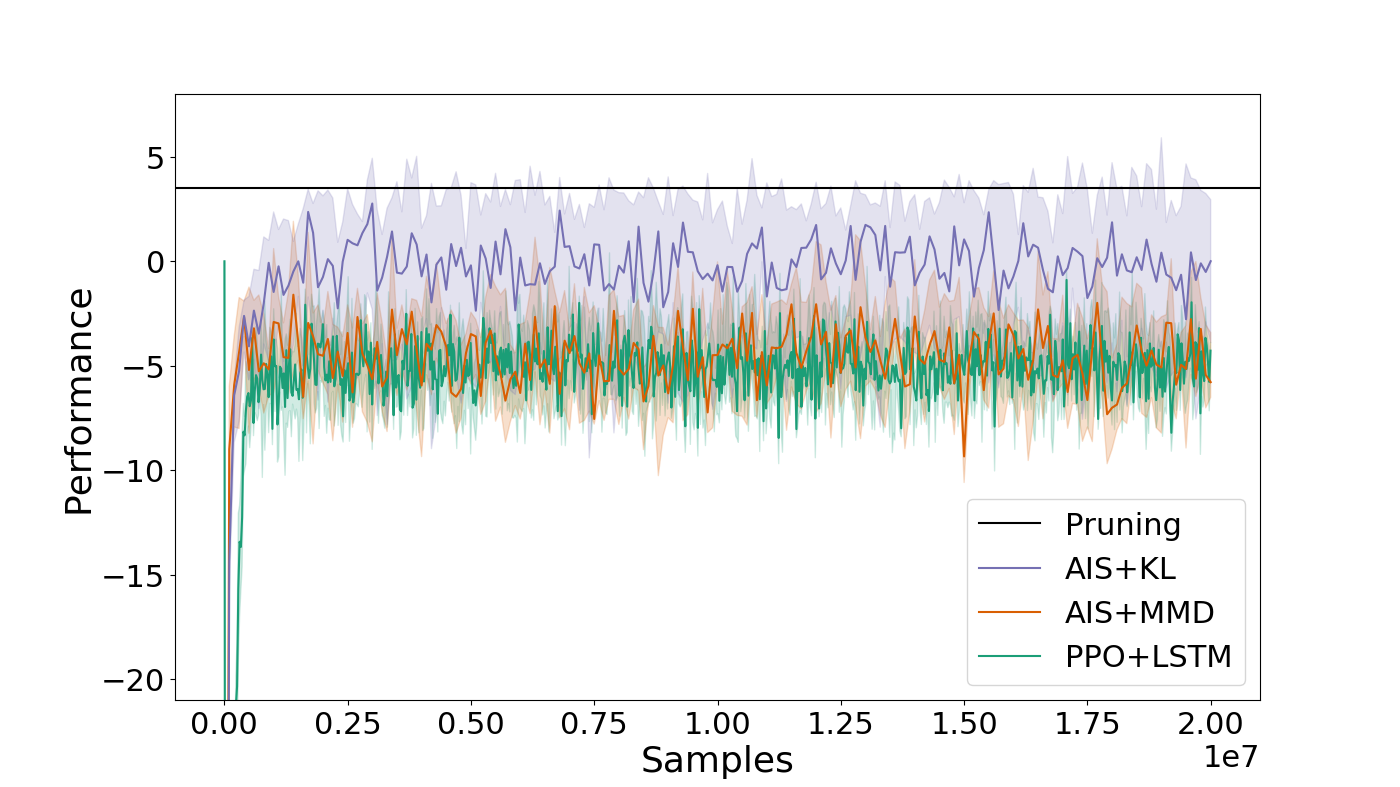}
        \caption{Voicemail}
    \end{subfigure}%

    \begin{subfigure}[t]{0.475\textwidth}
      \includegraphics[width=\textwidth]{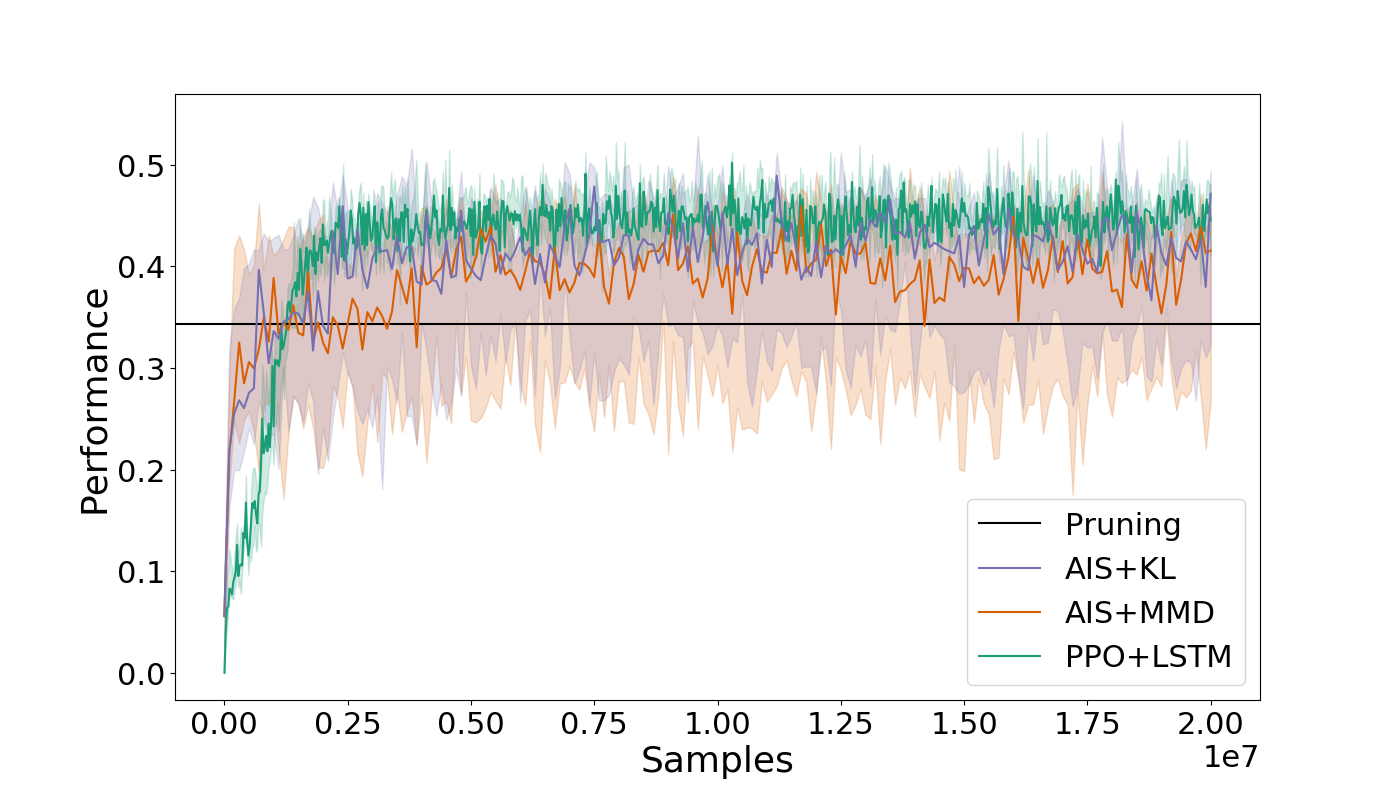}
        \caption{Cheese Maze}
    \end{subfigure}
    \caption{Comparison of AIS-based actor only PORL algorithm with LSTM+PPO
    baseline for low-dimensional environments (for 10 random seeds).} 
    \label{fig:low}
\end{figure}

\begin{enumerate}
  \item \textsc{Tiger:} The Tiger environment is a sequential
    hypothesis testing task proposed in \cite{KaelblingLittmanCassandra_1998}.
    The environment consists of two doors, with a tiger behind one door and a
    treasure behind the other. The agent can either perform a \textsc{listen}
    action, which has a small negative reward of $-1$ and gives a noisy
    observation about the location of the tiger, or the agent can open one of
    the doors. Opening the door with the treasure gives a reward of $+10$
    while opening the door with a tiger gives a large negative reward of
    $-100$. After opening a door, the environment is reset.

  \item \textsc{Voicemail:} The Voicemail enviroment is also a sequential
    hypothesis testing task propsed in \cite{WilliamsYoung_2007}. This
    environment models a dialog system for managing voicemails. The agent can
    either perform an \textsc{ask} action, wich has a small negative reward of
    $-1$ and gives a noisy observation about the intent of the user, or the
    agent can execute \textsc{save} or \textsc{delete}. Choosing a
    \textsc{save}/\textsc{delete} action which matches the intent of the user
    gives a reward of $+5$. The agent receives a negative reward of $-20$ for 
    action \textsc{delete} when the user intent is 
    \textsc{save}, while choosing action \textsc{save} when the user intent 
    is \textsc{delete} gives a smaller but still significant negative reward
    of $-10$. Since the user prefers  \textsc{save} more than \textsc{delete}, 
    the initial belief is given by $[0.65,0.35]$ for \textsc{save} and
    \textsc{delete} respectively. After taking a \textsc{save}/\textsc{delete}
    action, the agent moves on to the next voicemail message.

    % https://tex.stackexchange.com/questions/232105/wrapfigure-in-an-enumerate-environment
 \item
    \parbox[t]{\dimexpr\textwidth-\leftmargin}{%
      \vspace{-2.5mm}
      \begin{wrapfigure}[7]{r}{0.25\textwidth}
        \centering
        \vspace{-1.5\baselineskip}
        \resizebox{0.9\linewidth}{!}{%
          \begin{mpost}[mpsettings={input metafun;}]
            ux := 0.75cm;
            uy := 0.75cm;

          save box;
          path box;

          box := fullsquare xysized (ux, uy) ;

          % Top row
          draw box; 
          label(\btex $1$ etex, origin);

          draw box shifted (1ux, 0); 
          label(\btex $2$ etex, (1ux, 0));

          draw box shifted (2ux, 0); 
          label(\btex $3$ etex, (2ux, 0));

          draw box shifted (3ux, 0); 
          label(\btex $2$ etex, (3ux, 0));

          draw box shifted (4ux, 0); 
          label(\btex $4$ etex, (4ux, 0));

          % 2nd row
          draw box shifted (0ux, -1uy); 
          label(\btex $5$ etex, (0ux, -1uy));

          draw box shifted (2ux, -1uy); 
          label(\btex $5$ etex, (2ux, -1uy));

          draw box shifted (4ux, -1uy); 
          label(\btex $5$ etex, (4ux, -1uy));

          % 3rd row
          draw box shifted (0ux, -2uy); 
          label(\btex $6$ etex, (0ux, -2uy));

          draw image (draw box shifted (2ux, -2uy)) anglestriped (1, 45, 2) ;
          fill fullsquare xysized (0.5ux, 0.5uy) shifted (2ux, -2uy)
               withcolor white;
          label(\btex $7$ etex, (2ux, -2uy));

          draw box shifted (4ux, -2uy); 
          label(\btex $6$ etex, (4ux, -2uy));
        \end{mpost}}
    \end{wrapfigure}
   \textsc{CheeseMaze:} The CheeseMaze environment is a POMDP with masked
    states proposed in \cite{McCallum_1993}. The environment consists of
    11~states and 7~observations as shown on the right. The objective is to
    reach the goal state, which is indicated by observation~$7$. The agent only
    receives a reward of $+1$, when the goal state is reached. 
  }
\end{enumerate}

For all three environments, we compare the performance of AIS-based PORL with
the LSTM+PPO baseline, described earlier. We also compare
with the best performing planning solution from the JuliaPOMDP repository
\citep{JuliaPOMDPs}. The results are
presented in Fig.~\ref{fig:low}, which shows both AIS-based PORL and LSTM+PPO
converge close to the planning solutions relatively quickly.\footnote{The
  performance of all learning algorithms for \textsc{CheeseMaze} are better
  than the best planning solution. We solved the \textsc{CheeseMaze} model
  with other solvers available in the JuliaPOMDP \cite{JuliaPOMDPs}, and all
  these solution performed worse than the solution obtained by incremental
pruning presented here.}

\subsection{Moderate-dimensional environments}

\begin{figure}[!ht]
    \centering
    \hfill
    \begin{subfigure}[t]{0.475\textwidth}
      \includegraphics[width=\textwidth]{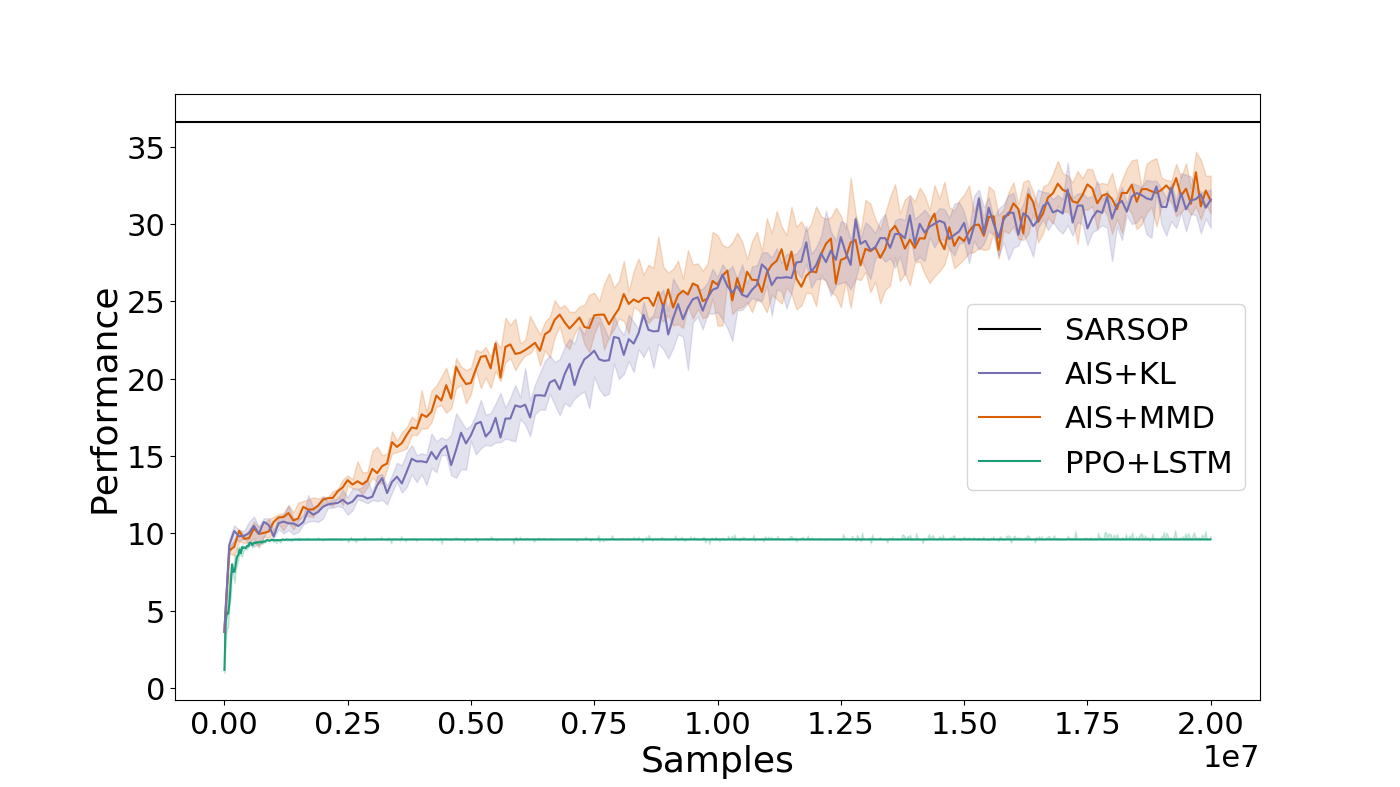}
        \caption{Rock Sampling} \label{fig:rocksampling_withoutprior_results}
    \end{subfigure}
    \hfill
    \begin{subfigure}[t]{0.475\textwidth}
      \includegraphics[width=\textwidth]{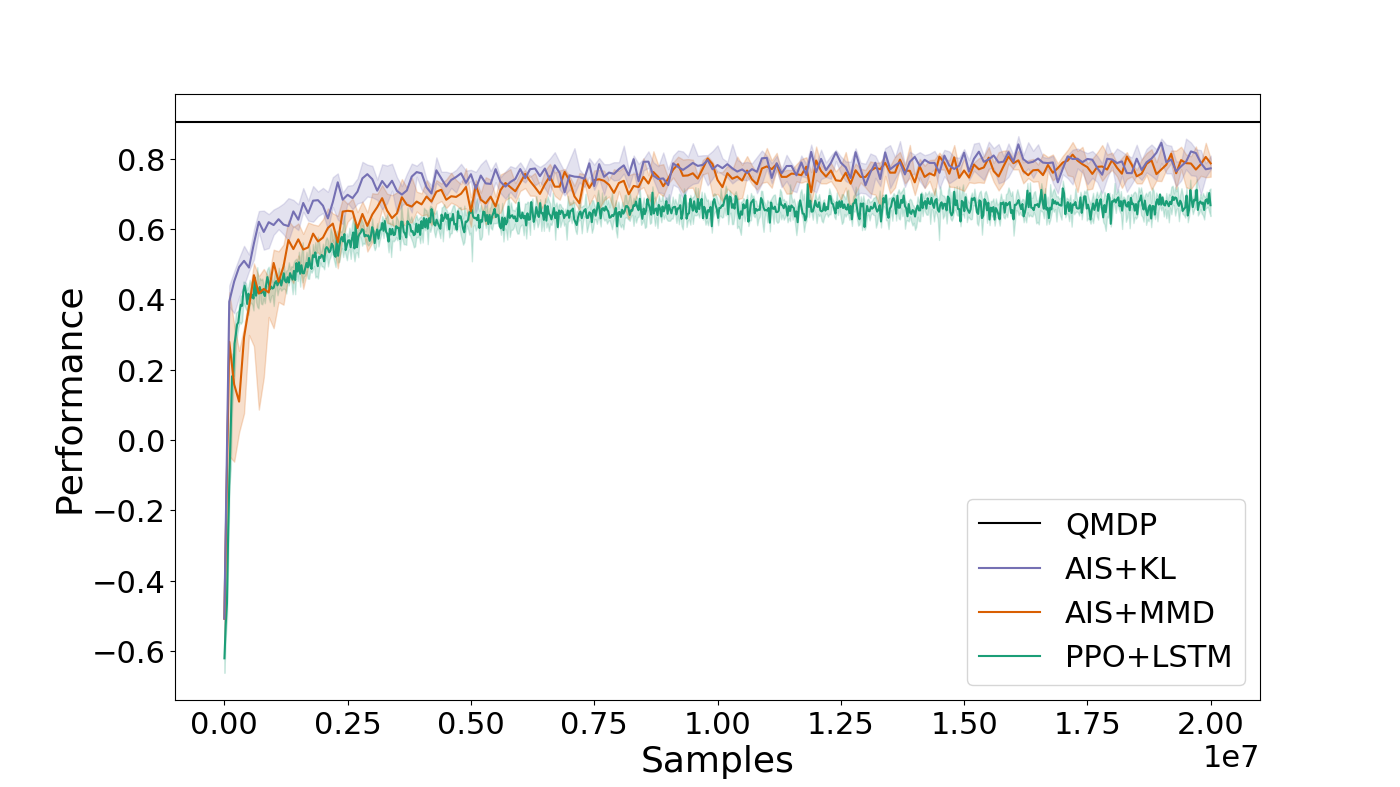}
        \caption{Drone Surveillance} \label{fig:dronesurveillance_results}
    \end{subfigure}
    \hfill
    \caption{Comparison of AIS-based actor only PORL algorithm with LSTM+PPO
    baseline for moderate-dimensional environments (for 10 random seeds).} \label{fig:moderate}
\end{figure}

In these environments, the size of the unobserved state is moderately large
(of the order of $10^2$ to $10^3$ unobserved states) and the optimal planning
solution cannot be easily obtained using standard POMDP solvers. However, an
approximate planning solution can be easily obtained using standard
approximation algorithms for POMDPs. 

\begin{enumerate}
    % https://tex.stackexchange.com/questions/232105/wrapfigure-in-an-enumerate-environment
 \item
    \parbox[t]{\dimexpr\textwidth-\leftmargin}{%
      \vspace{-2.5mm}
      \begin{wrapfigure}[9]{r}{0.25\textwidth}
        \centering
        \vspace{-1\baselineskip}
        \includegraphics[width=\linewidth]{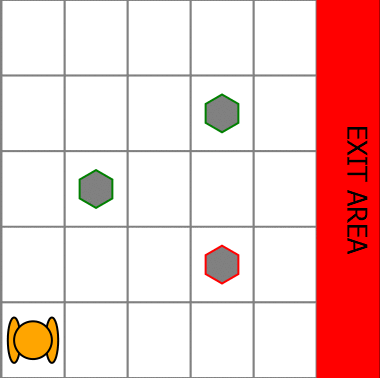}
      \end{wrapfigure}
   \textsc{RockSample:} RockSample is a scalable POMDP environment
    introduced in \cite{Smith2004} which models the rover science exploration.
    The RockSample$(n,k)$ environment consists of a $n \times n$ grid with $k$
    rocks. The rocks are at known positions. Some of the rocks which are labeled as
    \textsc{good} rocks have scientific values; other rocks which are labeled as
    \textsc{bad} rocks do not. Sampling a rock is expensive and the agent has a
    noisy long-range sensor to help determine if a rock is \textsc{good}
    before choosing to approach and sample it. At each stage, the agent can
    choose from $k+5$ actions: \textsc{north}, \textsc{south}, \textsc{east},
    \textsc{west}, \textsc{sample}, \textsc{check}\textsubscript{1}, \ldots,
    \textsc{check}\textsubscript{$k$}. The first four are deterministic
    single-step motion actions. The \textsc{sample} action samples the rock at
    the current location; if the rock is \textsc{good}, there is a reward of
    $+20$ and the rock becomes \textsc{bad} (so that no further reward can be
    gained from sampling it); if the rock is \textsc{bad}, there is a negative
    reward of $-10$. The right edge of the map is a terminal state and
    reaching it gives a reward of $+10$. In our experiments, we use a
    RockSample$(5,3)$ environment.
  }

  \item 
    \parbox[t]{\dimexpr\textwidth-\leftmargin}{%
      \vspace{-2.5mm}
      \begin{wrapfigure}[9]{r}{0.25\textwidth}
        \centering
        \vspace{-1\baselineskip}
        \includegraphics[width=\linewidth]{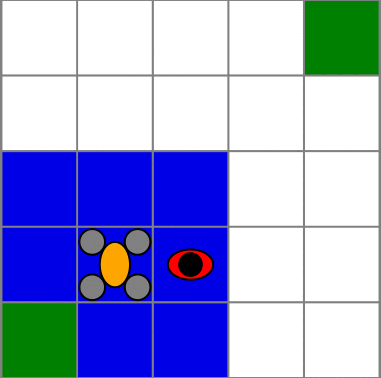}
      \end{wrapfigure}
    \textsc{DroneSurveillance:} DroneSurveillance is a POMDP model of
    deploying an autonomous aerial vehicle in a partially observed, dynamic,
    indoor environment introduced in \cite{Svorenova2015}. The environment is a
    $5 \times 5$ grid with two agents: a ground agent which moves randomly and
    an aerial agent, whose motion has to be controlled. The aerial agent
    starts at the bottom-left cell and has to reach the upper-right cell (the
    goal state) without being in the same location as the ground agent. The
    ground agent cannot enter the start or goal states. The aerial agent has a
    downward facing camera which can view a $3\times3$ grid centered at its
    current location and it can perfectly see the location of the ground agent
    if it is in this view. At each stage, the aerial agent may choose from $5$
    actions: \textsc{north}, \textsc{south}, \textsc{east}, \textsc{west},
    \textsc{hover}. The first four are deterministic single-step motion
    actions and the \textsc{hover} action keeps the aerial vehicle at its
    current position. Reaching the goal gives a reward of $+1$ and ends the
    episode. If both agents are in the same cell, there is a negative reward
    of $-1$ and the episode ends. 
    }
\end{enumerate}

The visualizations above are taken from the JuliaPOMDP environments
\citep{JuliaPOMDPs}. For both environments, we compare the performance of
AIS-based PORL with the LSTM+PPO baseline described earlier. We also compare
with the best performing planning solution from the JuliaPOMDP repository
\citep{JuliaPOMDPs}. The results are shown in Fig.~\ref{fig:moderate} which
shows that both AIS-based PORL algorithms converge close to the best planning
solution in both environments. The performance of LSTM+PPO is similar in
\textsc{DroneSurveillance} but LSTM+PPO gets stuck in a local minima in
\textsc{RockSample}.

\subsection{High-dimensional environments} \label{subsec:Minigrid_Experiments}

We use the MiniGrid environments from the BabyAI platform
\citep{gym_minigrid}, which are partially observable 2D grid environments
which has tasks of increasing complexity level. The environment has multiple
entities (agent, walls, lava, boxes, doors, and keys); objects can be picked up,
dropped, and moved around by the agent; doors can be unlocked via keys of the
same color (which might be hidden inside boxes). The agents can see a
$7\times7$ view in front of it but it cannot see past walls and closed
doors. At each time, it can choose from the
following actions: $\{$\textsc{Move Forward}, \textsc{Turn Left}, \textsc{Turn
Right}, \textsc{Open Door/Box}, \textsc{Pick up item}, \textsc{Drop Item}, \textsc{Done}$\}$.
The agent can only hold one item at a time. 
The objective is to reach a goal state in the quickest amount of time (which
is captured by assigning to the goal state a reward which decays over time).

Most of the environments have a certain theme, and we cluster the environments
accordingly. The visualizations below are taken from the Gym Minigrid environments
\citep{gym_minigrid}.

\begin{enumerate}
  \item 
    \parbox[t]{\dimexpr\textwidth-\leftmargin}{%
      \vspace{-2.5mm}
      \begin{wrapfigure}[7]{r}{0.2\textwidth}
        \centering
        \vspace{-1\baselineskip}
        \includegraphics[width=\linewidth]{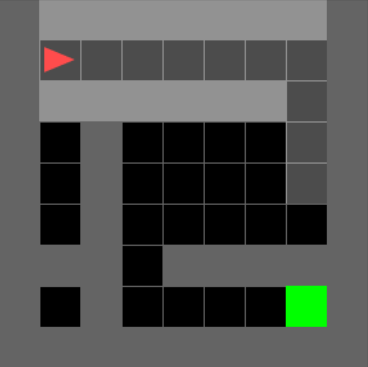}
      \end{wrapfigure}
   \textsc{Simple Crossing:} A simple crossing environment is a 2D grid with
   columns of walls with an opening (or a crossing). The agent can traverse
   the wall only through the openings and needs to find a path from the start
   to the goal state. There are four such environments
    (MGSCS9N1, MGSCS9N2, MGSCS9N3, and MGSCS11N5) where the label S$n$N$m$
    means that the size of the environment  is $n \times n$ and there are $m$
    columns of walls.
  }

  \item 
    \parbox[t]{\dimexpr\textwidth-\leftmargin}{%
      \vspace{-2.5mm}
      \begin{wrapfigure}[7]{r}{0.2\textwidth}
        \centering
        \vspace{-1\baselineskip}
        \includegraphics[width=\linewidth]{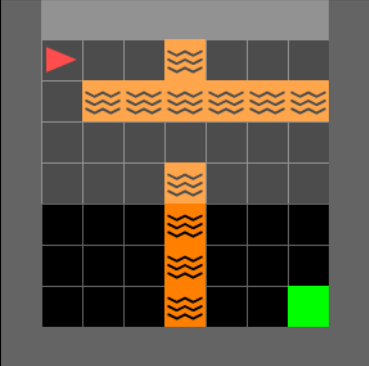}
      \end{wrapfigure}
   \textsc{Lava Crossing:} The lava crossing environments are similar to the
   simple crossing environments, but the walls are replaced by lava. If the
   agent steps on to the lava block then it dies and the episode ends.
   Therefore, exploration is more difficult in lava crossing as compared to
   simple crossing. There are two such environments (MGLCS9N1 and MGLCS9N2)
   where the label S$n$N$m$ has the same interpretation as simple crossing.
  }

  \item 
    \parbox[t]{\dimexpr\textwidth-\leftmargin}{%
      \vspace{-2.5mm}
      \begin{wrapfigure}[7]{r}{0.2\textwidth}
        \centering
        \vspace{-1\baselineskip}
        \includegraphics[width=\linewidth]{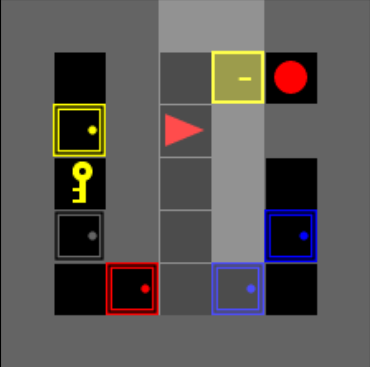}
      \end{wrapfigure}
   \textsc{Key Corridor:} The key corridor environments consist of a central
   corridor which has rooms on the left and right sides which can be accessed
   through doors. When the door is locked it can be opened using a key of the
   same color. The agent has to move to the location of the key, pick it up,
   move to the location of the correct door, open the door, drop the key, and
   pick up the colored ball. There are three such environments (MGKCS3R1,
   MGKCS3R2, and MGKCS3R3), where the label S$n$R$m$ means that the size of
   the grid is proportional to $n$ and the number of rooms present is
   $2 m$.
  }

  \item 
    \parbox[t]{\dimexpr\textwidth-\leftmargin}{%
      \vspace{-2.5mm}
      \begin{wrapfigure}[6]{r}{0.3\textwidth}
        \centering
        \vspace{-1\baselineskip}
        \includegraphics[width=\linewidth]{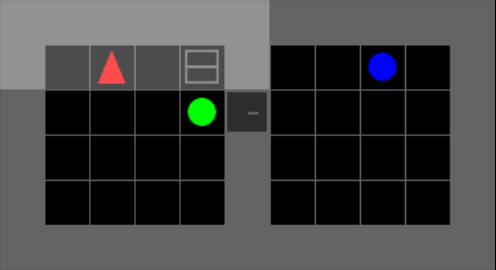}
      \end{wrapfigure}
   \textsc{Obstructed Maze:} The obstructed maze environments are similar to 
   key corridor environments but the key is inside a box and the box has to be
   opened to find the key. We consider two such environments (MGOM1Dl and
   MGOM1Dlh). In MGOM1Dl box is already open while in MGOM1Dlh the box is
   closed. There is an additional such environment in the BabyAI platform
   (MGOM1Dlhb), which is more suitable for continual learning algorithms so
   we exclude it here.
  }

\end{enumerate}

The number of observations in a given Minigrid environment is discrete but is
too large to model it as a one-hot encoded discrete observation as done in the previous environments. Instead we compress the observations as described in Section~\ref{sec:obsais} by using an autoencoder to convert a large discrete space to a continuous space with a tractable size. A separate autoencoder is trained for each environment using a dataset that is created by performing random rollouts. Once the autoencoder is trained over the fixed dataset for several epochs, it is fixed and used to generate the observations for learning the AIS. This is very similar to \cite{ha2018world}, where they learn the autoencoder in a similar fashion and then fix it, following which their training procedure for the next observation distribution prediction and policy takes place.

\begin{figure}[!p]
    \centering
    \begin{subfigure}[t]{0.475\textwidth}
        \includegraphics[width=\textwidth]{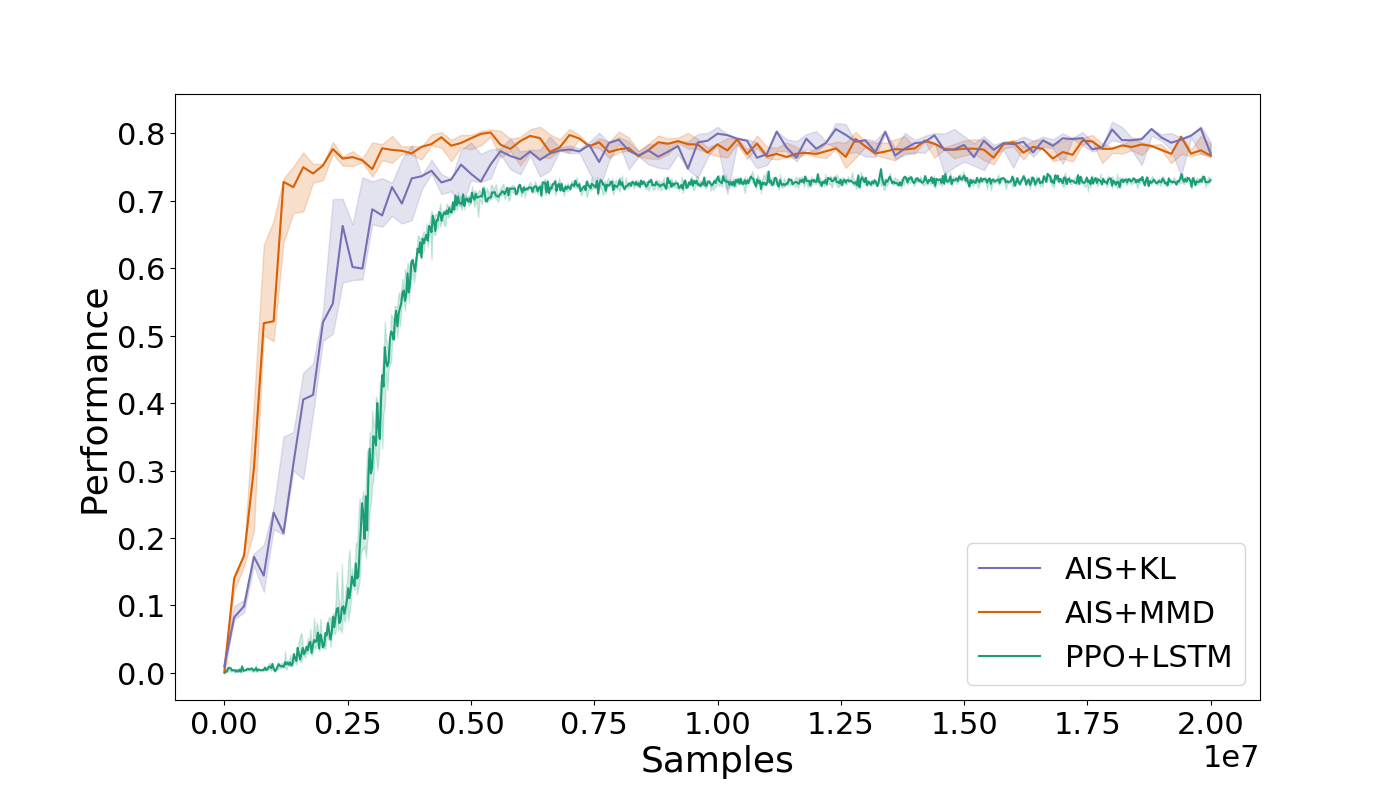}
        \caption{MGSCS9N1}
    \end{subfigure}
    \begin{subfigure}[t]{0.475\textwidth}
        \includegraphics[width=\textwidth]{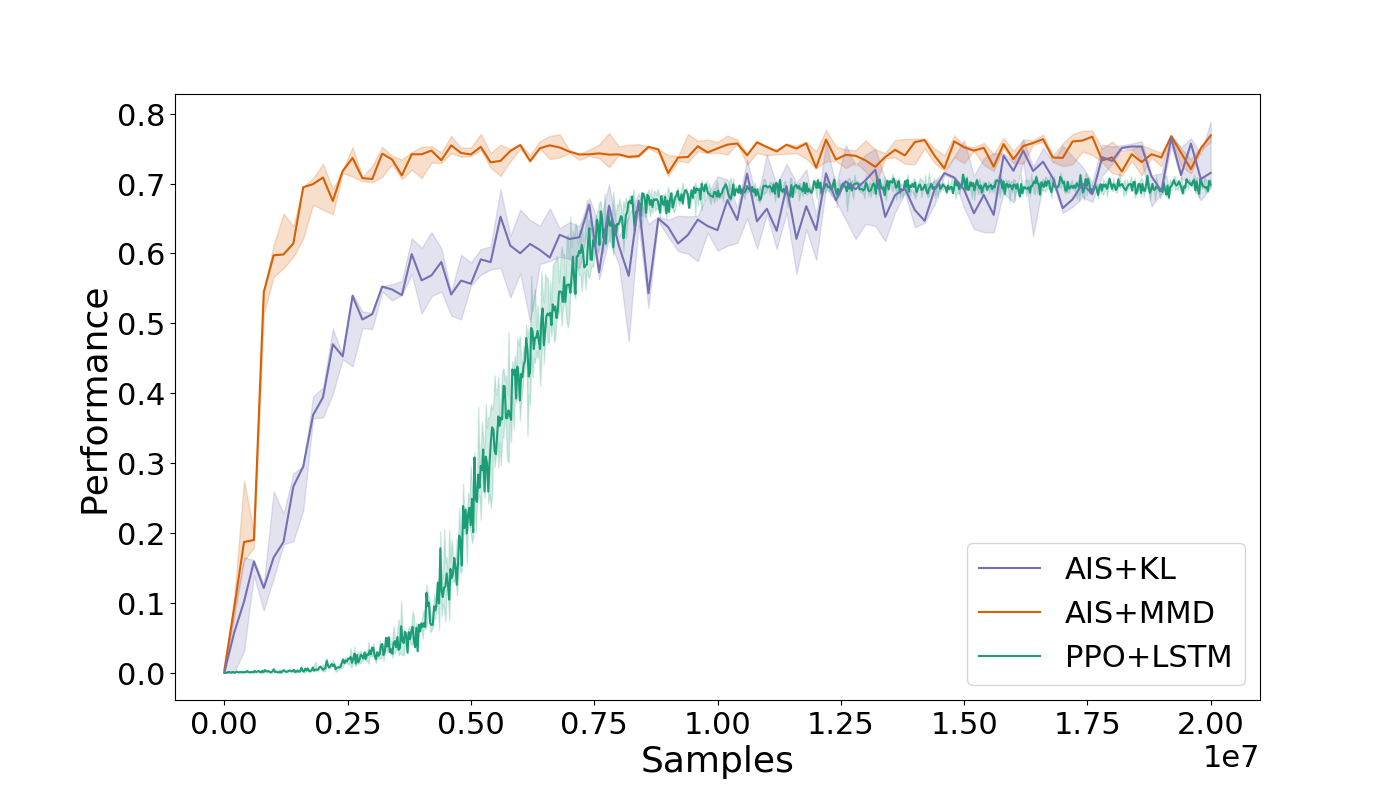}
        \caption{MGSCS9N2}
    \end{subfigure}

    \begin{subfigure}[t]{0.475\textwidth}
        \includegraphics[width=\textwidth]{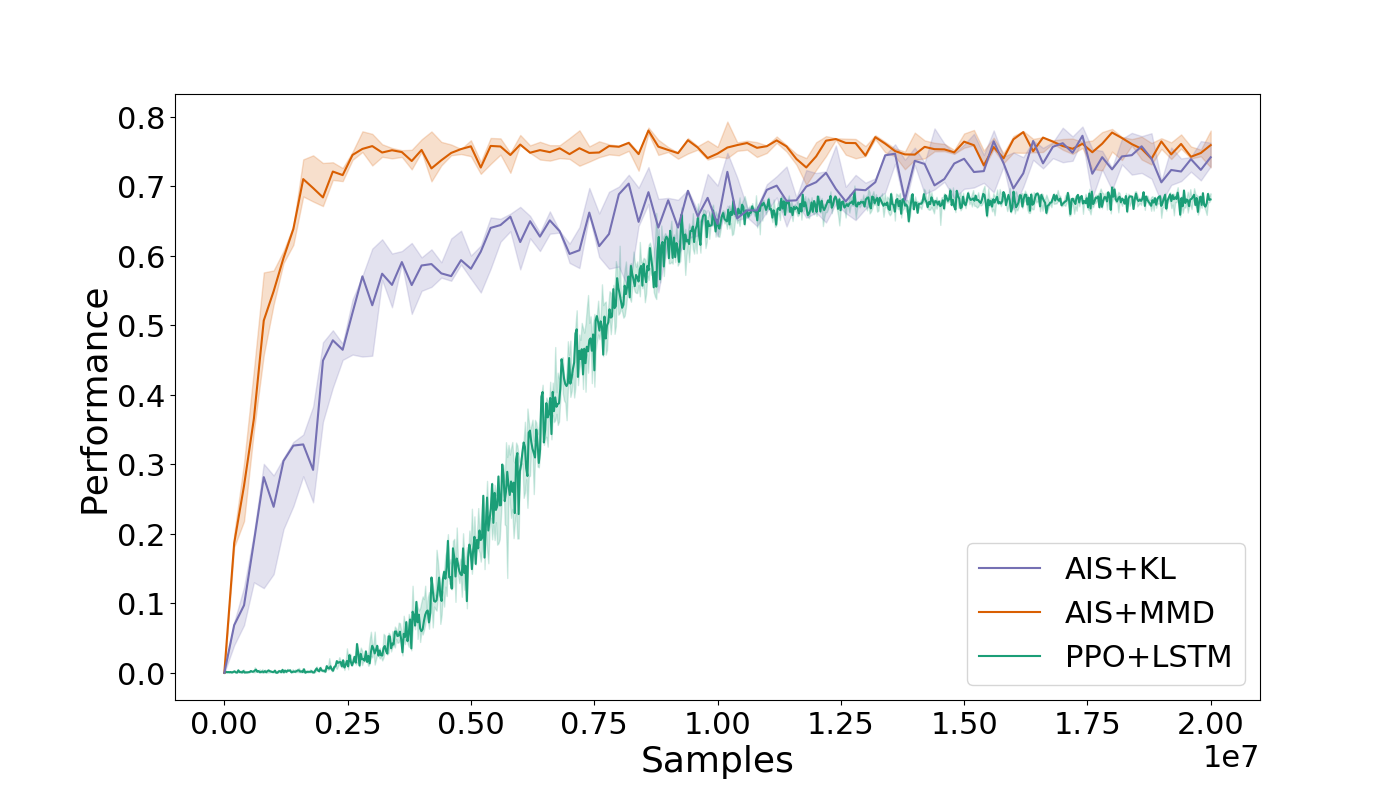}
        \caption{MGSCS9N3}
    \end{subfigure}
    \begin{subfigure}[t]{0.475\textwidth}
        \includegraphics[width=\textwidth]{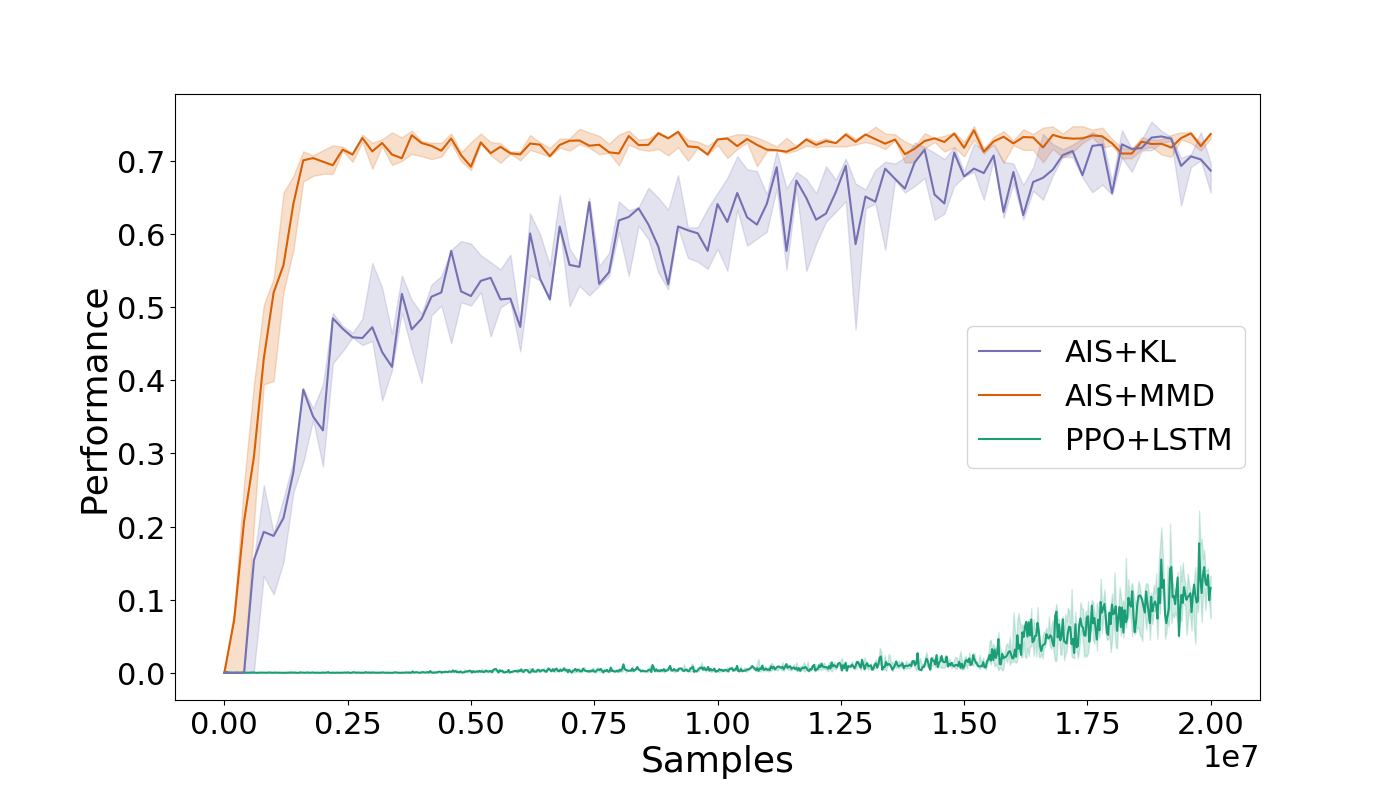}
        \caption{MGSCS11N5}
    \end{subfigure}

    \begin{subfigure}[t]{0.475\textwidth}
        \includegraphics[width=\textwidth]{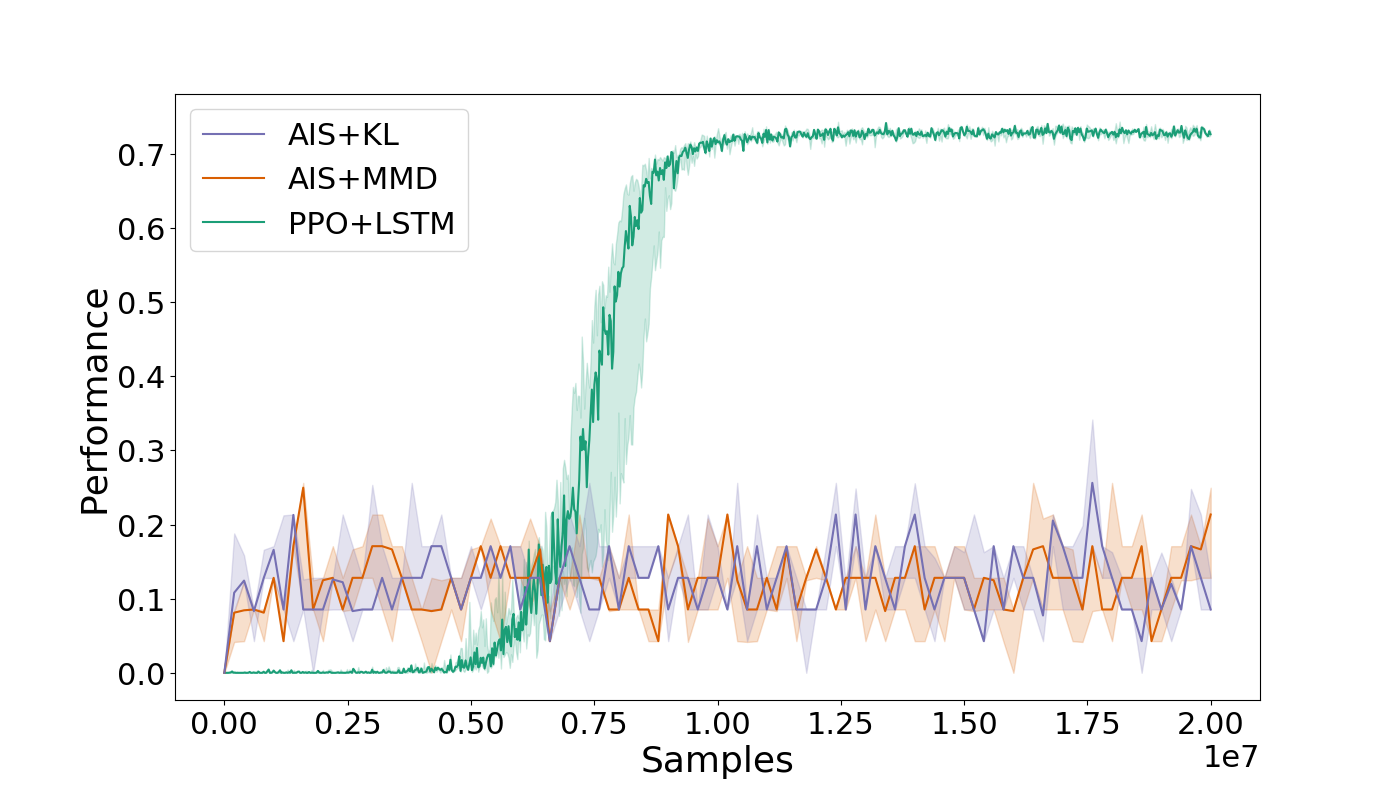}
        \caption{MGLCS9N1} \label{fig:MGLCS9N1_results}
    \end{subfigure}
    \begin{subfigure}[t]{0.475\textwidth}
        \includegraphics[width=\textwidth]{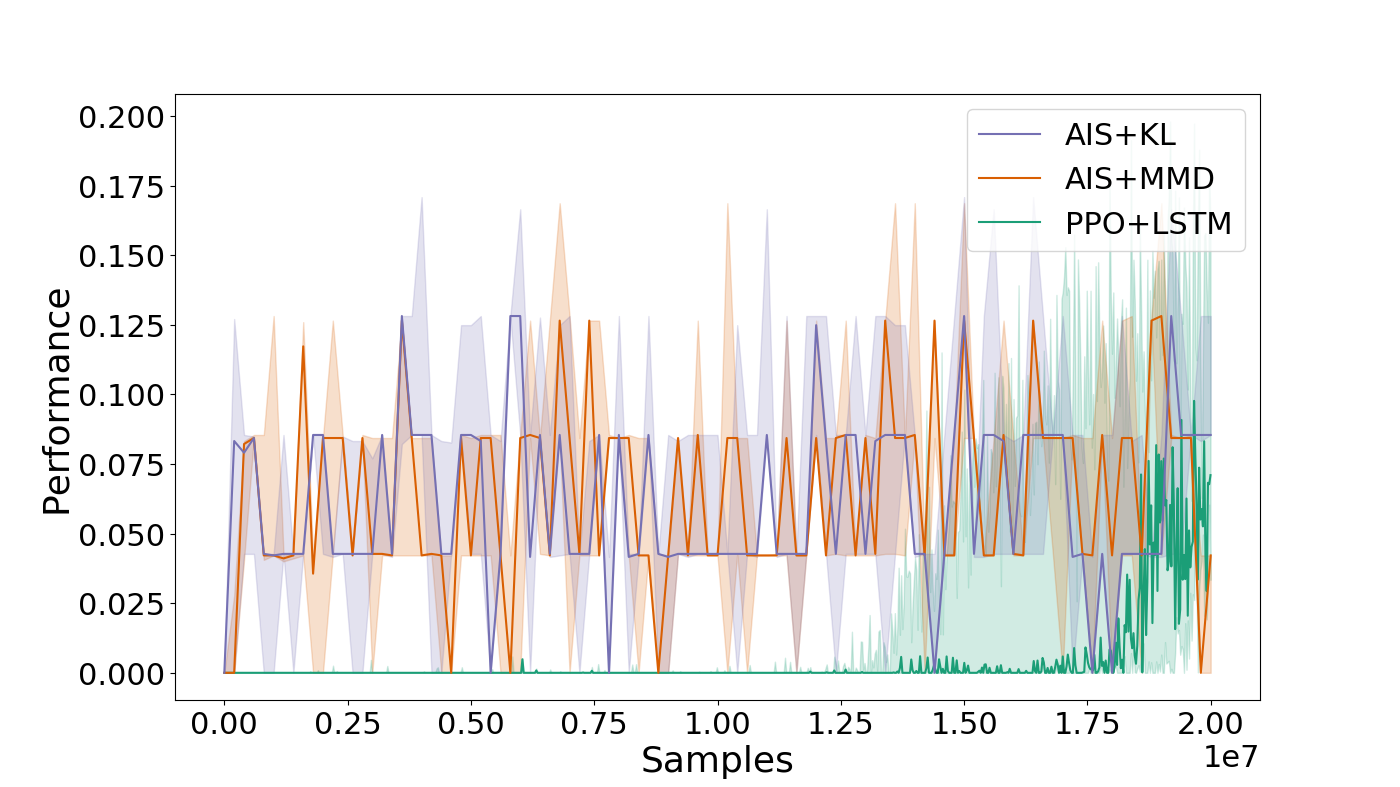}
        \caption{MGLCS9N2} \label{fig:MGLCS9N2_results}
    \end{subfigure}
    
    \begin{subfigure}[t]{0.475\textwidth}
        \includegraphics[width=\textwidth]{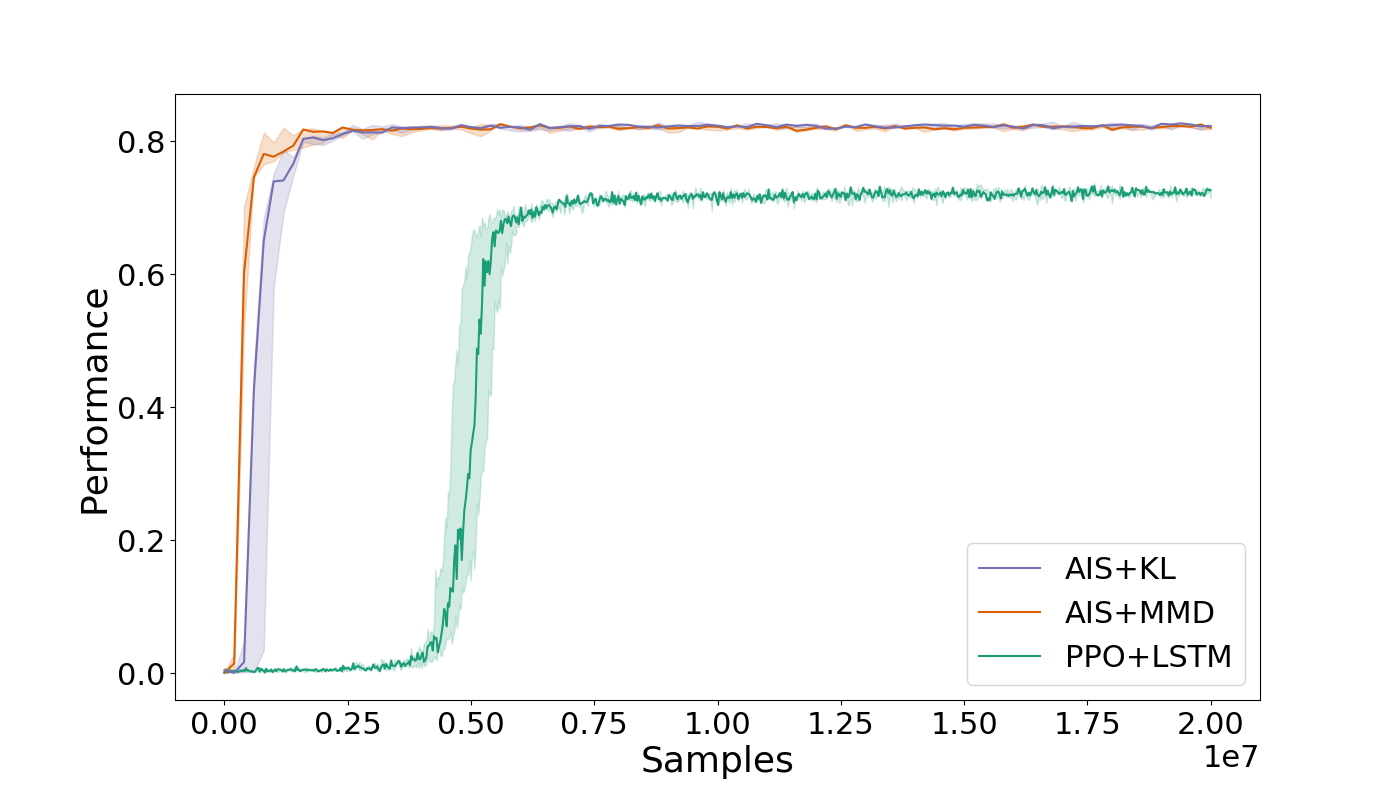}
        \caption{MGKCS3R1}
    \end{subfigure}
    \begin{subfigure}[t]{0.475\textwidth}
        \includegraphics[width=\textwidth]{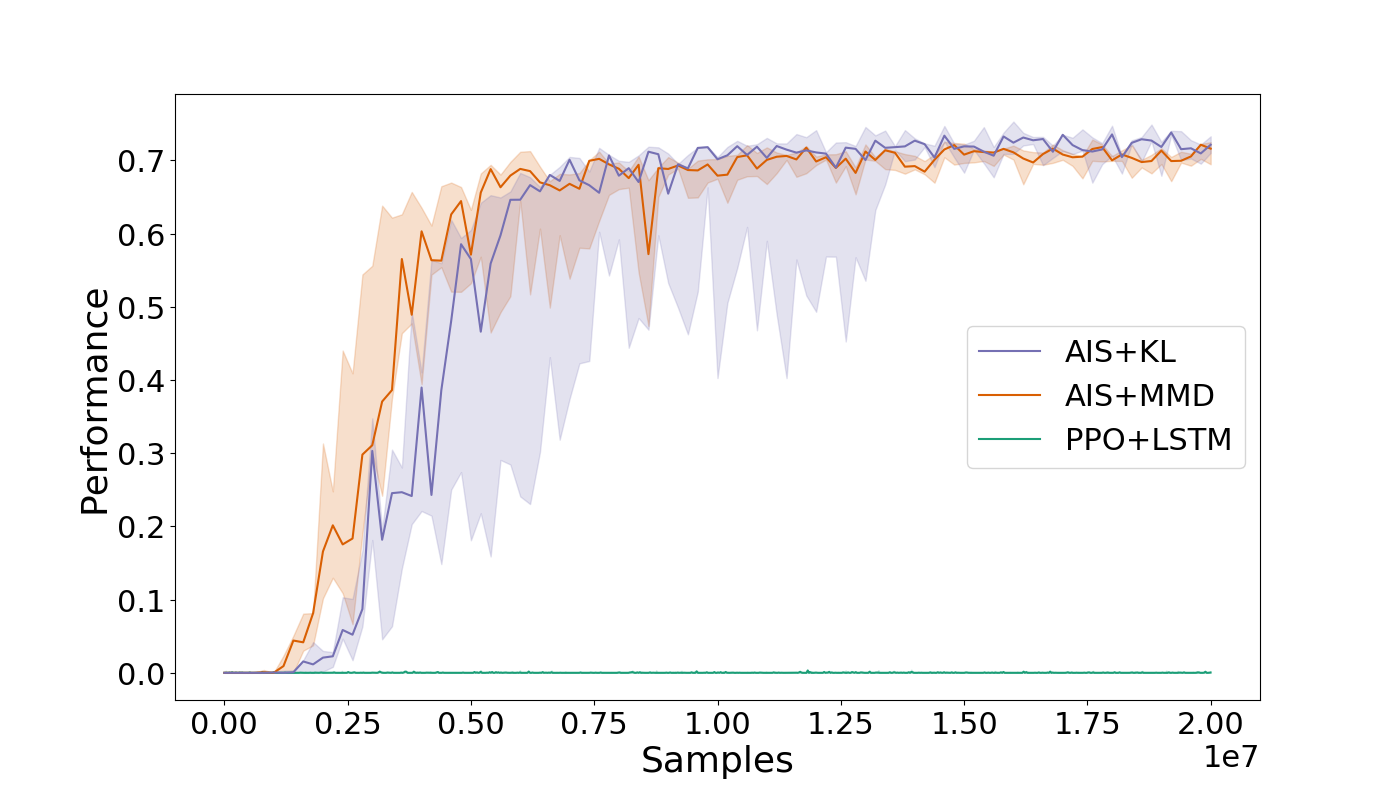}
        \caption{MGKCS3R2}
    \end{subfigure}
    \caption{Comparison of AIS-based actor only PORL algorithm with LSTM+PPO
    baseline for high-dimensional environments (for 5 random seeds).} \label{fig:high}
\end{figure}

\begin{figure}[!t]\ContinuedFloat
    \centering
    \begin{subfigure}[t]{0.475\textwidth}
        \includegraphics[width=\textwidth]{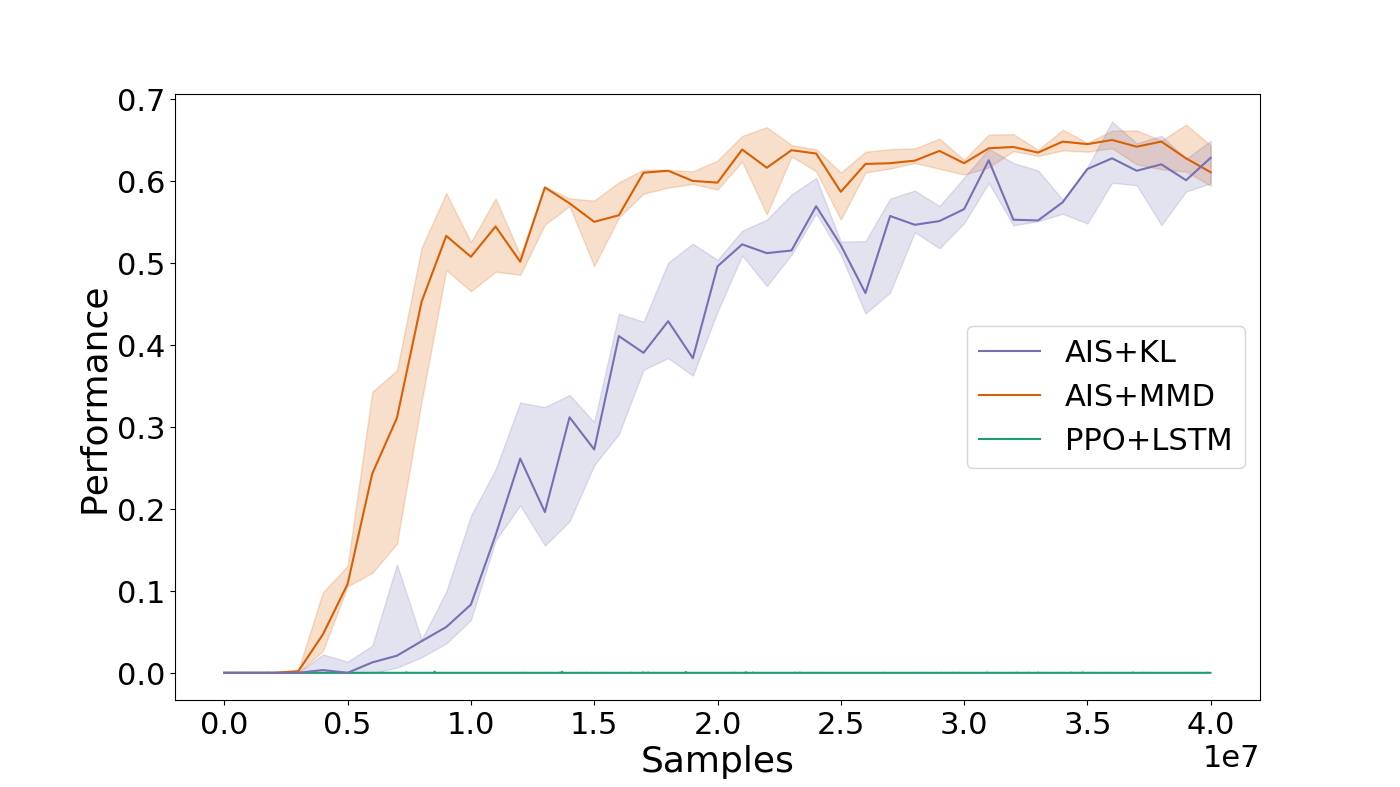}
        \caption{MGKCS3R3} \label{fig:MGKCS3R3_results}
    \end{subfigure}
    \begin{subfigure}[t]{0.475\textwidth}
        \includegraphics[width=\textwidth]{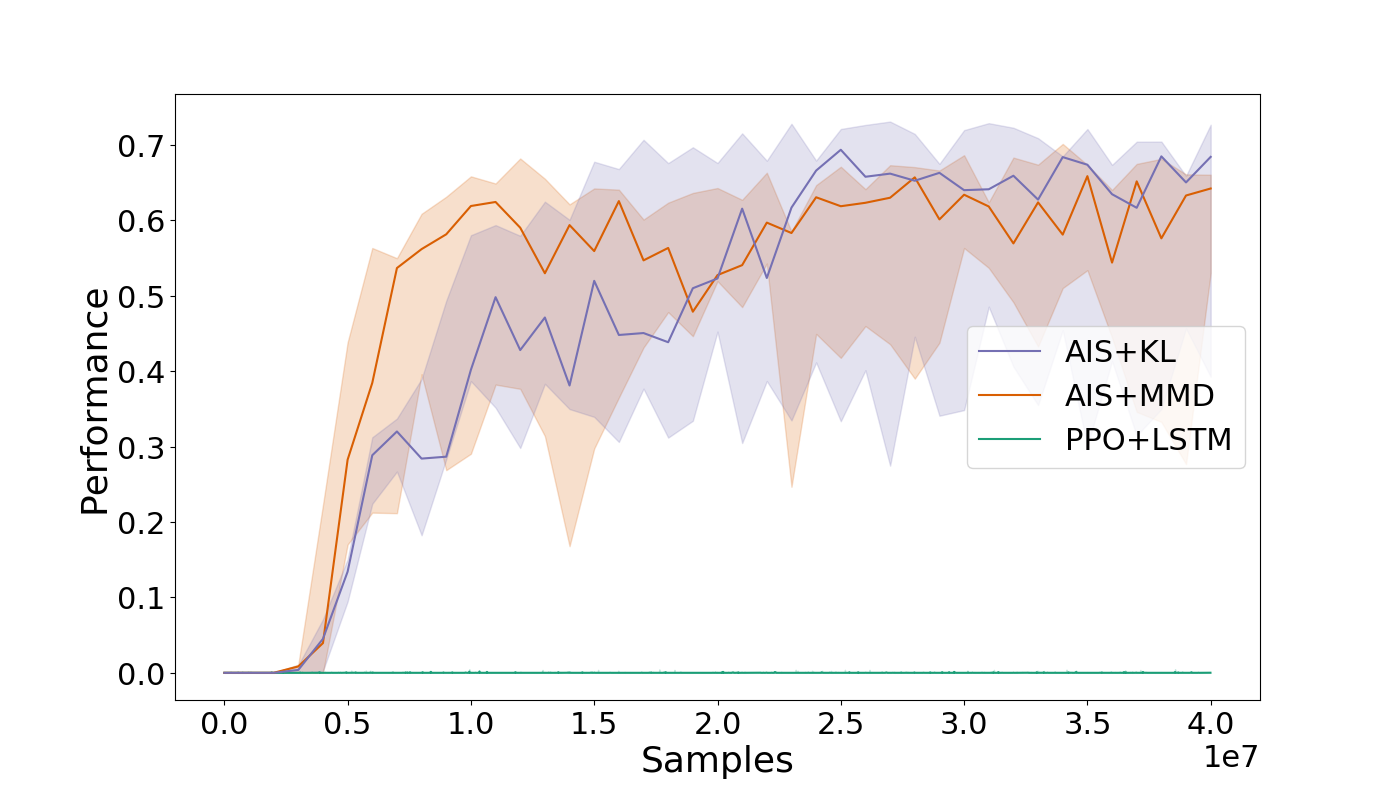}
        \caption{MGOM1Dl} \label{fig:MGOM1Dl_results}
    \end{subfigure}

    \begin{subfigure}[t]{0.475\textwidth}
        \includegraphics[width=\textwidth]{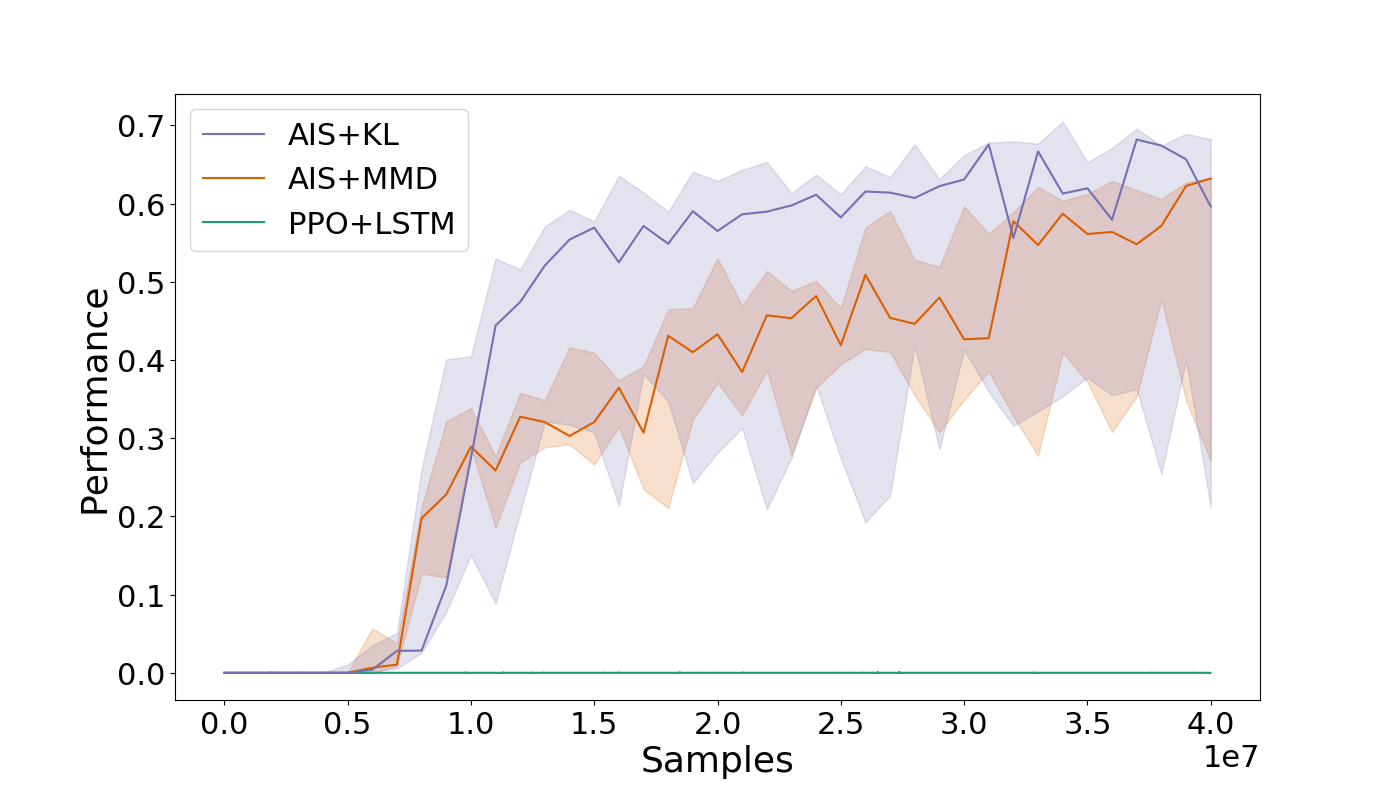}
        \caption{MGOM1Dlh} \label{fig:MGOM1Dlh_results}
    \end{subfigure}
    \caption*{Figure~\ref{fig:high} (continued): Comparison of AIS-based actor only PORL algorithm with LSTM+PPO
    baseline for high-dimensional environments (for 5 random seeds).}
\end{figure}

Note that the output of the autoencoder is a continuous variable and we are
using MMD with $p=2$ as an IPM. As explained in Section~\ref{sec:learn-IPM},
$d_{\F_2}(\mu,\nu)^2$ only depends on the mean of $\mu$ and $\nu$. So, to
simplify the computations, we assume that $\nu$ is a Dirac delta distribution
centered at its mean. Thus, effectively, we are predicting the mean of the
next observation. In general, simply predicting the mean of the observations
may not lead to a good representation, but in the Minigrid environments, the
transitions are deterministic and the only source of 
stochasticity in the observations is due to the initial configuration of the
environment. So, in practice, simply predicting the mean of the next
observation works reasonably well. We emphasize that for other more general
environments with truly stochastic observations, such a choice of IPM may not
work well and it may be better to choose the MMD $d_{\F_p}$ defined in
Proposition~\ref{prop:RKHS} for a different value of~$p$, say $p=1$ (which
corresponds to the energy distance \citep{Szekely2004}).

For all minigrid environments, we compare the performance of
AIS-based PORL with the LSTM+PPO baseline proposed in
\cite{chevalier2018babyai}. The results are shown in Fig.~\ref{fig:high} which
shows that for most environments AIS-based PORL converges to better
performance values. Note that AIS-based PORL fails to learn in the
\textsc{Lava Crossing} environments (MGLCS9N1 and MGLCS9N2) while LSTM+PPO
fails to learn in the larger \textsc{Key Crossing} environments (MGKCS3R2 and
MGKCS3R3) and in the \textsc{Obstructed Maze} environments (MGOM1Dl and
MGOM1Dlh).

\blue{The results indicate that one IPM does not necessarily lead to better
  performance than others in all cases. The performance of a particular IPM
  depends on whether the observation and AIS spaces are discrete or
  continuous, on the size of these spaces, and the stochasticity of the
  environment. The fact that we are approximating the policy using non-linear
  neural networks makes it difficult to quantify the impact of the choice of
  IPM on the accuracy of learning. It will be important to understand this
  impact in more detail and develop guidelines on how to choose an IPM based
on the features of the environment.}

\section{Conclusion}
In this paper, we present a theoretical framework for
approximate planning and learning in partially observed system. Our
framework is based on the fundamental notion of information state. We
provide two equivalent definitions of information state. An information
state is a function of history which is sufficient to compute the expected
reward and predict its next value. Equivalently, an information state is a
function of the history which can be recursively updated and is sufficient
to compute the expected reward and predict the next observation. We show
that an information state always leads to a dynamic programming
decomposition and provide several examples of simplified dynamic programming
decompositions proposed in the literature which may be viewed as specific
instances of information states.

We then relax the definition of an information state to describe an
\acf{AIS}, which is a function of the history that approximately satisfies
the properties of the information state. We show that an \acs{AIS} can be
used to identify an approximately optimal policy with the approximation
error specified in terms of the ``one-step'' approximation errors in the
definition of the \acs{AIS}. We present generalizations of \acs{AIS} to
setups with observation and action compression as well as to multi-agent
systems. We show that various approximation approaches 
for both fully and partially observed setups proposed in the literature may
be viewed as special cases of \acs{AIS}.

One of the salient features of the \ac{AIS} is that it is defined in terms of
properties that can be estimated from data, and hence the corresponding
\ac{AIS} generators can be learnt from data. These can then be used as
history representations in partially observed reinforcement learning (PORL)
algorithms. We build up on this idea to present policy gradient algorithms
which learn an \acs{AIS} representation and an optimal policy and/or
action-value function using multi time-scale stochastic gradient descent.

We present detailed numerical experiments which compare the performance of
\acs{AIS}-based PORL algorithms with a state-of-the-art PORL algorithm for 
three classes of partially observed problems---small, medium and large scale problems---and find out that \acs{AIS}-based PORL outperforms the state-of-the-art baseline in most cases.

We conclude by observing that in this paper we restricted attention to the
simplest classes of algorithms but the same idea can be extended to develop
\acs{AIS}-based PORL algorithms which uses value-based approaches such as
Q-learning and its improved variants such as DQN, DDQN, distributional RL,
etc. Finally, we note that the \acs{AIS} representation includes a model of
the system, so it can be used as a component of model-based reinforcement
learning algorithms such as Dyna~\cite[Sec 8.2, page 161]{SuttonBarto_2018}.
Such an approach will provide anytime guarantees on the approximation error
which will depend on the ``one-step'' approximation error of the current
\acs{AIS}-representation. Therefore, we believe that \acs{AIS} presents a
systematic framework to reason about learning in partially observed
environments.

\section*{Acknowledgment}
The authors are grateful to Demosthenis Teneketzis, Peter Caines, and Dileep
Kalathil for useful discussions and feedback. The work of JS and AM was
supported in part by the Natural Science and Engineering Research Council of
Canada through Discovery Grant RGPIN-2016-05165. The work of AS, RS, and AM
was supported in part by the Innovation for Defence Excellence and Security
(IDEaS) Program of the Canadian Department of National Defence through grant
CFPMN2-037. AS was also supported by an FRQNT scholarship. The numerical
experiments were enabled in part by support provided by Calcul Québec and
Compute Canada.

\bibliography{IEEEabrv,multiagent.bib}

\newpage

\appendices
\addappheadtotoc
\section{Comparison with the results of \texorpdfstring{\cite{Abel2016}}{Abel
et al. (2016)} for state aggregation in MDPs} \label{app:Abel}

\cite{Abel2016} introduce four models of state aggregation and derive
approximation bounds for all four. In this section, we show that one of these
models, which they call \emph{approximate model similarity} may be viewed as
an \acs{AIS}. We also show that the approximation bounds of
Theorem~\ref{thm:inf-ais} for this model are stronger than those derived in
\cite{Abel2016} by a factor of $\mathcal O(1/(1-\discount))$.

Since we follow a slightly different notation than \cite{Abel2016} and for the
sake of completeness, we start by describing the notion of approximate model
similarity defined in \cite{Abel2016}.

Consider an infinite horizon finite-state finite-action MDP with state space
$\StSp$, action space $\ActSp$, transition probability matrix $P \colon \StSp
\times \ActSp \to \Delta(\StSp)$, per-step reward function $r \colon \StSp
\times \ActSp \to \reals$, and discount factor $\discount$.   

Let $\hat \StSp$ be an aggregated state space and it is assumed that the
following two functions are available: a compression function $q \colon \StSp
\to \hat \StSp$ and a weight function $w \colon \StSp \to [0, 1]$ such that 
for all $\hat \st \in \hat \StSp$, $\sum_{\st \in q^{-1}(\hat \st)} w(\st) =
1$. Given these functions, define an aggregated MDP with state space $\hat
\StSp$, action space $\ActSp$, transition probability function $\hat P \colon
\hat \StSp \times \ActSp \to \hat \StSp$ given by
\[
   \hat P(\hat \st' | \hat \st, \act) = 
   \sum_{\st \in q^{-1}(\hat \st)} \sum_{\st' \in q^{-1}(\hat \st')}
   P(\st' | \st, \act) w(\st),
   \quad
   \forall \hat \st, \hat \st' \in \hat \StSp, \act \in \ActSp,
\]
and a per-step reward $\hat r \colon \hat \StSp \times \ActSp \to \reals$
given by 
\[
  \hat r(\hat \st, \act) = \sum_{\st \in q^{-1}(\hat \st)} r(\st, \act) w(\st), 
  \quad \forall \hat \st \in \hat \StSp, \act \in \ActSp.
\]

\begin{definition}[$\varepsilon$-approximate model similarity \citep{Abel2016}]
  The aggregated MDP is said to be $\varepsilon$-approximate model similar to
  the original MDP if it satisfies the following two properties:
  \begin{enumerate}
    \item For all $\hat \st \in \hat \StSp$, $s_1, s_2 \in q^{-1}(\hat \st)$, and
      $\act \in \ActSp$, we have
      \[
        \bigl| r(s_1, \act) - r(s_2, \act) \bigr| \le \varepsilon.
      \]
    \item For all $\hat \st, \hat \st' \in \hat \StSp$, $s_1, s_2 \in q^{-1}(\hat
      \st)$, and $\act \in \ActSp$, we have
      \[
        \biggl| 
        \sum_{\st' \in q^{-1}(\hat \st')} P(\st' | s_1, \act)
        -
        \sum_{\st' \in q^{-1}(\hat \st')} P(\st' | s_2, \act) 
        \biggr| \le \varepsilon.
      \]
  \end{enumerate}
\end{definition}

\begin{proposition}[Lemma 2 of \cite{Abel2016}]\label{prop:Abel}
  Let $\hat \pol \colon \hat \StSp \to \ActSp$ be the (deterministic) optimal
  policy for the aggregated MDP. Define $\pol \colon \StSp \to \ActSp$ by
  $\pol = \hat \pol \circ q$. Let $V \colon \StSp \to \reals$ denote the
  optimal value function and let $V^\pol \colon \StSp \to \reals$ denote the
  value function for policy~$\pol$. Then, for all $\st \in \StSp$
  \[
     \bigl| V(\st) - V^{\pol}(\st) \bigr| \le 
     \frac{2 \varepsilon}{(1 - \discount)^2} + 
     \frac{ 2 \discount \varepsilon |\StSp| \|r\|_\infty}
     {(1-\discount)^3}.
  \]
\end{proposition}
Note that the result is presented slightly differently in \cite{Abel2016}. They
assume that $\|r\|_\infty = 1$ and simplify the above expression.

We now show an approximate model similarity is also an \acs{AIS} and
directly using the result of Theorem~\ref{thm:inf-ais} for this model gives a
stronger bound than Proposition~\ref{prop:Abel}.

\begin{proposition}\label{prop:model-similarity}
  Let $(q,w)$ be such that the aggregated model is $\varepsilon$-approximate
  model similar to the true model. Then, $(q, \hat P, \hat r)$ is an
  $(\varepsilon, \varepsilon |\hat \StSp|)$-\acs{AIS} with respect to the
  total variation distance.
\end{proposition}
\begin{proof}
  We first establish (AP1). For any $\st \in \StSp$ and $\act \in \ActSp$,
  \begin{align*}
    \bigl| r(\st, \act) - \hat r(q(\st), \act) \bigr| &
    \stackrel{(a)}\le
    \biggl|        
      \sum_{\tilde \st \in q^{-1}(q(\st))} w(\tilde \st)r(       \st, \act)
    - \sum_{\tilde \st \in q^{-1}(q(\st))} w(\tilde \st)r(\tilde \st, \act)
    \biggr| \\
    &\stackrel{(b)}\le
      \sum_{\tilde \st \in q^{-1}(q(\st))} w(\tilde \st)
      \bigl| r(       \st, \act) - r(\tilde \st, \act) \bigr|
    \\
    &\stackrel{(c)}\le \varepsilon
  \end{align*}
  where $(a)$ follows from the basic property of the weight function $w$
  and the definition of the aggregated reward $\hat r$; $(b)$ follows form the
  triangle inequality; and $(c)$ follows from the definition of approximate
  model similarity and the basic property of the weight function $w$. This
  establishes property (AP1).
    
  Now, we establish (AP2).
  Let $d_\F$ denote the total variation distance. Define probability measures
  $\mu, \nu$ on $\Delta(\hat \StSp)$ in the definition of (AP2), i.e., 
  for any $\st \in \StSp$, $\hat \st' \in \hat \StSp$, and $\act \in \ActSp$, 
  \begin{align*}
    \mu(\hat \st') &\coloneqq \sum_{\st' \in q^{-1}(\hat \st')} P(\st' | \st, \act)
    \\
    \nu(\hat \st') &\coloneqq \hat P(\hat \st' | q(\st), \act)
    = 
    \sum_{\tilde \st \in q^{-1}(q(\st))} \sum_{\st' \in q^{-1}(\hat \st')}
    P(\st'|\tilde \st,\act) w(\tilde \st)
  \end{align*}

  Now consider (see footnote~\ref{fnt:TV} on page~\pageref{fnt:TV})
  \begin{align*}
    d_\F(\mu, \nu) &= \sum_{\hat \st' \in \hat \StSp}
    | \mu(\hat \st') - \nu(\hat \st') |
    \\
    &= \sum_{\hat \st' \in \hat \StSp} \biggl|
      \sum_{\st' \in q^{-1}(\hat \st')} P(\st' | \st, \act)
      -
    \sum_{\tilde \st \in q^{-1}(q(\st))} \sum_{\st' \in q^{-1}(\hat \st')}
    P(\st'|\tilde \st,\act) w(\tilde \st)
    \biggr|
    \\
    &\stackrel{(a)}\le
    \sum_{\hat \st' \in \hat \StSp} 
    \sum_{\tilde \st \in q^{-1}(q(\st))} 
    w(\tilde \st)
    \biggl|
    \sum_{\st' \in q^{-1}(\hat \st')} P(\st' | \st,\act) 
    -
    \sum_{\st' \in q^{-1}(\hat \st')} P(\st' | \tilde \st,\act) 
    \biggr|
    \\
    &\stackrel{(b)}\le 
    \sum_{\hat \st' \in \hat \StSp} 
    \sum_{\tilde \st \in q^{-1}(q(\st))} 
    w(\tilde \st) \varepsilon
    \stackrel{(c)}=
    \sum_{\hat \st' \in \hat \StSp} \varepsilon
    = |\hat \StSp| \varepsilon,
  \end{align*}
  where $(a)$ follows from triangle inequality, $(b)$ follows from definition
  of approximate model similarity and $(c)$ follows from the basic property of
  the weight function. This proves (AP2).
\end{proof}

\begin{lemma}\label{lem:MS-V-bound}
  For any MDP
  \[
    \Span(V) \le \frac{\Span(r)}{1 - \discount}.
  \]
  Therefore, when $d_\F$ is the total variation distance, $\Minkowski_\F(V) \le
  \tfrac12 \Span(r)/(1-\discount)$. 
\end{lemma}
\begin{proof}
  This result follows immediately by observing that the per-step cost
  $r(\St_t, \Act_t) \in [ \min(r), \max(r) ]$. Therefore, $\max(V) \le
  \max(r)/(1-\discount)$ and $\min(V) \ge \min(r)/(1-\discount)$. 
\end{proof}

\begin{proposition}\label{prop:MS-ais}
  Let $\hat \pol$, $\pol$, $V$, and $V^\pol$ be defined as in
  Proposition~\ref{prop:model-similarity}. Then, for all $\st \in \StSp$,
  \[
    \bigl| V(\st) - V^\pol(\st) \bigr| \le
     \frac{2 \varepsilon}{(1 - \discount)} + 
     \frac{ \discount \varepsilon |\hat \StSp| \Span(r)}
     {(1-\discount)^2}.
  \]
\end{proposition}
\begin{proof}
  This follows immediately from Theorem~\ref{thm:inf-ais},
  Proposition~\ref{prop:model-similarity} and Lemma~\ref{lem:MS-V-bound}.
\end{proof}

Note that the error bounds of Propositions~\ref{prop:model-similarity}
and~\ref{prop:MS-ais} have similar structure but the key difference is that
the bound of Proposition~\ref{prop:MS-ais} is tighter than a factor of
$1/(1-\discount)$ as compared to Proposition~\ref{prop:model-similarity}.
There are other minor improvements as well ($|\hat \StSp|$ instead of
$|\StSp|$ and $\tfrac12\Span(r)$ instead of $\|r\|_\infty$).

\section{Comparison with the results of \texorpdfstring{\cite{DeepMDP}}{Gelada
  et al. (2019)} for latent space models for MDPs}\label{app:LipMDP}

\cite{DeepMDP} propose a latent space model for an MDP and show that
minimizing the losses in predicting the per-step reward and repredicting the
distribution over next latent space provides a bound on the quality of the
representation. In this section, we show that latent space representation
defined in \cite{DeepMDP} may be viewed as an instance of an AIS and show that
the approximation bounds of Theorem~\ref{thm:inf-ais} are similar to those
derived in \cite{DeepMDP}.

Since we follow a slightly different notation than \cite{DeepMDP} and for the
sake of completeness, we start by describing the notion of latent space
representation used in \cite{DeepMDP}. 

Consider an MDP with infinite horizon, finite-state and finite-action having state
space $\StSp$, action space $\ActSp$, transition probability matrix $P \colon
\StSp \times \ActSp \to \Delta(\StSp)$, per-step reward function $r \colon
\StSp \times \ActSp \to \reals$ and discount factor $\discount$. 

Let $(\hat \StSp, d)$ be a Banach space and it is assumed that we are given an
embedding function $\phi \colon \StSp \to \hat \StSp$, along with transition
dynamics $\hat P \colon \hat \StSp \times \ActSp \to \Delta(\hat \StSp)$ and reward
function $\hat r \colon \hat \StSp \times \ActSp \to \reals$. The MDP
$\widehat {\mathcal M} = (\hat \StSp,
\ActSp, \hat P, \hat r, \discount)$ along with the embedding function $\phi$
is called the \emph{latent space model} of the original MDP. 

\begin{definition}
The MDP $\widehat {\mathcal M}$ is said to be $(L_r, L_p)$-Lipschitz if for
any $\hat s_1, \hat s_2 \in \hat \StSp$ and $\act \in \Act$, 
\[
  \bigl| \hat r(\hat \st_1, \act) - \hat r(\hat \st_2, \act) \bigr| 
  \le L_r d(\hat \st_1, \hat \st_2) 
  \quad\text{and}\quad
  d_\F(\hat P(\cdot | \hat \st_1, \act), \hat P(\cdot | \hat \st_2, \act))
  \le L_p d(\hat s_1, \hat s_2),
\]
where $d_\F$ denotes the Kantorovich distance. 
\end{definition}

Given a latent space embedding, define 
\[
  \varepsilon = \sup_{\st \in \StSp, \act \in \ActSp}
  \bigl| r(\st, \act) - \hat r(\phi(\st), \act) \bigr| 
  \quad\text{and}\quad
  \delta = \sup_{\st \in \StSp, \act \in \ActSp}
  d_\F( \mu, \hat P(\cdot | \phi(\st), \act) ),
\]
where $\mu \in \Delta(\hat \StSp)$ given by $\mu(\ALPHABET B) = P(
\phi^{-1}(\ALPHABET B) | \st, \act)$ for any Borel subset $\ALPHABET B$ of
$\hat \StSp$. 

\begin{proposition}[Theorem 5 of \cite{DeepMDP}]\label{prop:DeepMDP}
  Let $\hat \pol \colon \hat \StSp \to \ActSp$ be the (deterministic)
  optimal policy of the latent space MDP. Define $\pol \colon \StSp \to
  \ActSp$ by $\pol = \hat \pol \circ \phi$. Let $V \colon \StSp \to \reals$
  denote the optimal value function and let $V^\pol \colon \StSp \to \reals$
  denote the value function for policy $\pol$. 

  If the latent space MDP $\widehat {\mathcal M}$ is $(L_r, L_p)$-Lipschitz,
  then,
  \[
    \bigl| V(\st) - V^\pol(\st) \bigr|  \le \frac{2\varepsilon}{1-\discount}
    + \frac{2\discount \delta L_r}{(1-\discount)(1 - \discount L_p)}. 
  \]
\end{proposition}

We show that a latent space model is an AIS and directly using the result
of Theorem~\ref{thm:inf-ais} gives the same approximation bound. 

\begin{proposition}
  Let $\widehat {\mathcal M} = (\hat \StSp, \ActSp, \hat P, \hat r,
  \discount)$ be a latent space model with embedding function $\phi$. Then, 
  $(\phi, \hat P, \hat r)$ is an $(\varepsilon, \delta)$-AIS with respect to
  the Kantorovich distance. 
\end{proposition}
\begin{proof}
  The result is an immediate consequence of the definition of $\varepsilon$ and
  $\delta$ for latent space model.
\end{proof}

\begin{lemma}\label{lem:LipMDP}
  For any $(L_r, L_p)$-Lipschitz MDP, if $\discount L_p < 1$, then
  \[
      \| V \|_\Lip \le \frac{L_r}{1 - \discount L_p}.
  \]
  Therefore, when $d_\F$ is the Kantorovich distance, $\Minkowski_\F(V) = \| V
  \|_{\Lip} \le L_r/(1 - \discount L_p)$. 
\end{lemma}
\begin{proof}
  This result follows immediately from Theorem 4.2 of \cite{Hinderer2005}.
\end{proof}

\begin{proposition}\label{prop:DeepMDP-ais}
  Let $\hat \pol$, $\pol$, $V$, and $V^\pol$ be defined in
  Proposition~\ref{prop:DeepMDP}. The, for all $s \in \StSp$,
  \[
    \bigl| V(\st) - V^\pol(\st) \bigr|  \le \frac{2\varepsilon}{1-\discount}
    + \frac{2\discount \delta L_r}{(1-\discount)(1 - \discount L_p)}. 
  \]
\end{proposition}
\begin{proof}
  This follows immediately from Theorem~\ref{thm:inf-ais},
  Proposition~\ref{prop:DeepMDP}, and Lemma~\ref{lem:LipMDP}.
\end{proof}

Note that the error bounds in Propositions~\ref{prop:DeepMDP}
and~\ref{prop:DeepMDP-ais} are exactly the same. 

\section{Comparison with the results of
  \texorpdfstring{\cite{FrancoisLavet2019}}{Francois-Lavet et al. (2019)} for belief
approximation in POMDPs}\label{app:FrancoisLavet}

\cite{FrancoisLavet2019} analyze the trade off between asymptotic bias and
overfitting in reinforcement learning with partial observations. As part of
their analysis, they express the quality of state representation in terms of
the bounds on the $L_1$ error of the associated belief states. We show that
these approximation bounds may be viewed as an instance of AIS-based bounds of
Theorems~\ref{thm:ais} and~\ref{thm:inf-ais}. We also show that the bounds of
Theorem~\ref{thm:inf-ais} for this model are stronger than those derived in
\cite{FrancoisLavet2019} by a factor of $\mathcal O(1/(1-\discount))$. 

Since we follow a slightly different notation than \cite{FrancoisLavet2019}
and for the sake of completeness, we start by describing the notion of
$\varepsilon$-sufficient statistics defined in \cite{FrancoisLavet2019}. 

Consider an infinite-horizon finite-state finite-action POMDP with state space
$\StSp$, action space $\ActSp$, observation space $\ObSp$, transition
probability matrix $P \colon \StSp \times \ActSp \to \Delta(\StSp)$,
observation matrix $P^\ob \colon \StSp \to \Delta(\ObSp)$, per-step reward
$r \colon \StSp \times \ActSp \to \reals$, and discount factor $\discount$. 

\begin{definition}[$\varepsilon$-sufficient statistic
  \citep{FrancoisLavet2019}]
  Given a family of Banach spaces $\{\Phi_t\}_{t=1}^L$, an
  $\varepsilon$-sufficient statistic is a collection of history
  compression function $\{ \phi_t \colon \HisSp_t \to \Phi_t \}_{t=1}^T$ and
  belief approximation functions $\{ \hat b_t \colon \Phi_t \to \Delta(\StSp)
  \}_{t=1}^T$ such that for any time~$t$ and any realization $\his_t$ of
  $\His_t$, we have
  \[
    \| \hat b_t(\cdot | \phi_t(\his_t)) - b_t (\cdot | h_t) \|_1 \le
    \varepsilon.
  \]
\end{definition}

Given an $\varepsilon$-sufficient statistic, \cite{FrancoisLavet2019} define
an MDP with state space $\Delta(\StSp)$, action space $\ActSp$, transition
probability kernel $\PR( \hat b_{t+1}(\cdot | \phi(\his_{t+1})) \mid \hat
b_t(\cdot | \phi(\his_t)), \act_t)$ computed from the underlying POMDP, 
and per-step reward given by 
\[
  \hat r(\hat b_t(\his_t), \act_t) = 
  \sum_{\st \in \StSp} r(\st, \act_t) \hat b_t(\st | \phi(\his_t)).
\]

\begin{proposition}[Theorem 1 of \cite{FrancoisLavet2019}]
  \label{prop:FrancoisLavet}
  Let $\{(\hat b_t, \phi_t)\}_{t=1}^T$ be an $\varepsilon$-sufficient statistic and $\hat
  \pol = (\hat \pol_1, \hat \pol_2, \dots)$ be an optimal policy for the MDP
  described above. Define a policy $\pol = (\pol_1, \pol_2, \dots)$ given by
  $\pol_t = \hat \pol_t \circ \phi_t$. Let $V_t \colon \HisSp_t \to \reals$
  denote the optimal value functions and $\hat V^\pol_t \colon \His_t \to
  \reals$ denote the value function for policy $\pol$. Then for any initial
  history $\his_1 \in \HisSp_1$,
  \[
    \bigl| V_1(\his_1) - V_1^\pol(\his_1) \bigr| \le 
    \frac{2 \varepsilon \|r\|_{\infty}}{(1-\discount)^3}.
  \]
\end{proposition}

We now show that an $\varepsilon$-sufficient statistic gives rise to an
\acs{AIS} and directly using the results of Theorem~\ref{thm:inf-ais} for this
model gives a stronger bound than Proposition~\ref{prop:FrancoisLavet}.

\begin{proposition}\label{prop:POMDP-ais}
  Let $\{(\hat b_t, \phi_t)\}_{t=1}^T$ be an $\varepsilon$-sufficient
  statistic. Let $\AisSp_t = \Delta(\StSp)$ and define the different
  components of an AIS as follows:
  \begin{itemize}[nosep]
    \item history compression functions $\ainfo_t = \hat b_t \circ \phi_t$,
    \item  \acs{AIS} prediction kernels $\nextinfo_t(\cdot
      | \ais_t, \act_t)$ is given by
      \[
        \nextinfo_t(\ALPHABET B | \ais_t, \act_t) 
      = \sum_{\ob_{t+1} \in \ObSp} \psi(\ob_{t+1} | \ais_t, \act_t)
      \IND_{\ALPHABET B}\{ \aupdate(\ais_t, \ob_{t+1}, \act_t) \},
    \]
    where
    \[
      \psi(\ob_{t+1} | \ais_t, \act_t) = 
      \sum_{\st_{t+1} \in \StSp} \sum_{\st_t \in \StSp}
      P^\ob(\ob_{t+1} | \st_{t+1}) P(\st_{t+1} | \st_t, \act_t) \ais_t(\st_t)
    \]
    and
    \[
      \aupdate(\ais_t, \ob_{t+1}, \act_t)(\st_{t+1}) = 
          \frac{ \sum_{\st_t \in \StSp} P^\ob(\ob_{t+1} | \st_{t+1}) P(\st_{t+1} | \st_t, \act_t)
          \ais_t(\st_t) }
        {\psi(\ob_{t+1} | \ais_t, \act_t)},
      \]
      where $\aupdate$ is the same as the Bayes'-rule based update of the belief
      state,
    \item reward approximation functions
      $\rewinfo(\ais_t, \act_t) = \sum_{\st \in \St}\ais_t(\st) r(\st, \act_t)$.
  \end{itemize}
  Then, $\{(\ainfo_t, \nextinfo_t, \rewinfo_t)\}_{t=1}^T$ is an $(\varepsilon
    \|r\|_\infty, 3\varepsilon)$-AIS with respect to the bounded-Lipschitz metric.
\end{proposition}
\begin{proof}
  We need to equip $\AisSp = \Delta(\StSp)$ with a metric in order to define a
  bounded-Lipschitz metric over $\Delta(\AisSp)$. We use the total variation as
  the metric and denote it by $d_\TV$. We use $\F$ to denote $\{ f \colon
  \Delta(\AisSp) \to \reals : \| f \|_{\infty} + \| f \|_\Lip \le 1\}$ and
  denote the corresponding bounded-Lipschitz metric over
  $\Delta(\AisSp)$ by $d_\F$. 

  We first establish (AP1). For any time~$t$, realization $\his_t$ of
  history $\His_t$, and action $\act_t \in \ActSp$, we have
  \begin{align*}
    \bigl|
    \EXP[ r(\St_t, \act_t) &\mid \His_t = \his_t, \Act_t = \act_t] - 
    \rewinfo_t(\ainfo_t(\his_t), \act_t) 
    \bigr|
    \\
    &= \biggl|
      \sum_{\st \in \StSp} r(\st, \act_t) b_t(\st | \his_t)
      -
      \sum_{\st \in \StSp} r(\st, \act_t) \hat b_t(\st | \phi(\his_t))
    \biggr|
     \\ 
   &\stackrel{(a)}\le \|r\|_\infty d_\TV(b_t, \hat b_t) 
   \\
   &\stackrel{(b)}\le \varepsilon \|r\|_\infty
  \end{align*}
  where $(a)$ follows from~\eqref{eq:minkowski} and the fact that for total
  variation distance $\Minkowski_\TV(r) \le \|r\|_\infty$; and $(b)$ follows from
  definition of $\varepsilon$-sufficient statistic.

  Before  establishing (AP2), we 
  note that $\aupdate$ is the Bayes'-rule based update of the true belief;
  therefore, 
  \[
    b_{t+1}(\cdot | \his_{t+1}) = \aupdate(b_t(\cdot | \his_t), \ob_{t+1}, 
    \act_t).
  \]
  For ease of notation, we use $b_t(\cdot)$ and $\hat b_t(\cdot)$ instead of
  $b_t(\cdot | \his_t)$ and $\hat b_t(\cdot | \phi_t(\his_t))$, when the
  conditioning is clear from context.

  Now consider $\mu_t$ and $\nu_t$ as defined in the definition of (AP2). In
  particular, for any Borel set $\ALPHABET B$, 
  \begin{align*}
    \mu_t(\ALPHABET B) 
      &= \sum_{\ob_{t+1} \in \ObSp} \psi(\ob_{t+1} | b_t, \act_t)
      \IND_{\ALPHABET B}\{ \hat b_{t+1}(\cdot | \phi(\his_t, \ob_{t+1},
      \act_t)) \}
    \\
    \nu_t(\ALPHABET B) &= \nextinfo_t(\ALPHABET B | \ais_t, \act_t).
  \end{align*}
  We also define an additional measure $\xi_t$ given by
  \[
    \xi_t(\ALPHABET B) 
      = \sum_{\ob_{t+1} \in \ObSp} \psi(\ob_{t+1} | b_t, \act_t)
      \IND_{\ALPHABET B}\{ \aupdate(b_t, \ob_{t+1}, \act_t) \},
  \]

  Now, by the triangle inequality
  \begin{equation} \label{eq:POMDP-diff}
    d_\F(\mu_t, \nu_t) \le d_\F(\mu_t, \xi_t) + d_\F(\xi_t, \nu_t).
  \end{equation}
  Now consider the first term of~\eqref{eq:POMDP-diff}:
  \begin{align}
    d_\F(\mu_t, \xi_t) &= \sup_{f \in \F} \biggl|
    \int_{\AisSp} f d\mu_t - \int_{\AisSp} f d\xi_t \biggr|
    \notag\\
    &= \sup_{f \in \F} \biggl|
    \sum_{\ob_{t+1} \in \ObSp} f(\hat b_{t+1}(\cdot | \phi(\his_t, \ob_{t+1}, \act_t))) 
      \psi(\ob_{t+1} | b_t, \act_t)
    \notag \\
    \displaybreak[1]
    &\qquad \qquad
    - 
    \sum_{\ob_{t+1} \in \ObSp} f(b_{t+1}(\cdot | \his_t, \ob_{t+1}, \act_t)) 
      \psi(\ob_{t+1} | b_t, \act_t)
    \biggr|
    \notag \\
    &\stackrel{(a)}\le
    \sum_{\ob_{t+1} \in \ObSp}
    d_\TV(\hat b_{t+1}(\cdot | \phi(\his_{t+1})), b_{t+1}(\cdot | \his_{t+1}))
    \psi(\ob_{t+1} | \his_t, \act_t)
    \notag\\
    &\stackrel{(b)}\le \varepsilon
    \label{eq:POMDP-diff-1}
  \end{align}
  where $(a)$ follows from triangle inequality and the fact that slope of $f$
  is bounded by $1$; and $(b)$ follows from the definition of
  $\varepsilon$-sufficient statistic (see footnote~\ref{fnt:TV} on
  page~\pageref{fnt:TV}). 
  Now consider the second term of~\eqref{eq:POMDP-diff} (for ease of notation, 
  we use $b_t(\cdot)$ instead of $b_t(\cdot | \his_t)$):
  \begin{align}
    d_\F(\xi_t, \nu_t) &= \sup_{f \in \F} \biggl|
    \int_{\AisSp} f d\xi_t - \int_{\AisSp} f d\nu_t \biggr|
    \notag\\
    \displaybreak[1]
    &= \sup_{f \in \F} \biggl|
    \sum_{\ob_{t+1} \in \ObSp} f(\aupdate(b_t, \ob_{t+1}, \act_t)) 
      \psi(\ob_{t+1} | b_t, \act_t)
    - 
    \sum_{\ob_{t+1} \in \ObSp} f(\aupdate(\ais_t, \ob_{t+1}, \act_t)) 
      \psi(\ob_{t+1} | \ais_t, \act_t)
    \biggr|
    \notag \\
    &\stackrel{(c)}\le
    \sup_{f \in \F} \biggl|
    \sum_{\ob_{t+1} \in \ObSp} f(\aupdate(b_t, \ob_{t+1}, \act_t)) 
      \psi(\ob_{t+1} | b_t, \act_t)
    - 
    \sum_{\ob_{t+1} \in \ObSp} f(\aupdate(\ais_t, \ob_{t+1}, \act_t)) 
      \psi(\ob_{t+1} | b_t, \act_t)
    \biggr|
    \notag \\
    \displaybreak[1]
    &\quad + \sup_{f \in \F} \biggl|
    \sum_{\ob_{t+1} \in \ObSp} f(\aupdate(\ais_t, \ob_{t+1}, \act_t)) 
      \psi(\ob_{t+1} | b_t, \act_t)
    - 
    \sum_{\ob_{t+1} \in \ObSp} f(\aupdate(\ais_t, \ob_{t+1}, \act_t)) 
      \psi(\ob_{t+1} | \ais_t, \act_t)
    \biggr|
    \notag \\
    &\stackrel{(d)}\le
    \sum_{\ob_{t+1} \in \ObSp}
    d_\TV( \aupdate(b_t,\ob_{t+1}, \act_t), 
      \aupdate(\ais_t, \ob_{t+1}, \act_t)) 
    \psi(\ob_{t+1} | b_t, \act_t) 
    \notag \\
    &\quad + 
    \Contraction_{\F,\TV}(\aupdate(\ais_t, \cdot, \act_t)) 
    \,
    d_\TV(\psi( \cdot | b_t, \act_t), 
    \psi( \cdot | \ais_t, \act_t)),
    \label{eq:POMDP-diff-2}
  \end{align}
  where $(c)$ follows from the triangle inequality; the first step of
  $(d)$ follows from an argument similar to step $(a)$
  of~\eqref{eq:POMDP-diff-1}; and the second part of $(d)$ follows
  from~\eqref{eq:contraction}. 
  
  Now, we obtain bounds for both terms of~\eqref{eq:POMDP-diff-2}. 
  For $\ob_{t+1} \in \ObSp$, define 
  \begin{align*}
    \xi^\ob_t(\ob_{t+1} ) &\coloneqq \psi(\ob_{t+1} | b_t, \act_t) = 
    \sum_{\st_{t+1} \in \StSp} \sum_{\st_t \in \StSp} 
    P^\ob(\ob_{t+1} | \st_{t+1}) P(\st_{t+1} | \st_t, \act_t) b_t(\st_t | \his_t), 
    \\
    \nu^\ob_t(\ob_{t+1} ) &\coloneqq \psi(\ob_{t+1} | \ais_t, \act_t) = 
    \sum_{\st_{t+1} \in \StSp} \sum_{\st_t \in \StSp} 
    P^\ob(\ob_{t+1} | \st_{t+1}) P(\st_{t+1} | \st_t, \act_t) \ais_t(\st_t), 
  \end{align*}
  
  Total variation is also an $f$-divergence\footnote{%
    Let $f \colon \reals_{\ge 0} \to \reals$ be a convex function such that
    $f(1) = 0$. Then the $f$-divergence between two measures $\mu$ and $\nu$
    defined on a measurable space $\ALPHABET X$ is given by 
    \[
      D_f(\mu \| \nu) = \int_{\ALPHABET X} f\Bigl( \frac{d\mu}{d\nu} \Bigr) d\nu.
    \]
    Total variation is a $f$-divergence with $f(x) = \lvert x - 1 \rvert$
    (also see footnote~\ref{fnt:TV} on page~\ref{fnt:TV}).
    \cite{Sriperumbudur2009} showed that total variation is the only non-trivial
    IPM which is also an $f$-divergence.}
  therefore, it satisfies the strong data processing inequality.\footnote{Let
    $\ALPHABET X$ and $\ALPHABET Y$ be measurable spaces, $\mu$ and $\nu$
    be measures on $\ALPHABET X$ and $P \colon \ALPHABET X \to
    \Delta(\ALPHABET Y)$ be a stochastic kernel from $\ALPHABET X$ to
    $\ALPHABET Y$. We use $\mu  P$ to denote the
    measure $\mu_{\ALPHABET Y}$ on $\ALPHABET Y$ given by $\mu_{\ALPHABET
    Y}(dy) = \int_{\ALPHABET X} P(dy|x) \mu(dx)$. Similar interpretation holds
    for $\nu  P$. Then, the \emph{strong data processing inequality}
    \citep{Sason2019} states that for any
    $f$-divergence, \( D_f(\mu  P \| \nu  P) \le D_f(\mu \| \nu)\).}
    Note that the
  definition of both $\xi^\ob_t$ and $\nu^\ob_t$ may be viewed as outputs of a
  ``channel'' from $\st_t$ to $\ob_{t+1}$.  In case of $\xi^\ob_t$, the
  channel input is distributed according to $b_t(\cdot | \his_t)$ and in case
  of $\nu^\ob_t$, the channel input is distributed according to $\ais_t$.
  Therefore, from the data processing inequality, 
  \begin{equation} \label{eq:data-processing-1}
    d_\TV(\xi^\ob_t, \nu^\ob_t) \le d_\TV(b_t(\cdot | \his_t), \ais_t)
    \le \varepsilon
  \end{equation}
  where the last inequality follows from the definition of
  $\varepsilon$-sufficient statistic.

  A similar argument can be used to bound 
  \(
    d_\TV( \aupdate(b_t,\ob_{t+1}, \act_t), 
      \aupdate(\ais_t, \ob_{t+1}, \act_t))
  \). In particular, we can think of $\aupdate(\cdot, \ob_{t+1}, \act_t)$ as a
  channel from $\st_t$ to $\st_{t+1}$. Then, by the data processing
  inequality, 
  \begin{equation}\label{eq:data-processing-2}
    d_\TV( \aupdate(b_t,\ob_{t+1}, \act_t), 
      \aupdate(\ais_t, \ob_{t+1}, \act_t))
    \le d_\TV(b_t, \ais_t) \le \varepsilon
  \end{equation}
  where the last inequality follows from the definition of
  $\varepsilon$-sufficient statistic. 

  The final part of~\eqref{eq:POMDP-diff-2} that needs to be characterized is
  $\Contraction_{\F, \TV}(\aupdate(\ais_t, \cdot, \act_t))$.
  From~\eqref{eq:F-contraction}
  \[
    \Contraction_{\F, \TV}(\aupdate(\ais_t, \cdot, \act_t)) = 
    \sup_{f \in \F} \Minkowski_\TV(f \circ \aupdate(\ais_t, \cdot, \act_t)) 
    \le
    \sup_{f \in \F} \| f \circ \aupdate(\ais_t, \cdot, \act_t) \|_{\infty}
    \le 1.
  \]
  Substituting this bound along with~\eqref{eq:data-processing-1}
  and~\eqref{eq:data-processing-2} in~\eqref{eq:POMDP-diff-2}, we get
  $d_\F(\xi_t, \nu_t) \le 2\varepsilon$. Substituting this along
  with~\eqref{eq:POMDP-diff-1} in~\eqref{eq:POMDP-diff}, we get that
  $d_\F(\mu_t, \nu_t) \le 3\varepsilon$. Hence (AP2) is satisfied.
\end{proof}

\begin{lemma}\label{lem:POMDP-bound}
  For any POMDP,
  \[
    \Minkowski_\F(V) = \| V \|_\infty + \| V \|_\Lip \le 
    \frac{2 \| r \|_\infty }{1 - \discount}.
  \]
\end{lemma}
\begin{proof}
  The result follows immediately from the sup norm on the value function
  (Lemma~\ref{lem:MS-V-bound} and the bounds on the Lipschitz constant of the
  value function (Lemma~1 of \cite{Lee2008}).
\end{proof}

\begin{proposition}\label{prop:POMDP-bound}
  Let $\hat \pol$, $\pol$, $V$, and $V^\pol$ be as defined in
  Proposition~\ref{prop:FrancoisLavet}. Then, for any initial history $\his_1 \in
  \HisSp_1$, 
  \[
    \bigl| V(\his_1) - V^\pi(\his_1) \bigr| \le 
    \frac{2 \varepsilon \| r \|_{\infty} }{(1-\discount)}
    + 
    \frac{6 \discount \varepsilon \| r\|_{\infty}}{(1-\discount)^2}.
  \]
\end{proposition}
\begin{proof}
  This follows immediately from Theorem~\ref{thm:inf-ais},
  Proposition~\ref{prop:POMDP-ais}, and Lemma~\ref{lem:POMDP-bound}.
\end{proof}

Note that the error bounds of Propositions~\ref{prop:FrancoisLavet}
and~\ref{prop:POMDP-bound} have similar structure but the key difference is
that the bound of Proposition~\ref{prop:POMDP-bound} is tighter by a factor of
$1/(1-\discount)$. 

\section{Comparison with the results of
  \texorpdfstring{\cite{chandak2020lifelong}}{Chandak et al. (2020)} on lifelong
learning for time-varying action spaces}\label{app:ais-ac-only}

Lifelong learning refers to settings where a reinforcement learning agent
adapts to a time-varying environment. There are various models for lifelong
learning and \cite{chandak2020lifelong} recently proposed a model where the
action spaces change over time. The environment has an underlying finite state
space $\StSp$, finite action space $\ActSp$, and reward $r \colon \StSp \to
\reals$. Note that the reward depends only on the current state and not the
current action. 

It is assumed that there is an underlying finite dimensional representation
space $\ALPHABET E$ and for any feasible action $\act \in \ActSp$, there is an
underlying representation $e \in \ALPHABET E$. This relationship is captured
via an invertible map $\phi$, i.e., $a = \phi(e)$. There is a transition
kernel $P \colon \StSp \times \ALPHABET E \to \Delta(\StSp)$ with respect to
this representation space. This induces a transition kernel $P^a \colon \StSp
\times \ActSp \to \Delta(\StSp)$ with respect to
the action, where $P^a(s'|s,a) = P(s'|s,\phi^{-1}(a))$. It is assumed that the
transition kernel $P$ is $\rho$-Lipschitz, i.e., for all $\st,\st' \in \StSp$
and $e_i, e_j \in \ALPHABET E$,
\[
  \bigl\| P(s' | s, e_i) - P(s' | s, e_j) \bigr\|_{1} \le 
  \rho \| e_i - e_j \|_{1}.
\]
\cite{chandak2020lifelong} consider infinite horizon discounted
setup with discount factor $\discount$.

Initially, the RL agent is not aware of the action space and learns about the
actions in discrete stages indexed by $k \in \integers_{\ge 0}$. At stage~$k$,
the agent becomes aware of a subset $\ALPHABET U_k$ of $\ALPHABET E$, where
$\ALPHABET U_k \supseteq \ALPHABET U_{k-1}$. Thus, the environment at stage
$k$ may be modelled as an MDP $\mathcal{M}_k = \{ \StSp, \ActSp_k, P^{a}_k, r \}$,
where $\ActSp_k = \{ \phi(e) : e \in \ALPHABET U_k \}$ and $P^{a}_k(\st' | \st,
\act) = P(\st' | \st, \phi^{-1}(a))$.

Two main results are established in \cite{chandak2020lifelong}. The first one
is the following.

\begin{proposition}[Theorem 1 of \cite{chandak2020lifelong}]\label{prop:chandak1}
  Let $\pi_k$ and $V^{\pi_k}$ denote the optimal policy for MDP
  $\mathcal{M}_k$ and its performance. Let $V$ denote the value function for
  the hypothetical model when the agent has access to all actions. Let
  \[
    \eta_k = \sup_{a_i, a_j \in \ActSp_k}
    \| \phi^{-1}(a_i) - \phi^{-1}(a_j) \|_1.
  \]
  Then, for any $s \in \StSp$,
  \[
    V(s) - V^{\pi_k}(s) \le \frac{\discount \rho
    \eta_k \| r \|_{\infty}}{(1-\discount)^2}.
 \]
\end{proposition}

We now show that this result may be viewed as a corollary of 
Corollary~\ref{cor:ais-ac-only}. In particular, we have the following.

\begin{lemma}\label{lem:chandak1}
  The action set $\ActSp_k$ may be viewed as a ``quantization'' of $\ActSp$
  using a function $\aquant \colon \ActSp \to \ActSp_k$, which maps any
  action $a = \phi(e) \in \ActSp$ to action $a' = \phi(e') \in \ActSp_k$ such
  that $e' = \arg \min_{e'' \in \ALPHABET U_k} \| e - e'' \|_{1}$. Then, 
  $\aquant$ is $(0, \rho \eta_k)$-action-quantizer with respect to the
  total variation distance. 
\end{lemma}
\begin{proof}
  Since the per-step reward does not depend on the action, there is no
  approximation error in the reward and, therefore, $\varepsilon = 0$. Now
  note that for any $s \in \StSp$ and $a \in \ActSp$, we have
  \[
    d_{\TV}(P^a(\cdot | s, a), P^a(\cdot | s, \aquant(a))) \le 
    \sup_{e_i, e_j \in \ALPHABET U_k}
    \| P(\cdot | s, e_i) - P(\cdot | s, e_j) \|_{1}
    \le \rho \eta_k
  \]
  where the last equality follows from the $\rho$-Lipschitz continuity of
  $P$ and the definition of $\eta_k$. Thus, $\delta = \rho \eta_k$.
\end{proof}

\begin{proposition}\label{prop:chandak2}
  Let $\pi_k$, $V^{\pi_k}$ and $V$ be as defined in
  Proposition~\ref{prop:chandak1}. 
  Then, for any $s \in \StSp$,
  \[
    V(s) - V^{\pi_k}(s) \le \frac{\discount \rho
    \eta_k \Span(r)}{2 (1-\discount)^2} 
 \]
\end{proposition}
\begin{proof}
  The result can be established from the following observations: (i)~The
  result of Corollary~\ref{cor:ais-ac-only} continues to hold in the infinite horizon 
  discount reward setup with $\alpha_t$ replaced by $(\varepsilon + \discount
  \Minkowski_\F(\hat V^*) \delta)/(1-\discount)$. This can be established in a
  manner similar to Theorem~\ref{thm:inf-ais}. 
  (ii)~From Lemma~\ref{lem:MS-V-bound}, we know that for total variation
  distance $\Minkowski_\F(\hat V^*) \le \frac12 \Span(r)/(1 - \discount)$. 
  The result follows from substituting the values of
  $(\varepsilon,\delta)$ from Lemma~\ref{lem:chandak1} and the value of
  $\Minkowski_\F(\hat V^*)$ from~(ii) in~(i).
\end{proof}

Note that if the rewards $r(s)$ belongs in a symmetric interval, say
$[-R_{\max}, R_{\max}]$, as is assumed in \cite{chandak2020lifelong}, the result
of Proposition~\ref{prop:chandak2} matches that of Proposition~\ref{prop:chandak1}.

The second result of \cite{chandak2020lifelong} is for the setting when the
mapping $\phi$ is not known. They assume that the agent selects some finite
dimensional representation $\hat{\ALPHABET E}$ and, for every~$k$,
parameterizes the policy using two components: (i) a map $\beta \colon \StSp
\to \Delta(\hat{\ALPHABET E})$ 
and (ii)~an estimator $\hat{\phi}_k \colon \hat{\ALPHABET E} \to
\Delta(\ActSp_k)$. Then the action at any state $S_t \in \StSp$ is chosen by
first sampling $\hat e \sim \beta(s)$ and then choosing the action $a \sim
\hat{\phi}_k(\hat e)$. 
The second main result in \cite{chandak2020lifelong} is the
following.\footnote{This result is stated slightly different in
  \cite{chandak2020lifelong} using an \emph{inverse dynamics}
function $\varphi \colon \StSp \times \StSp \to \Delta(\hat {\ALPHABET E})$,
where $e \sim \varphi(s, s')$ is a prediction of a latent action $e$ which
might have caused the transition from $s$ to $s'$. However, the bounds hold for the
simpler form presented here as well.}

\begin{proposition}[Theorem 2 of \cite{chandak2020lifelong}]\label{prop:chandak3}
  Let $\hat \pi_k$ denote the best overall policy that
  can be represented using the above structure, $V^{\hat \pi_k}$ denotes its
  performance, and $V$ denote the value function when the agent has access to
  the complete model. Suppose there exists a $\zeta \in \reals_{\ge 0}, \beta \colon \StSp 
  \to \Delta(\hat{\ALPHABET E})$ and $\hat \phi_k \colon \hat {\ALPHABET
  E} \to \Delta(\ActSp_k)$, such that for 
  \[
    \sup_{s \in \StSp, a_k \in \ActSp_k}
    \KL( P^a(\cdot | s, a_k) \| P^a(\cdot | s, \hat A) ) \le \zeta_k^2/2,
  \]
  where $\hat A \sim \hat \phi_k(\hat E)$ and $\hat E \sim \beta(s)$. 
  Then, for any $s \in \StSp$,
  \[
    V(s) - V^{\pi_k}(s) \le \frac{\discount (\rho \eta_k + \zeta_k) \| r \|_{\infty}}{(1-\discount)^2}.
 \]
\end{proposition}

We now show that this result may be viewed as a corollary of 
Corollary~\ref{cor:ais-ac-only}. In particular, we have the following.

\begin{lemma}\label{lem:chandak2}
  The action set $\hat {\ALPHABET E}$ may be viewed as a ``compression'' of
  the ``quantized'' action set $\ActSp_k$. In particular, let $\aquant \colon
  \ActSp \to \ActSp_k$ be as defined in Lemma~\ref{lem:chandak1}. Then, the
  function $\hat \phi_k^{-1} \circ \aquant$ is a $(0, \rho \eta_k + \zeta_k)$-action-
  quantizer with respect to the total variation distance.
\end{lemma}
\begin{proof}
  As argued in the proof of Lemma~\ref{lem:chandak1}, since the reward
  function does not depend on action, $\varepsilon = 0$. Now, recall that from
  Pinsker's inequality, for any distribution $\mu$ and $\nu$, $d_\TV(\mu, \nu)
  \le \sqrt{ 2 D_{\KL}(\mu \| \nu)}$. Thus,
  \[
    \sup_{s \in \StSp, a_k \in \ActSp_k}
    d_\TV( P^a(\cdot | s, a_k) , P^a(\cdot | s, \hat A) ) \le \zeta_k
  \]
  where $\hat A \sim \hat \phi_k(\hat E)$ and $\hat E \sim \beta(s)$. 
  Now, by the triangle inequality, for any $s \in \StSp$ and $a \in \ActSp$
  \begin{align*}
    d_\TV( P^a(\cdot | s, a) , P^a(\cdot | s, \hat A) ) 
    &\le 
    d_\TV( P^a(\cdot | s, a) , P^a(\cdot | s, \aquant(a)) ) + 
    d_\TV( P^a(\cdot | s, \aquant(a)) , P^a(\cdot | s, \hat A) ) 
    \\
    &\le
    \rho \eta_k + \zeta_k,
  \end{align*}
  Thus, $\delta = \rho \eta_k + \zeta_k$.
\end{proof}

\begin{proposition}\label{prop:chandak4}
  Let $\hat \pi_k$, $V^{\hat \pi_k}$ and $V$ be as defined in
  Proposition~\ref{prop:chandak1}. 
  Then, for any $s \in \StSp$,
  \[
    V(s) - V^{\hat \pi_k}(s) \le \frac{\discount (\rho \eta_k + \zeta_k) \Span(r)}{2 (1-\discount)^2} 
 \]
\end{proposition}
\begin{proof}
  The proof is similar to the proof of Proposition~\ref{prop:chandak2}, where
  we replace the values of Lemma~\ref{lem:chandak1} with those of
  Lemma~\ref{lem:chandak2}.
\end{proof}

As before, if the rewards $r(s)$ belongs in a symmetric interval, say
$[-R_{\max}, R_{\max}]$, as is assumed in \cite{chandak2020lifelong}, the result
of Proposition~\ref{prop:chandak4} matches that of Proposition~\ref{prop:chandak3}.

{\color{black}
\section{Convergence of the PORL algorithm}
\label{sec:porl-convergence}

In this section, we discuss the convergence of the PORL algorithm presented in 
Sec.~\ref{sec:grad-ascent} and~\ref{sec:PORL}. The proof of convergence relies
on multi-timescale stochastic approximation~\cite{Borkar_1997} under
conditions similar to the standard conditions for convergence of policy
gradient algorithms with function approximation stated
below:

\begin{assumption} \label{ais_assmpt}
  The following conditions are satisfied:
  \begin{enumerate}[leftmargin=1pc,nosep]
    \item All network parameters $(\bar \xi_k, \zeta_k, \theta_k)$
      lie in convex and bounded subsets of Euclidean spaces.
    \item The gradient of the loss function $\GRAD_{\bar \xi}\mathcal{L}(\bar \xi_k)$ of the state approximator is
      Lipschitz in $\bar \xi_k$, the gradient of the TD loss $\GRAD_{\zeta}\mathcal{L}_{\text{TD}}(\bar \xi_k, \theta_k, \zeta_k)$
      and the policy gradient $\widehat \GRAD_{\theta_k} J(\bar \xi_k,
      \theta_k, \zeta_k)$ is Lipschitz in $(\bar \xi_k, \theta_k, \zeta_k)$
      with respect to the sup norm.
    \item All the gradients---$\GRAD_{\bar \xi}\mathcal{L}(\bar \xi_k)$ at the state
      approximator; $\GRAD_{\zeta}\mathcal{L}_{\text{TD}}(\bar \xi_k, \theta_k, \zeta_k)$ at the critic; 
      and $\widehat \GRAD_{\theta_k}J(\bar \xi_k, \theta_k, \zeta_k)$ at the actor---are unbiased with bounded variance. Furthermore, the critic and the actor function approximators are compatible as given in~\cite{Sutton2000}, i.e., 
      \[
      \frac{\partial Q_{\zeta_k}(\Ais_t, \Act_t)}{\partial \zeta} = \frac{1}{\pol_{\theta_k}(\Ais_t, \Act_t)}\frac{\partial \pol_{\theta_k}(\Ais_t, \Act_t)}{\partial \theta}.
      \]
    \item The learning rates are sequences of positive numbers $\{a_k\}_{k \ge 0}, \{b_k\}_{k \ge 0}, \{c_k\}_{k \ge 0}$ that satisfy:
      \(
        \sum a_k = \infty, 
      \)
      \(
        \sum b_k = \infty, 
      \)
      \(
        \sum c_k = \infty, 
      \)
      \(
       \sum a_k^2 < \infty,
     \)
      \(
       \sum b_k^2 < \infty,
     \)
      \(
       \sum c_k^2 < \infty,
     \)
      \(
        \lim_{k \to \infty} c_k/a_k = 0,
      \)
      and
      \(
        \lim_{k \to \infty} b_k/c_k = 0.
      \)
  \end{enumerate}
\end{assumption}

\begin{assumption}\label{ass:regularity}
  The following regularity conditions hold:
  \begin{enumerate}
    \item The ODE
      corresponding to $\theta$ in~\eqref{eq:ais_ac} is locally asymptotically
      stable.
    \item The ODEs corresponding to $\bar \xi$ and $\zeta$ in~\eqref{eq:ais_ac} 
      are globally asymptotically stable. In addition, the ODE corresponding to
      $\zeta$ has a fixed point which is 
      Lipschitz continuous in $\theta$. 
  \end{enumerate}
\end{assumption}

The proposed RL framework has the following convergence guarantees.

\begin{theorem} \label{thm:ais_rl_convergence}
  Under Assumptions~\ref{ais_assmpt} and~\ref{ass:regularity},
  along any sample path, almost surely we have the following:
  \begin{enumerate}
    \item[\textup{(a)}] The iteration for $\bar \xi$ in~\eqref{eq:ais_ac}
      converges to a state estimator that minimizes the loss function
      $\mathcal L(\bar \xi)$;
    \item[\textup{(b)}] The iteration for $\zeta$ in~\eqref{eq:ais_ac} converges
      to a critic that minimizes the error with respect to the true
      $Q$-function; 
    \item[\textup{(c)}] The iteration for $\theta$ in~\eqref{eq:ais_ac}
      converges to a local maximum of the performance $J(\bar \xi^*, \zeta^*,
      \theta)$, where $\bar \xi^*$ and $\zeta^*$ are the converged values of
      $\bar \xi$ and $\zeta$.
  \end{enumerate}
\end{theorem}
\begin{proof}
  The assumptions satisfy all the four conditions stated in~\cite[page 35]{Leslie_2004},
  \cite[Theorem 23]{Borkar_1997}. The proof follows from combining this two-time scale algorithm proof with the fastest third time-scale of learning the state representation. Due to the specific choice of learning rates, the state representation algorithm sees a stationary actor and critic, while the actor and critic in turn see a converged state approximator ietration due to its faster learning rate. The convergence of the state approximator follows from~\cite[Theorem
  2.2]{Borkar:book} and the fact that the model satisfies conditions (A1)--(A4) of~\cite[pg~10--11]{Borkar:book}. The Martingale difference condition (A3) of \cite{Borkar:book} is satisfied due to the unbiasedness assumption of the state approximator.
  The result then follows from by combining the theorem given in~\cite[page 35]{Leslie_2004},
  \cite[Theorem 23]{Borkar_1997} along with ~\cite[Theorem
  2.2]{Borkar:book} and using a third fastest time scale for the state apparoximator.
\end{proof}
}

\section{Details about the network architecture, training, and hyperparameters}
\label{sec:network}

As explained in Sec.~\ref{sec:experiments}, the \acs{AIS}-generator consists
of four components: the history compression function $\ainfo$, the \acs{AIS}
update function $\aupdate$, the reward prediction function $\rewinfo$, and the
observation prediction kernel $\nextobs$. We model the first as an LSTM, where
the memory update unit of LSTM acts as $\aupdate$. We model $\rewinfo$,
$\nextobs$, and the policy $\apol$ as feed-forward neural networks. We
describe the details for each difficulty class of environment separately. In
the description below,  we use $\Linear(n,m)$ to denote a linear layer
$\Tanh(n,m)$ to denote a tanh layer, $\Relu(n,m)$ to denote a ReLU layer, and
$\LSTM(n,m)$ to denote an LSTM layer, where $n$ denotes the number of inputs
and $m$ denotes the number of outputs of each layer. The size of the input of
the outputs depend on the size of the observation and action spaces, which we
denote by $n_O$ and $n_A$, respectively as well as on the dimension of
\acs{AIS} and for the case of minigrid environments, the dimension of the
latent space for observations, we denote by $d_{\Ais}$ and $d_O$. We also use $\Conv(IC, OC, (FSx, FSy))$ to denote a 2D convolutional layer with $IC$, $OC$, $(FSx, FSy)$ represent the number of input channels, output channels and kernel size (along $x$ and $y$) respectively. Note that the strides are the same as the kernel size in this case. $\ELU$ represents Exponential Linear Unit and is used to model the prediction of variance. Finally, $\GMM(n_{\textup{comp}})$ represents a Gaussian Mixture Model with $n_{\textup{comp}}$ Gaussian components. Most of the details are common for both the AIS+KL and the AIS+MMD cases, we make a distinction whenever they are different by indicating KL or MMD.

\subsection{Details for low dimensional environments:}

\begin{itemize}
  \item \textsc{Environment Details:}
    \par\nopagebreak[4]
    \begin{tabular}{@{}cccc@{}}
      \toprule
      Environment & Discount & No.~of actions & No.~of obs. \\
      &     $\discount$ & $n_A$ & $n_O$ \\
      \midrule
      \textsc{Voicemail} & 0.95 & 3 & 2 \\
      \textsc{Tiger}     & 0.95 & 3 & 2 \\
      \textsc{CheeseMaze}& 0.7  & 4 & 7 \\
      \bottomrule
    \end{tabular}

    The discount factor for \textsc{CheeseMaze} is chosen to match with
    standard value used in that environment \citep{McCallum_1993}.
    
  \item \textsc{AIS and Network details:}
    \par\nopagebreak[4]
    \begin{tabular}{@{}llcl}
      $\bullet$ & Dimensions of AIS ($d_{\Ais}$) & : & $40$ \\
      $\bullet$ & Weight in \acs{AIS} loss ($\lambda$) (KL)& : & $0.0001$ \\
      			& Weight in \acs{AIS} loss ($\lambda$) (MMD)& : & $0.001$ \\
    \end{tabular}
    
    \begin{tabular}{@{}cccc@{}}
      \toprule
      $\ainfo$ & $\rewinfo$ & $\nextobs$ & $\apol$ \\
      \midrule
      $\Linear(n_O + n_A + 1, d_{\Ais})$ & $\Linear(n_A + d_{\Ais}, \tfrac12 d_{\Ais})$ &
      $\Linear(n_A + d_{\Ais}, \tfrac12 d_{\Ais})$ & $\Linear(d_{\Ais},d_{\Ais})$
      \\ \Arrow & \Arrow & \Arrow & \Arrow \\
      $\Tanh(d_{\Ais},d_{\Ais})$ & $\Tanh(\tfrac12d_{\Ais}, \tfrac12d_{\Ais})$ & 
      $\Tanh(\tfrac12d_{\Ais}, \tfrac12d_{\Ais})$ & $\Tanh(d_{\Ais},d_{\Ais})$ 
      \\ \Arrow & \Arrow & \Arrow & \Arrow \\
      $\LSTM(d_{\Ais},d_{\Ais})$ & $\Linear(\tfrac12d_{\Ais}, 1)$ & $\Linear(\tfrac12d_{\Ais}, n_O)$ &
        $\Linear(d_{\Ais}, n_A)$
      \\
      & & \Arrow & \Arrow \\
      & &  $\Softmax$ & $\Softmax$ \\
      \bottomrule
    \end{tabular}
    
  \item \textsc{Training details:}
    As explained in Section~\ref{sec:PORL}, we update the parameters after a
    rollout of $T$, which we call a \emph{training batch}. The choice of
    parameters for the training batch are as follows:
    \par\nopagebreak[4]
    \begin{tabular}{@{}llcl}
      $\bullet$ & Samples per training batch & : & $200$ \\
      $\bullet$ & Number of training batches & : & $10^5$ \\
    \end{tabular}

    In addition, we use the following learning rates:
    \par\nopagebreak[4]
    \begin{tabular}{@{}llcl}
      $\bullet$ & AIS learning rate & : & ADAM(0.003) \\
      $\bullet$ & Policy learning rate (KL)& : & ADAM(0.0006) \\
		        & Policy learning rate (MMD)& : & ADAM(0.0008) \\
    \end{tabular}
    \par\nopagebreak[4]
    In the above description, we use ADAM($\alpha$) to denote the choice of
    $\alpha$ parameter of ADAM. All other parameters have their default value.

  \item \textsc{Evaluation details:}
    \par\nopagebreak[4]
    \begin{tabular}{@{}llcl}
      $\bullet$ & No.~of batches after which evaluation is done& : & $500$ \\
      $\bullet$ & Number of rollouts per evaluation & : & $50$ \\
    \end{tabular}
\end{itemize}

\subsection{Details for moderate dimensional environments:}

\begin{itemize}
  \item \textsc{Environment Details:}
    \par\nopagebreak[4]
    \begin{tabular}{@{}cccc@{}}
      \toprule
      Environment & Discount & No.~of actions & No.~of obs. \\
      &     $\discount$ & $n_A$ & $n_O$ \\
      \midrule
      \textsc{Drone Surveillance} & 0.99 & 5 & 10 \\
      \textsc{Rock Sampling}      & 0.99 & 8 & 3  \\
      \bottomrule
    \end{tabular}
    
  \item \textsc{AIS and Network details:}
    \par\nopagebreak[4]
    \begin{tabular}{@{}llcl}
      $\bullet$ & Dimensions of AIS ($d_{\Ais}$) & : & $128$ \\
      $\bullet$ & Weight in \acs{AIS} loss ($\lambda$) (KL)& : & $0.0001$ \\
      			& Weight in \acs{AIS} loss ($\lambda$) (MMD)& : & $0.001$ \\
    \end{tabular}
    
    \begin{tabular}{@{}cccc@{}}
      \toprule
      $\ainfo$ & $\rewinfo$ & $\nextobs$ & $\apol$ \\
      \midrule
      $\LSTM(n_O + n_A + 1, d_{\Ais})$ & $\Linear(n_A + d_{\Ais}, \tfrac12 d_{\Ais})$ &
      $\Linear(n_A + d_{\Ais}, \tfrac12 d_{\Ais})$ & $\Linear(d_{\Ais},n_A)$
      \\ & \Arrow & \Arrow & \Arrow \\
      & $\Relu(\tfrac12d_{\Ais}, \tfrac12d_{\Ais})$ & 
      $\Relu(\tfrac12d_{\Ais}, \tfrac12d_{\Ais})$ & $\Softmax$
      \\ & \Arrow & \Arrow & \\
      & $\Linear(\tfrac12d_{\Ais}, 1)$ & $\Linear(\tfrac12d_{\Ais}, n_O)$ & \\
      & & \Arrow & \\
      & &  $\Softmax$ & \\
      \bottomrule
    \end{tabular}
    
  \item \textsc{Training details:}
    As explained in Section~\ref{sec:PORL}, we update the parameters after a
    rollout of $T$, which we call a \emph{training batch}. The choice of
    parameters for the training batch are as follows:
    \par\nopagebreak[4]
    \begin{tabular}{@{}llcl}
      $\bullet$ & Samples per training batch & : & $200$ \\
      $\bullet$ & Number of training batches & : & $10^5$ \\
    \end{tabular}

    In addition, we use the following learning rates:
    \par\nopagebreak[4]
    \begin{tabular}{@{}llcl}
      $\bullet$ & AIS learning rate & : & ADAM(0.003) \\
      $\bullet$ & Policy learning rate & : & ADAM(0.0007) \\
    \end{tabular}
    \par\nopagebreak[4]
    In the above description, we use ADAM($\alpha$) to denote the choice of
    $\alpha$ parameter of ADAM. All other parameters have their default value.

  \item \textsc{Evaluation details:}
    \par\nopagebreak[4]
    \begin{tabular}{@{}llcl}
      $\bullet$ & No.~of batches after which evaluation is done& : & $500$ \\
      $\bullet$ & Number of rollouts per evaluation & : & $100$ \\
    \end{tabular}
\end{itemize}

\subsection{Details for high dimensional environments:}

\begin{itemize}
  \item \textsc{Environment Details:}
    \par\nopagebreak[4]
    Note that here $n_O$ represents the number of possible observations that a
    general minigrid environment can have. With the actual rules of the
    environment plugged in, this number is smaller since some combinations of the encoded observation are not possible. The actual input that we get from the environment is a vector of size $147$ ($d_O$) which is basically an observation grid of $7 \times 7$ with $3$ channels containing characteristic information about the observation.
    
    \begin{tabular}{@{}ccccc@{}}
      \toprule
      Environment & Discount & No.~of actions & No.~of obs. & Obs. dimen.\\
      &     $\discount$ & $n_A$ & $n_O$ & $d_O$ \\
      \midrule
      \textsc{Minigrid Envs} & 0.99 & 7 & $(6 \times 11 \times 3)^{7 \times 7}$ & $7 \times 7 \times 3$ \\
      \bottomrule
    \end{tabular}
  \item \textsc{Autoencoder ($q$) details:}
    \par\nopagebreak[4]
    \begin{tabular}{@{}llcl}
      $\bullet$ & Latent space dimensions ($d_L$) & : & $64$ \\
	  $\bullet$ & Type of autoencoder used						& : & Basic autoencoder \\      
      $\bullet$ & Reconstruction Loss Criterion Used			& : & Mean Square Error \\
      
    \end{tabular}
    
    \begin{tabular}{@{}c@{}}
      \toprule
      $q$ \\
      \midrule
      $\Linear(d_O, \tfrac32 d_L)$		\\
      \Arrow 					\\
      $\Relu(\tfrac32 d_L, \tfrac32 d_L)$			\\
      \Arrow					\\
      $\Linear(\tfrac32 d_L, d_L)$ \\
      \bottomrule
    \end{tabular}
       
  \item \textsc{AIS and Network details:}
    \par\nopagebreak[4]
    \begin{tabular}{@{}llcl}
      $\bullet$ & Dimensions of AIS ($d_{\Ais}$) & : & $128$ \\
      $\bullet$ & Weight in \acs{AIS} loss ($\lambda$) & : & $0.1$ \\
      $\bullet$ & Number of GMM components used ($n_{\textup{comp}}$) (only for KL) & : & $5$ \\
    \end{tabular}
    
    \begin{tabular}{@{}cccc@{}}
      \toprule
      $\ainfo$ & $\rewinfo$ & $\nextobs$ & $\apol$ \\
      \midrule
      $\LSTM(d_L + n_A + 1, d_{\Ais})$ & $\Linear(n_A + d_{\Ais}, \tfrac12 d_{\Ais})$ &
      $\Linear(n_A + d_{\Ais}, \tfrac12 d_{\Ais})$ & $\Linear(d_{\Ais},d_{\Ais})$\\
      & \Arrow & \Arrow & \Arrow \\
      & $\Relu(\tfrac12d_{\Ais}, \tfrac12d_{\Ais})$ & 
      $\Relu(\tfrac12d_{\Ais}, \tfrac12d_{\Ais})$ & $\Relu(d_{\Ais}, d_{\Ais})$
      \\ & \Arrow & \Arrow & \Arrow\\
      & $\Linear(\tfrac12d_{\Ais}, 1)$ & $\Linear(\tfrac12d_{\Ais}, d_L)$ & $\Linear(d_{\Ais}, n_A)$\\
      & & & \Arrow \\
      & & & $\Softmax$ \\
      \bottomrule
    \end{tabular}

    For KL, $\nextobs$ is replaced by the following while other networks remain the same:
    \par\nopagebreak[4]
    \begin{tabular}{@{}ccc@{}}
      \toprule
      & $\nextobs$ & \\
      \midrule
      & $\Linear(n_A + d_{\Ais}, \tfrac12 d_{\Ais})$ & \\
      & \Arrow 	&				\\
      &$\Relu(\tfrac12d_{\Ais}, \tfrac12d_{\Ais})$ &\\
      \LLArrow & \Arrow	& \LRArrow				\\
      $\Linear(\tfrac12d_{\Ais}, d_L n_{\textup{comp}})$ &
	  $\ELU(\Linear(\tfrac12d_{\Ais}, d_L n_{\textup{comp}}))$ + $1$
      + $10^{-6}$ &
	  $\Softmax(\Linear(\tfrac12d_{\Ais}, n_{\textup{comp}}))$\\
      \LRArrow & \Arrow & \LLArrow					\\
	  & $\GMM(n_{\textup{comp}})$ & \\
      \bottomrule
    \end{tabular}
    \par\nopagebreak[4]
	Note that the third layer generates the mean vector of each component, the
    diagonal vector for variance of each component and the mixture weights of
    each component of the GMM model in the last layer.
    
  \item \textsc{Training details:}
    As explained in Section~\ref{sec:PORL}, we update the parameters after a
    rollout of $T$, which we call a \emph{training batch}. The choice of
    parameters for the training batch are as follows:
    \par\nopagebreak[4]
    \begin{tabular}{@{}llcll}
      $\bullet$ & Samples per training batch & : & $200$ \\
      $\bullet$ & Number of training batches & : & $2 \times 10^5$ & (MGKCS3R3,
      MGOM1Dl, MGOM1Dlh) \\
        & & & $10^5$ & (others) \\
    \end{tabular}

    In addition, we use the following learning rates:
    \par\nopagebreak[4]
    \begin{tabular}{@{}llcl}
      $\bullet$ & AIS learning rate & : & ADAM(0.001) \\
      $\bullet$ & Policy learning rate & : & ADAM(0.0007) \\
    \end{tabular}
    \par\nopagebreak[4]
    In the above description, we use ADAM($\alpha$) to denote the choice of
    $\alpha$ parameter of ADAM. All other parameters have their default value.

  \item \textsc{Evaluation details:}
    \par\nopagebreak[4]
    \begin{tabular}{@{}llcll}
      $\bullet$ & No.~of batches after which evaluation & : & $5000$
      & (MGKCS3R3, MGOM1Dl, MGOM1Dlh) \\
        & is done &  & $1000$ & (others) \\
      $\bullet$ & Number of rollouts per evaluation & : & $20$ \\
    \end{tabular}
\end{itemize}

\subsection{Details for PPO with LSTM and Critic:}

\begin{itemize}
  \item \textsc{Environment Details:}
    \par\nopagebreak[4]
    The environment details are the same as mentioned previously.
   
  \item \textsc{Network details:}
    \par\nopagebreak[4]
    
    \begin{itemize}
    \item
	Low and moderate dimensionality environments:
    \par\nopagebreak[4]
    \begin{tabular}{@{}ccc@{}}
      \toprule
      Feature Extractor & Actor Head & Critic Head\\
      \midrule
      $\LSTM(n_O, n_O)$ & $\Linear(n_O, 64)$ & $\Linear(n_O, 64)$\\
      & \Arrow & \Arrow \\
      & $\Tanh(64, 64)$ & $\Tanh(64, 64)$\\
      & \Arrow & \Arrow \\
      & $\Linear(64, n_A)$ & $\Linear(64, 1)$\\
      & \Arrow & \\
      & Softmax & \\
      \bottomrule
    \end{tabular}
      \item High dimensionality environments:
        \par\nopagebreak[4]
	\begin{tabular}{@{}llcll}
		$\bullet$ & Observation tensor & : & $7 \times 7 \times 3$ \\
		$\bullet$ & Embedding size ($d_E$) & : & $64$ \\
	\end{tabular}
	
    \begin{tabular}{@{}ccc@{}}
      \toprule
      Conv. Feature Extractor & Actor Head & Critic Head\\
      \midrule
      $\Conv(3, \tfrac14 d_E, (2,2))$ & $\Linear(d_E, d_E)$ & $\Linear(d_E, d_E)$\\
      \Arrow & \Arrow & \Arrow \\
      $\Relu$ & $\Tanh(d_E, d_E)$ & $\Tanh(d_E, d_E)$\\
      \Arrow & \Arrow & \Arrow \\
      $\MP$ & $\Linear(d_E, n_A)$ & $\Linear(d_E, 1)$\\
      \Arrow & \Arrow & \\
      $\Conv(\tfrac14 d_E, \tfrac12 d_E, (2,2))$ & Softmax & \\
      \Arrow & & \\
      $\Relu$ & & \\
      \Arrow & & \\
      $\Conv(\tfrac12 d_E, d_E, (2,2))$ & & \\
      \Arrow & & \\
      $\Relu$ & & \\
      \Arrow & & \\
      $\LSTM(d_E, d_E)$ & & \\
      \bottomrule
    \end{tabular}
\end{itemize}
    
  \item \textsc{Training details:}
    \par\nopagebreak[4]
    \begin{tabular}{@{}llcll}
      $\bullet$ & Number of parallel actors & : & $64$ \\
      $\bullet$ & Number of training batches & : & $4\times10^{7}$ & (MGKCS3R3, MGOM1Dl, MGOM1Dlh) \\
        		& & & $2\times10^{7}$ & (others) \\
      $\bullet$ & Epochs per training batch & : & $4$ \\
      $\bullet$ & Samples per training batch & : & $1280$ \\
      $\bullet$ & Frames per parallel actor & : & $40$ \\
      $\bullet$ & GAE ($\lambda_{GAE}$) & : & $0.99$ \\
      $\bullet$ & Trajectory recurrence length & : & $20$ \\
    \end{tabular}

    In addition, we use ADAM with the following details:
    \par\nopagebreak[4]
    \begin{tabular}{@{}llcl}
      $\bullet$ & Learning rate $\alpha$ & : & 0.0001 \\
      $\bullet$ & ADAM parameter $\epsilon$ & : & 0.00001 \\
    \end{tabular}
    \par\nopagebreak[4]

  \item \textsc{Evaluation details:}
    \par\nopagebreak[4]
    \begin{tabular}{@{}llcll}
      $\bullet$ & No.~of batches after which evaluation &   &  \\
        		& is done 								& : & $200$ \\
      $\bullet$ & Rollouts used for evaluation		 	& : & All recent episodes completed by all actors\\
    \end{tabular}
\end{itemize}

\subsection{Details about hyperparameter tuning}

Hyperparameter tuning was carried out by searching a grid of values, but exhaustive grid search was not carried out due to the prohibitive computational cost. Instead, coarse values were used initially as starting points and finer tuning was done around promising values, which was essentially an iterative process of performing experiments, observing results and trying similar parameters to the ones generating good results. Hyperparameters observed in each previous environment class (low, moderate, high dimensionality) were used as a starting point for the search in the new environment class.

Performance was quite sensitive to different learning rates used for the AIS and policy in most environments. Performance generally improved or remained the same when a larger AIS State Size was used (values considered were 128, 256, 512 for moderate/high-dimensional environments and 5, 10, 20, 40 for low-dimensional environments), although in some cases, it was more unstable during training. $\lambda$ values considered were between 0 and 1 and generally only made a difference (in terms of performance results) when the rewards were very large. The choice of activation function between ReLU and Tanh did not seem to make a significant difference for the considered environments.

%\input sections/app-main-algorithm
%\input sections/appendix

% \newpage
% 
% \input sections/old

\end{document}